%% file: main.tex
\documentclass{article} 
\usepackage{wda_preprint}
\usepackage{fontspec,xeCJK}
\setCJKmonofont{FandolSong-Regular.otf}
\setCJKsansfont{FandolHei-Regular.otf}
\usepackage{fvextra,longtable,array,xurl,graphicx,adjustbox,needspace,tcolorbox}
\tcbuselibrary{breakable,skins}
\newlength{\RecordTableWidth}

\input{math_commands.tex}

\usepackage{hyperref}
\hypersetup{hypertexnames=false,hidelinks,pdftitle={Streamlined Reflective Evolution for Task-Adaptive Self-Refinement Pipelines},pdfauthor={Xiaofan Zhou and Lu Cheng}}
\usepackage{url}
\usepackage{placeins}
\usepackage{algorithm}
\usepackage{algpseudocode}
\usepackage{tikz}
\usetikzlibrary{arrows.meta,positioning,shapes.geometric,calc,fit}

\title{Streamlined Reflective Evolution\\ for Task-Adaptive Self-Refinement Pipelines}

\author{Xiaofan Zhou \qquad Lu Cheng \\
Pennsylvania State University \\
\texttt{\{xpz5429,lqc5822\}@psu.edu}}

\iclrfinalcopy
\begin{document}

\maketitle
\pagestyle{plain}

\begin{abstract}
Reflective prompt optimization improves large language model (LLM) systems without updating model weights,
but fixed architectures constrain how self-refinement is organized.
We introduce Workflow-Designing Agents (WDA), a framework for streamlined reflective evolution
of task-adaptive self-refinement pipelines.
Starting from a minimal prompt, WDA jointly evolves stage instructions and their sequential structure.
During evolution, we find that repeated revisions can accumulate redundant instructions in a single prompt.
In WDA, we propose to address this problem with \textsc{Split}, which redistributes these instructions across specialized stages.
Three-example reflection and local screening guide selective search, while calibration
scores guide Pareto admission and rollback of unhelpful trailing updates.
The resulting pipelines are task-adaptive: their instructions and depth are learned from task data,
then fixed for all test inputs within that task.
We evaluate WDA on five benchmarks spanning knowledge, mathematical reasoning,
multi-hop question answering, and instruction following.
On Qwen3.5-9B, WDA achieves an average score of 51.24\%, improving over the initial
solver by 8.63 percentage points and the variant without \textsc{Split} by 2.60 points.
On GPT-4.1-mini, it achieves 49.00\%, with corresponding gains of 5.62 and 3.69 points.
These results support task-adaptive self-refinement as a complementary
direction to broader agentic workflow search.
\end{abstract}

\section{Introduction}
LLMs are increasingly used for knowledge-intensive question answering and reasoning \citep{lewis2020rag,wei2022chain}. Their performance depends on instructions and examples, including in mathematical and medical tasks \citep{wei2022chain,nori2023medprompt}. Designing effective prompts can require substantial manual effort \citep{pryzant2023automatic}. Automatic prompt optimizers instead use task feedback to search for better instructions \citep{pryzant2023automatic,yang2024optimizers}. The cost of this search motivates sample-efficient prompt adaptation \citep{agrawal2026gepa}.

One widely used remedy is self-refinement: the model produces an answer, critiques it, and
revises it, repeating the loop at inference time without any weight update
\citep{madaan2023selfrefine}. The recipe is attractive because it needs no training data and
no second model. It is also fragile. When the critique comes from the model itself and no
external signal is available, self-correction frequently fails to improve reasoning and can
make accuracy worse than the first attempt \citep{huang2024selfcorrect}. These findings
motivate examining two commonly hand-specified choices: the number of refinement rounds and
the critique instruction. A fixed depth and a generic request to find mistakes may not
provide the task-specific checks needed to improve the initial answer.

A separate line of work makes prompts themselves learnable. Automatic prompt optimizers search
over natural-language instructions using calibration data, and reflective optimizers use the
system's own execution traces to diagnose failures and propose targeted rewrites. Genetic-Pareto (GEPA)
exemplifies this approach: it collects trajectories and textual feedback, has an LLM
propose reflective mutations, and preserves complementary candidates by instance-level Pareto
selection \citep{agrawal2026gepa}. It can therefore optimize critique instructions, but takes
the system architecture as given. Broader approaches such as Automated Design of Agentic Systems (ADAS) and
Automating Agentic Workflow Generation (AFlow) search agentic designs or executable workflows \citep{hu2025adas,zhang2025aflow}.
Our focus is narrower: learning the depth and division of responsibilities within sequential
self-refinement, which can itself serve as a component of a larger workflow.

We propose WDA, which streamlines GEPA-inspired reflective evolution to learn task-adaptive
self-refinement pipelines. Starting from one minimal, task-agnostic instruction, it samples a feasible operator
and compatible configuration. \textsc{Replace} specializes an instruction, \textsc{Add} appends
a refinement stage, and \textsc{Split} decomposes an overloaded prompt into consecutive stages.
\textsc{Replace} and \textsc{Add} proposals must improve a three-example batch before calibration evaluation;
\textsc{Split} proposals compete on a shared batch. Pareto admission retains non-dominated
candidates, and calibration-guided rollback discards trailing updates without a calibration gain.
Local screening and rollback keep evolution selective. Task data determine what each stage
checks and how many stages to retain; the selected pipeline remains fixed during test inference.

Our contributions are threefold. First, we formulate task-adaptive self-refinement:
reflection learns stage instructions, and calibration guides pipeline structure and depth,
instead of prescribing a fixed refine loop. Second, we introduce \textsc{Split} to redistribute
instructions from an evolved prompt across specialized sequential stages.
This is motivated by the redundant instructions that can accumulate during evolution
and the need for more focused refinement stages. Third, we evaluate WDA on
five benchmarks with a proprietary and an open-weight backbone, comparing against direct
inference, fixed-depth self-refinement, fixed-architecture prompt optimizers, and
preference-based fine-tuning.

\section{Related work}
\subsection{LLM Recursive Self-Improvement}
We use recursive self-improvement here to refer to repeated LLM-guided revisions of prompts
or agentic programs using execution feedback, rather than autonomous updates to model weights.
Early prompt optimizers use textual gradients, black-box search, or evolutionary operators
\citep{pryzant2023automatic,yang2024optimizers,guo2024evoprompt,fernando2024promptbreeder}.
GEPA uses trajectories and textual feedback to propose reflective mutations and preserves
complementary candidates through instance-level Pareto selection \citep{agrawal2026gepa}.
Consensus-Evolve (C-Evolve) instead evolves prompt groups using each member's contribution to majority-vote
consensus \citep{li2026cevolve}, while Multi-Agent System Prompt Optimization via Bandits (MASPOB) optimizes role-specific prompts with
topology-aware dependency modeling in a fixed multi-agent workflow \citep{hong2026maspob}.
Mixture-of-Prompts (MoP) routes inputs to specialized prompts associated with semantic regions \citep{wang2024mop}; combining such routing with region-specific WDA pipelines could make refinement instructions and depth input-adaptive.

Beyond prompt optimization, ADAS uses a meta-agent to propose and evaluate agentic systems
represented in code \citep{hu2025adas}. AFlow searches code-represented workflows with
Monte Carlo Tree Search, execution feedback, and reusable operators such as Ensemble and
Review \& Revise \citep{zhang2025aflow}. These approaches can change workflow structure;
they should not be characterized as fixed-architecture prompt optimizers.
WDA targets a more focused search space: the instructions, depth, and stage responsibilities
of sequential self-refinement. Self-refinement is a common workflow building block,
explicitly represented by AFlow's Review \& Revise operator. WDA could therefore supply an
optimized refinement component to a broader workflow optimizer, although we do not evaluate
such an integration here.

\subsection{Self-Refinement}
Self-Refine iterates answer generation, feedback, and revision with one model and no weight
updates \citep{madaan2023selfrefine}. Without an external verifier, however, intrinsic
self-correction can fail to improve reasoning or reduce accuracy \citep{huang2024selfcorrect}.
Fixed-depth pipelines with generic critique instructions may spend unnecessary computation
or miss task-specific errors. WDA retains sequential refinement but learns its depth through
calibration and its instructions through reflection on failed traces.

Relatedly, multi-agent systems elicit diverse candidate solutions and let agents critique or
aggregate one another's outputs. Multiagent debate exchanges independently generated reasoning
traces \citep{du2024multiagent}, role-based systems such as MedAgents organize domain experts
into multi-round deliberation \citep{tang2024medagents}, and Mixture-of-Agents uses a layered
proposer--aggregator architecture \citep{wang2024mixture}. Controlled comparisons find that
debate does not reliably beat self-consistency \citep{wang2023selfconsistency} or simple ensembles without careful configuration
\citep{smit2024mad}. These systems differ from ours in both structure and provenance: their
modules answer in parallel and are combined by a fixed protocol, whereas our modules form an
ordered refinement chain, and their composition is specified in advance rather than discovered.

\section{Problem statement}
We study how to learn a task-adaptive, multi-prompt LLM system from task data,
starting from a single simple prompt and keeping the learned pipeline fixed at test time.
Following prior work \citep{agrawal2026gepa}, we represent the system as
$\Phi=(M,C,\mathcal{X},\mathcal{Y})$, where
$M=\langle M_1,\ldots,M_{|M|}\rangle$ is a sequence of LLM modules,
$C$ specifies the control logic, and $\mathcal{X}$ and $\mathcal{Y}$
denote the input and output schemas, respectively.
Each module is defined as
$M_i=(\pi_i,\theta_i,\mathcal{X}_i,\mathcal{Y}_i)$,
where $\pi_i$ is its prompt, $\theta_i$ denotes its model weights,
and $\mathcal{X}_i$ and $\mathcal{Y}_i$ specify its input and output
schemas. We adopt sequential control logic: the first module receives
the input question, each subsequent module takes the original question and the preceding
module's output as input, and the last module produces the final answer.

Initially, the system contains only one module with a simple prompt.
During calibration, the system applies \textsc{Replace}, \textsc{Add}, and \textsc{Split},
automatically determining both the number of modules and their
instructions to obtain a multi-prompt system.
We denote the module prompts by
$\Pi_{\Phi}=\langle\pi_1,\ldots,\pi_{|M|}\rangle$
and the corresponding model weights by
$\Theta_{\Phi}=\langle\theta_1,\ldots,\theta_{|M|}\rangle$.
The model weights remain fixed; we optimize the module structure and prompts
contained in the complete system $\Phi$.

The objective is to maximize expected performance on task instances
$(x,m)\sim\mathcal{T}$, where $\mathcal{T}$ is the task distribution,
$x\in\mathcal{X}$ is an input, and $\mathcal{M}$ is the space of evaluation
metadata $m$, such as reference answers, scoring rubrics, or unit tests.
Given the system output $y=\Phi(x;\Pi_{\Phi},\Theta_{\Phi})$, abbreviated as $\Phi(x)$,
a task-specific metric
$\mu:\mathcal{Y}\times\mathcal{M}\rightarrow[0,1]$
measures its quality with respect to $m$.
The optimization problem is
\begin{equation}
\Phi^{\mathrm{opt}}
=
\operatorname*{arg\,max}_{\Phi}
\mathbb{E}_{(x,m)\sim\mathcal{T}}
\left[
\mu\!\left(\Phi(x),m\right)
\right].
\end{equation}
Here $\mathbb{E}$ denotes expectation and $\arg\max$ selects a maximizing system.
In practice, calibration data are used to estimate this objective
and guide the evolution from the initial single-prompt system
to a system of sequential prompt modules.
\section{Method}
\begin{figure}[t]
\centering
\includegraphics[width=\linewidth]{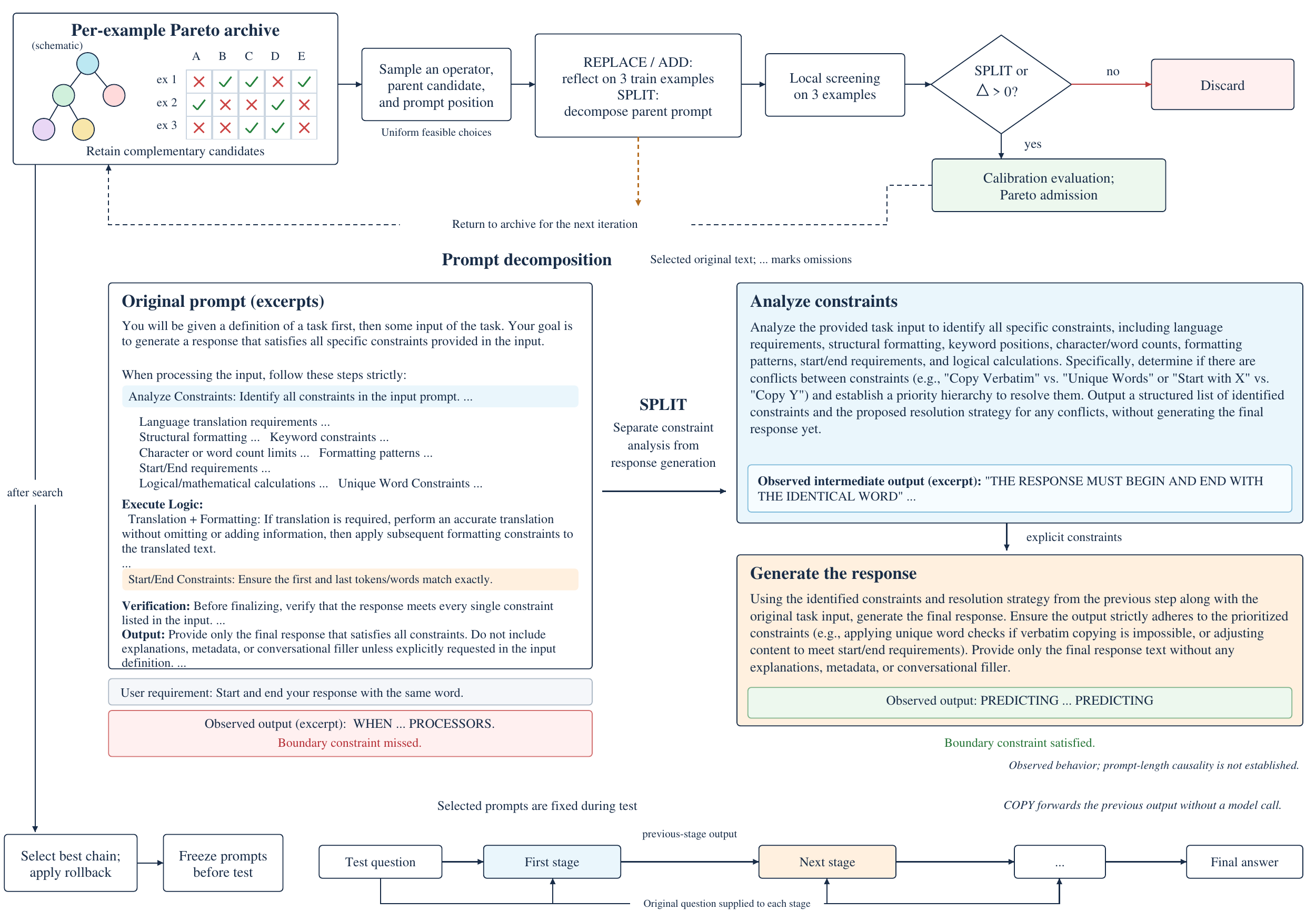}
\caption{WDA: reflective evolution with local screening, Pareto admission, and fixed test-time execution. The example illustrates \textsc{Split} separating constraint analysis from response generation; these excerpts do not establish prompt-length causality. Each stage receives the question and preceding output; \textsc{Copy} forwards that output without a model call.}
\label{fig:method-overview}
\end{figure}

WDA learns task-adaptive pipelines through minimal initialization (Section~\ref{sec:init}), reflective structure and prompt evolution (Section~\ref{sec:evolution}), and sequential test inference (Section~\ref{sec:collab}). Its inputs are an initial system $\Phi^{(0)}$, training set $\mathcal{D}_{\mathrm{train}}$ for reflection, calibration set $\mathcal{D}_{\mathrm{cal}}$ for candidate evaluation, task metric $\mu$, reflection LLM $R$, and evolution budget $B$. It evolves module structure and prompts while keeping weights $\Theta_{\Phi}$ fixed. Figure~\ref{fig:method-overview} summarizes the workflow.

\subsection{Initialization}
\label{sec:init}
We initialize $\Phi^{(0)}=(M^{(0)},C^{(0)},\mathcal{X},\mathcal{Y})$ with one module,
$M^{(0)}=\langle M_1^{(0)}\rangle$, where $M_1^{(0)}=(\pi_1^{(0)},\theta_1,\mathcal{X},\mathcal{Y})$.
Control $C^{(0)}$ passes input $x$ to this module and returns its answer.
The sole prompt $\pi_1^{(0)}$ asks it to solve the problem, without task-specific expertise
or a prescribed decomposition; hence $\Pi_{\Phi^{(0)}}=\langle\pi_1^{(0)}\rangle$.

We score this system on calibration data using $\mu$. Modules are introduced incrementally,
and calibration-guided rollback retains useful changes rather than expanding unconditionally.
This initialization makes refinement optional: tasks that do not benefit from an additional
stage can retain a compact solver. The search therefore allocates complexity in response to
observed errors instead of committing to a multi-stage architecture before evaluation.
Repeated revisions can introduce redundant instructions, motivating \textsc{Split} to distribute
responsibilities across focused stages.

\subsection{Incremental Prompt Evolution}
\label{sec:evolution}
Inspired by GEPA's instance-level feedback \citep{agrawal2026gepa}, we store admitted systems as $\mathcal{P}[k]=\Phi_k$, where $k$ is a candidate identifier (ID) and $\pi_{k,i}$ is its $i$-th prompt. Let $\mathcal{D}_{\mathrm{cal}}=\{(x_j,m_j)\}_{j=1}^{N_{\mathrm{cal}}}$ contain $N_{\mathrm{cal}}$ examples. The score matrix has entries $S_{k,j}=\mu(\Phi_k(x_j),m_j)$; $S_{k,:}$ denotes candidate $k$'s score vector. Let $A$ contain the current Pareto-front candidate IDs. For any candidate-ID set $I$, $\ell\succ k$ means that $\ell$ dominates $k$, and the non-dominated (ND) operator keeps candidates dominated by none:
\begin{equation}
\begin{aligned}
\ell\succ k
&\iff (\forall j,\ S_{\ell,j}\geq S_{k,j})
\ \land\ (\exists j,\ S_{\ell,j}>S_{k,j}),\\
\operatorname{ND}(I;S)
&=\{k\in I:\not\exists\ell\in I\text{ with }\ell\succ k\}.
\end{aligned}
\label{eq:pareto-selection}
\end{equation}
This preserves complementary correctness patterns rather than retaining only the highest average score.
Unlike GEPA's frequency-weighted selection, our implementation samples uniformly from
operator-compatible Pareto-front candidates.

We first sample an operator, then a compatible parent and target. Let $\mathcal{I}_o(k)$
be the eligible prompt positions in $\Phi_k$ for operator $o$, let
$A_o=\{k\in A:\mathcal{I}_o(k)\neq\emptyset\}$ be the compatible parent IDs, and let
$\mathcal{O}(A)=\{o:A_o\neq\emptyset\}$ be the available operators.
\textsc{Replace} rewrites an active prompt; \textsc{Add} appends a module when capacity permits;
\textsc{Split} replaces an eligible evolved prompt with two consecutive modules.
The unchanged initial prompt is not eligible for \textsc{Split}: it must first be revised by \textsc{Replace}.
These restrictions and the module limit $K_{\max}$ determine $\mathcal{I}_o(k)$.
At search iteration $t$, we sample an operator $o_t$, parent ID $k_t$, and position $i_t$.
The uniform distribution $\operatorname{Unif}(V)$ gives each element of a finite set $V$ probability $1/|V|$:
\[
o_t\sim\operatorname{Unif}(\mathcal{O}(A)),\qquad
k_t\sim\operatorname{Unif}(A_{o_t}),\qquad
i_t\sim\operatorname{Unif}(\mathcal{I}_{o_t}(k_t)).
\]
For \textsc{Add}, $\mathcal{I}_{o_t}(k_t)$ contains only the next insertion position, and
$\pi_{k_t,i_t}=\emptyset$ denotes the absence of an instruction at that position.

\paragraph{Reflection and local screening.}
The reflection model $R$ is the same configured backbone as the solver, with fixed weights.
For \textsc{Replace} and \textsc{Add}, it receives the selected instruction, the chosen operator, and three uniformly sampled
training examples $\mathcal{B}_t$, including their questions, intermediate reasoning traces
$\tau_t$, and reference-answer feedback $F_t$.
It returns a diagnosis of reusable errors and a prompt proposal $r_t$;
the instruction requests general strategies rather than copying example-specific answers.
For \textsc{Split}, the LLM sees only the parent instruction and decomposition requirements,
not the sampled examples or reference answers, and returns two complementary prompts.
Write $U_t=(\mathcal{B}_t,\tau_t,F_t)$ for \textsc{Replace}/\textsc{Add} and $U_t=\emptyset$ for \textsc{Split}.
No reference answers are provided at test time. \textsc{Replace} and \textsc{Add} proposals must improve
their three-example batch. For \textsc{Split}, valid decompositions are ranked on a shared batch,
and the highest-scoring proposal proceeds to calibration even without a local improvement.
We denote proposal generation and, for \textsc{Split}, this local proposal selection by $\operatorname{Reflect}$;
it returns $r_t$ or skips the iteration if no valid proposal is obtained. The batch is used
to score \textsc{Split} proposals, not as input to $R$. Applying operator $o_t$ produces
$\widetilde{\Phi}_t$; $\Delta_t$ is its mean batch-score gain and $g_t$ indicates whether it passes local screening:
\begin{equation}
\begin{aligned}
&r_t=\operatorname{Reflect}(R,\Phi_{k_t},i_t,o_t,U_t,\mathcal{B}_t),\\
&
\widetilde{\Phi}_t=o_t(\Phi_{k_t},i_t,r_t),\\
&\Delta_t=\frac{1}{|\mathcal{B}_t|}\sum_{(x,m)\in\mathcal{B}_t}
\bigl[\mu(\widetilde{\Phi}_t(x),m)-\mu(\Phi_{k_t}(x),m)\bigr],\\
&g_t=\mathbf{1}\!\left[o_t=\textsc{Split}\ \lor\ \Delta_t>0\right].
\end{aligned}
\label{eq:mutation-acceptance}
\end{equation}
Here $\mathbf{1}[\cdot]$ equals 1 when its condition holds and 0 otherwise.
Candidates with $g_t=1$ are evaluated on the full calibration set. Their configurations
and score vectors are appended to $\mathcal{P}$ and $S$ only if they enter the Pareto
front; candidates they dominate are removed from $A$. Duplicate configurations are rejected.
The archive has capacity $C_{\max}$. The $\operatorname{UpdateFront}$ operation first applies
ND, then retains the best mean-scoring member and prioritizes complementary instance
coverage when pruning to capacity. Historical records remain available for rollback;
$H[k]$ stores candidate $k$'s parent ID. They are not all eligible for sampling.

For calibration-guided rollback, following $H$ gives the lineage $\mathcal{L}(\Phi)=\langle\Phi^{(0)},\ldots,\Phi^{(L)}=\Phi\rangle$, where $L$ counts ancestral updates and $s$ indexes checkpoints, not global iterations. Let $J_{\mathrm{cal}}$ be the mean calibration score and $s^*$ the earliest best checkpoint. Rollback returns that checkpoint:
\begin{equation}
\begin{aligned}
J_{\mathrm{cal}}(\Phi)
&=\frac{1}{N_{\mathrm{cal}}}
\sum_{j=1}^{N_{\mathrm{cal}}}\mu(\Phi(x_j),m_j),\\
s^*(\Phi)
&=\min\operatorname*{arg\,max}_{0\leq s\leq L}
J_{\mathrm{cal}}(\Phi^{(s)}),\\
\operatorname{Rollback}(\Phi,H)&=\Phi^{(s^*(\Phi))}.
\end{aligned}
\label{eq:rollback}
\end{equation}
This discards trailing updates without a calibration gain, but need not reduce module count if they only rewrite prompts. In Algorithm~\ref{alg:mope}, $\operatorname{Evaluate}$ returns per-example scores; $S'$ and $A'$ are temporary scores and front IDs. Iteration $t$ also serves as a candidate ID, so rejected iterations leave gaps. We omit iteration subscripts inside the loop. The returned $\Phi^*$ is the selected system, not a guarantee of the ideal optimum $\Phi^{\mathrm{opt}}$.

\begin{algorithm}[t]
\caption{WDA}
\label{alg:mope}
\begin{algorithmic}[1]
\Require Initial system $\Phi^{(0)}$, $\mathcal{D}_{\mathrm{train}}$, $\mathcal{D}_{\mathrm{cal}}$, metric $\mu$, reflection model $R$, budget $B$
\Require Module limit $K_{\max}$, archive capacity $C_{\max}$; $B$ counts search iterations
\Ensure Optimized system $\Phi^*$
\Statex \textbf{Notation:} $\mathcal{P}[k]=\Phi_k$: stored system; $S$: calibration scores; $A$: front IDs; $H$: parent map.
\Statex $\mathcal{O}(A)$: available operators; $A_o$: compatible parent IDs; $\mathcal{I}_o(k)$: eligible prompt positions.
\Statex Uniform sampling gives every eligible choice equal probability.
\State $\mathcal{P}[0]\gets\Phi^{(0)}$; $S_{0,:}\gets\Call{Evaluate}{\Phi^{(0)},\mathcal{D}_{\mathrm{cal}},\mu}$
\State $A\gets\{0\}$; $H[0]\gets\textsc{None}$ \Comment{Initial front; no parent}
\For{$t=1,\ldots,B$}
    \State Sample $o$ uniformly from $\mathcal{O}(A)$, then $k$ uniformly from $A_o$
    \State Sample $i$ uniformly from $\mathcal{I}_o(k)$ \Comment{Eligible prompt position}
    \State $\mathcal{B}\gets$ three examples sampled from $\mathcal{D}_{\mathrm{train}}$
    \State Collect traces $\tau$ of $\Phi_k$ and reference-answer feedback $F$ on $\mathcal{B}$
    \State $U\gets\emptyset$ for \textsc{Split}; otherwise $U\gets(\mathcal{B},\tau,F)$
    \State $r\gets\Call{Reflect}{R,\Phi_k,i,o,U,\mathcal{B}}$; skip if no valid proposal
    \State $\widetilde{\Phi}\gets o(\Phi_k,i,r)$
    \State Compute $\Delta$ and local screening indicator $g$ using Eq.~\ref{eq:mutation-acceptance}
    \If{$g=1$ and $\widetilde{\Phi}$ is not a duplicate}
        \State $\widetilde{s}\gets\Call{Evaluate}{\widetilde{\Phi},\mathcal{D}_{\mathrm{cal}},\mu}$
        \State $S'\gets S$; $S'_{t,:}\gets\widetilde{s}$ \Comment{Candidate's calibration scores}
        \State $A'\gets\Call{UpdateFront}{A\cup\{t\},S',C_{\max}}$
        \If{$t\in A'$} \Comment{Admit only Pareto-front candidates}
            \State $\mathcal{P}[t]\gets\widetilde{\Phi}$; $S\gets S'$; $H[t]\gets k$; $A\gets A'$
        \EndIf
    \EndIf
\EndFor
\State $\widehat{k}\gets\arg\max_{k\in A}J_{\mathrm{cal}}(\Phi_k)$ \Comment{Best mean-score candidate}
\State $\Phi^*\gets\Call{Rollback}{\Phi_{\widehat{k}},H}$ using Eq.~\ref{eq:rollback}
\State \Return $\Phi^*$
\end{algorithmic}
\end{algorithm}

\subsection{Test Inference}
\label{sec:collab}
At test time, the calibration-selected system $\Phi$ and its prompts are fixed: no reflection, reference-answer feedback, or further search is used. Its modules execute sequentially according to $C$. Let $K=|M|$ count active LLM modules, excluding pass-through \textsc{Copy} slots, and let $h_i$ denote the output of module $i$:
\begin{equation}
\begin{aligned}
h_1&=M_1(x;\pi_1,\theta_1),\\
h_i&=M_i(x,h_{i-1};\pi_i,\theta_i),\quad i=2,\ldots,K,
\qquad y=h_K.
\end{aligned}
\end{equation}
Each active module is one inference step (one LLM call). It can refine, verify, or extend
the preceding result. A \textsc{Copy} slot forwards that output without an LLM call. The learned chain is
reused for all test inputs in a task, with no online prompt evolution.
Consequently, the inference procedure requires neither training examples nor reference-answer
feedback. Task-adaptive thus refers to learning instructions and depth for a task, not adapting
to individual test inputs. Input-adaptive stopping remains an extension discussed in
Section~\ref{sec:ablation}.
\section{Experiments}
\subsection{Research questions}
We organize our experiments around three research questions (RQs). \textbf{RQ1:} Does learning the refinement structure---how many stages and what each one checks---improve task performance over fixed-depth self-refinement and fixed-architecture prompt optimization? \textbf{RQ2:} How does \textsc{Split} affect the number of inference calls of the discovered pipelines? \textbf{RQ3:} How does test performance vary across inference steps in pipelines discovered with and without \textsc{Split}?

\subsection{Datasets}
We evaluate on five benchmarks that cover complementary capabilities and remain unsaturated for the two tested backbones under our evaluation setup. The initial-solver scores in Tables~\ref{tab:main-results} and~\ref{tab:gpt-results} indicate substantial headroom for improvement. \textbf{SuperGPQA} evaluates graduate-level knowledge and reasoning across a broad range of academic disciplines \citep{du2025supergpqa}. \textbf{LiveBench Math} contains challenging and contamination-resistant mathematical problems drawn from recently released sources \citep{white2025livebench}. \textbf{BIG-Bench Extra Hard (BBEH)} evaluates diverse reasoning skills using substantially more difficult successors to BIG-Bench Hard tasks \citep{suzgun2023challenging,kazemi2025bbeh}. \textbf{HotpotQA} tests multi-hop question answering over multiple supporting documents \citep{yang2018hotpotqa}. Finally, \textbf{IFBench} evaluates generalization to diverse, verifiable instruction-following constraints \citep{pyatkin2025ifbench}. We use each benchmark's official evaluator and report its default task score, with higher values indicating better performance.

\subsection{Models}
We use \texttt{gpt-4.1-mini} \citep{openai2025gpt41mini} and the open-weight \texttt{Qwen3.5-9B} \citep{qwen2026qwen35}. WDA uses the same backbone for reflection and solving. Prompt-based methods keep weights fixed; Direct Preference Optimization (DPO) \citep{rafailov2023dpo} provides a weight-updating comparison for Qwen3.5-9B.

\subsection{Baselines}
We compare our method with the initial solver, Self-Refine, two prompt optimizers, and, for Qwen3.5-9B, DPO. The \textbf{initial solver} uses the task prompt without optimization or additional modules. \textbf{Self-Refine} applies a fixed generate--critique--revise workflow \citep{madaan2023selfrefine}. \textbf{Multi-prompt Instruction Proposal Optimizer version 2 (MIPROv2)} optimizes instructions and demonstrations for a fixed program \citep{opsahlong2024mipro}. \textbf{GEPA} reflectively evolves prompts using execution feedback and instance-level Pareto selection \citep{agrawal2026gepa}. \textbf{DPO} updates model weights using preference optimization. We also compare with \textbf{WDA without \textsc{Split}}, which disables the prompt-decomposition operator.

\subsection{Experimental setup}
For the Qwen3.5-9B results reported below, WDA uses a budget of at most 20 search iterations, with a minimum of 10 iterations and early stopping after five iterations without improvement. The recorded WDA runs complete 10--20 iterations. Solver and reflection outputs are capped at 4,096 tokens per call; task evaluation uses temperature zero, and reflection uses temperature 0.7. The recorded runs use a train/calibration/test ratio of 1:2:2, with at most 300 test examples per benchmark. Each reflection candidate uses three sampled training examples. The recorded MIPROv2 baseline uses instruction-only optimization with 20 trials and no few-shot demonstrations. Optimizer iterations do not imply equal token or model-call budgets. We select the final WDA chain using calibration performance and apply the rollback rule in Eq.~\ref{eq:rollback}; test scores are used only for reporting.

\subsection{Main Results}
\label{sec:main-results}
Tables~\ref{tab:main-results} and~\ref{tab:gpt-results} report Qwen3.5-9B and GPT-4.1-mini results, respectively, across the same five benchmarks. We report the unweighted arithmetic average (Avg.) of the five displayed dataset scores for each method.

\begin{table}[t]
\centering
\small
\setlength{\tabcolsep}{5pt}
\renewcommand{\arraystretch}{1.12}
\caption{Test performance (\%) of Qwen3.5-9B across five benchmarks, including the DPO comparison. Avg. is the unweighted mean of the five displayed benchmark scores. Bold indicates the highest score in each column.}
\label{tab:main-results}
\begin{tabular}{lrrrrrr}
\hline
Method & SuperGPQA & \shortstack{LiveBench\\Math} & IFBench & BBEH & HotpotQA & Avg. \\
\hline
Initial solver & 53.00 & 57.38 & 38.67 & 23.33 & 40.67 & 42.61 \\
MIPROv2 & 51.33 & 51.64 & 31.33 & 22.00 & 49.33 & 41.13 \\
GEPA & 51.67 & 59.84 & 41.00 & 23.33 & 51.00 & 45.37 \\
Self-Refine & 54.33 & 65.57 & \textbf{47.67} & 25.67 & 39.00 & 46.45 \\
DPO & 50.00 & 57.38 & 40.33 & 21.67 & 40.00 & 41.88 \\
WDA (no \textsc{Split}) & 53.00 & 67.21 & 42.33 & 29.33 & 51.33 & 48.64 \\
WDA (\textsc{Split}) & \textbf{56.00} & \textbf{68.85} & \textbf{47.67} & \textbf{31.67} & \textbf{52.00} & \textbf{51.24} \\
\hline
\end{tabular}
\end{table}

\begin{table}[t]
\centering
\small
\setlength{\tabcolsep}{6pt}
\renewcommand{\arraystretch}{1.12}
\caption{Test performance (\%) of GPT-4.1-mini across five benchmarks. Avg. is the unweighted mean of the five displayed benchmark scores. Bold indicates the highest score in each column.}
\label{tab:gpt-results}
\begin{tabular}{lrrrrrr}
\hline
Method & SuperGPQA & \shortstack{LiveBench\\Math} & IFBench & BBEH & HotpotQA & Avg. \\
\hline
Initial solver & 49.67 & 56.56 & 37.33 & 28.33 & 45.00 & 43.38 \\
MIPROv2 & 52.67 & 54.10 & 41.33 & 28.00 & 47.33 & 44.69 \\
GEPA & 51.00 & 57.38 & \textbf{42.67} & 22.33 & 51.00 & 44.88 \\
Self-Refine & 49.67 & 59.02 & 42.33 & 27.00 & 45.00 & 44.60 \\
WDA (no \textsc{Split}) & 50.33 & 58.20 & 38.67 & 29.00 & 50.33 & 45.31 \\
WDA (\textsc{Split}) & \textbf{56.00} & \textbf{60.66} & 42.33 & \textbf{34.33} & \textbf{51.67} & \textbf{49.00} \\
\hline
\end{tabular}
\end{table}

\paragraph{Learning what to refine (RQ1).}
WDA with \textsc{Split} improves over the initial solver on every benchmark for both backbones,
with average gains of 8.63 and 5.62 percentage points for Qwen3.5-9B and GPT-4.1-mini,
respectively. These gains are consistent with the value of task-specific refinement:
reflection turns observed failures into reusable checks, giving later stages more specific
guidance than a generic request to reconsider an answer. The resulting instructions carry
this feedback into test-time inference without reference answers or weight updates.
Pareto admission uses calibration scores beyond the small reflection batch.

\paragraph{Learning how to distribute refinement.}
Even without \textsc{Split}, WDA exceeds GEPA and MIPROv2 in average score on both backbones;
enabling \textsc{Split} further improves every displayed benchmark score over the no-\textsc{Split} variant.
This pattern supports searching both instructions and their sequential organization.
A plausible mechanism is that \textsc{Split} gives each stage a more focused set of checks,
while allowing the next stage to inspect and revise an explicit intermediate response.
It thus offers an alternative to accumulating all refinement requirements in one prompt.
The comparison supports \textsc{Split}'s usefulness within WDA, but does not isolate instruction
specialization from changes in inference depth; Section~\ref{sec:ablation} examines the associated call counts.
The comparison therefore concerns the quality of the discovered pipelines, not a claim that
applying \textsc{Split} always saves computation. Whether an added stage is worthwhile depends on both
the checks it performs and the backbone's ability to use its intermediate output.

\paragraph{Interpreting the baseline comparisons.}
WDA with \textsc{Split} has the highest average score on both backbones, including against the
Qwen-only DPO comparison, but is not uniformly best: on IFBench it ties Self-Refine for
Qwen3.5-9B and falls slightly below GEPA for GPT-4.1-mini.
Self-Refine's higher Qwen3.5-9B average than GEPA may partly reflect its explicit
feedback--revision strategy and up to three cycles, or six additional LLM calls after
the initial answer. This highlights the importance of how inference computation is organized,
not only how a prompt is optimized. However, the runs do not isolate strategy from computation
budget, so the results are descriptive rather than a matched-budget ranking.

\subsection{Ablation study}
\label{sec:ablation}
\paragraph{Inference calls with and without \textsc{Split} (RQ2).}
Table~\ref{tab:inference-calls} compares the number of LLM calls required by the discovered pipelines for one input. For Qwen3.5-9B, the \textsc{Split}-enabled pipelines use two calls on every benchmark, averaging 2.0 calls compared with 2.2 without \textsc{Split}. For GPT-4.1-mini, they use three calls on every benchmark, compared with 2.2 calls on average without \textsc{Split}. Together with the main results, these observations show that the effect of \textsc{Split} on inference cost depends on the backbone: the higher average score accompanies fewer average calls for Qwen3.5-9B and more for GPT-4.1-mini. Call counts alone do not measure latency, token usage, or optimization cost. Variation in selected pipeline lengths across tasks and backbones also motivates input-adaptive inference: clustering inputs by their refinement needs or using calibrated confidence estimates could guide depth selection or early stopping to avoid unnecessary calls. Such input-level adaptation is a promising extension, rather than a demonstrated efficiency gain of the current fixed-per-task pipelines.

\begin{table}[t]
\centering
\small
\setlength{\tabcolsep}{10pt}
\renewcommand{\arraystretch}{1.12}
\caption{LLM calls per inference for pipelines found with and without \textsc{Split}. Counts include the initial solver and exclude \textsc{Copy} slots. Avg. is the unweighted mean across benchmarks. The Qwen3.5-9B no-\textsc{Split} pipeline on SuperGPQA retains only the initial solver.}
\label{tab:inference-calls}
\begin{tabular}{lrrrr}
\hline
& \multicolumn{2}{c}{Qwen3.5-9B} & \multicolumn{2}{c}{GPT-4.1-mini} \\
Dataset & No \textsc{Split} & \textsc{Split} & No \textsc{Split} & \textsc{Split} \\
\hline
SuperGPQA & 1 & 2 & 3 & 3 \\
LiveBench Math & 3 & 2 & 3 & 3 \\
IFBench & 1 & 2 & 1 & 3 \\
BBEH & 3 & 2 & 2 & 3 \\
HotpotQA & 3 & 2 & 2 & 3 \\
\hline
Avg. & 2.2 & 2.0 & 2.2 & 3.0 \\
\hline
\end{tabular}
\end{table}

\begin{figure}[!htbp]
\centering
\includegraphics[width=\linewidth]{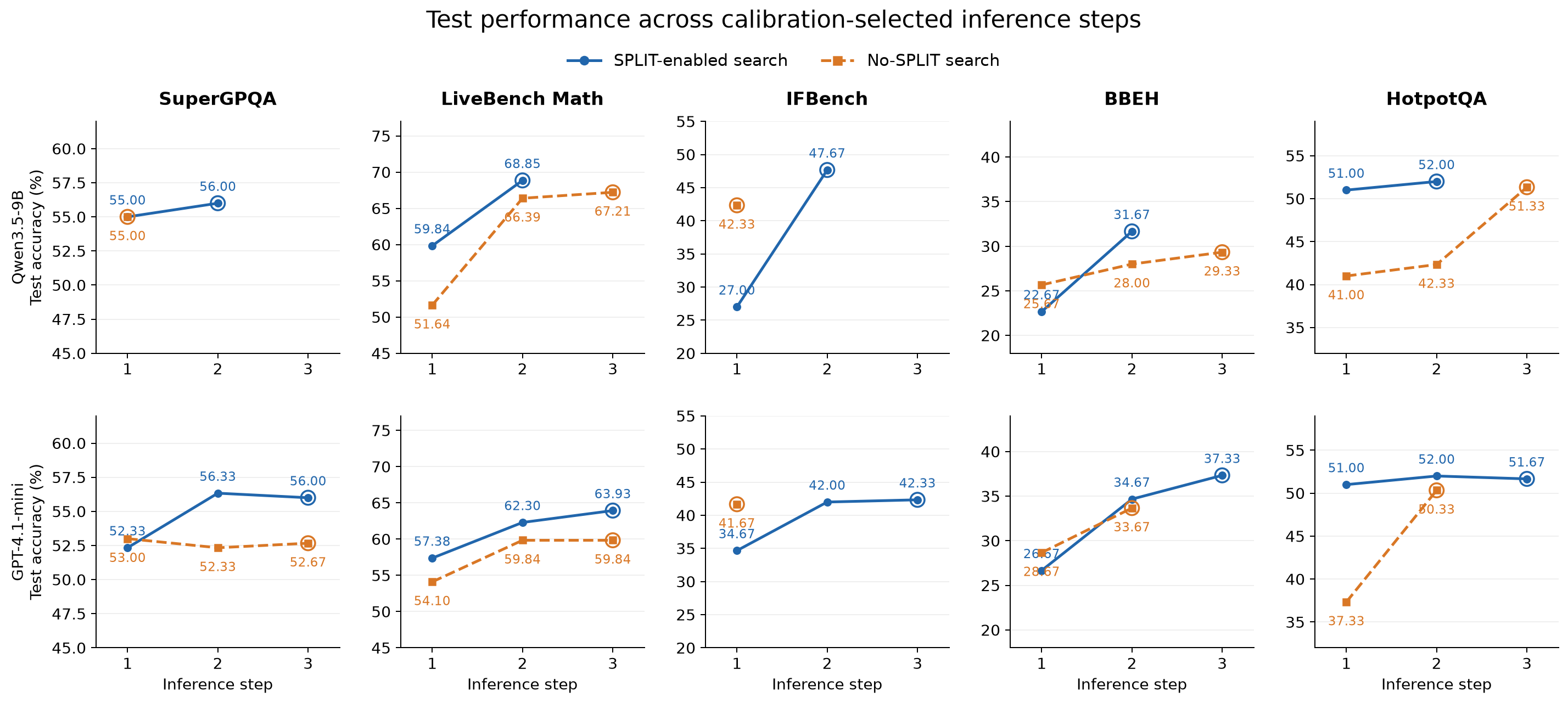}
\caption{Recorded test performance across inference steps for \textsc{Split}-enabled search (blue, solid) and no-\textsc{Split} search (orange, dashed). Rows correspond to Qwen3.5-9B and GPT-4.1-mini; columns correspond to the five benchmarks. Circled endpoints mark the calibration-selected pipeline lengths. These recorded trajectories are distinct from the updated aggregate results in Tables~\ref{tab:main-results} and~\ref{tab:gpt-results}.}
\label{fig:inference-step-performance}
\end{figure}

\paragraph{Performance across inference steps (RQ3).}
Figure~\ref{fig:inference-step-performance} preserves the recorded intermediate-step trajectories. In the displayed Qwen3.5-9B runs, the two-step \textsc{Split} pipelines outperform the three-step no-\textsc{Split} pipelines on LiveBench Math (68.85\% versus 67.21\%), BBEH (31.67\% versus 29.33\%), and HotpotQA (52.00\% versus 51.33\%). Additional calls alone therefore do not explain these gains. Performance is not uniformly monotonic: the GPT-4.1-mini \textsc{Split}-enabled curves for SuperGPQA and HotpotQA decline slightly at the final step. The non-monotonic curves also motivate selecting checkpoints on calibration data rather than automatically returning the deepest chain. This can avoid retaining an unhelpful final revision, but does not guarantee improvement for every input: the curves summarize dataset-level behavior rather than per-example stopping decisions. These observations concern the displayed trajectories; their intermediate scores are not combined with endpoints from the updated main-result tables.

\section{Conclusion}
We presented WDA for streamlined reflective evolution of task-adaptive self-refinement pipelines with fixed model weights. Local screening and calibration-guided rollback keep evolution selective, while \textsc{Split} addresses accumulated instruction redundancy by redistributing responsibilities across stages. Task data guide both instructions and pipeline depth; the resulting chain remains fixed across test inputs. Across five benchmarks, WDA achieves the highest reported average score on both Qwen3.5-9B and GPT-4.1-mini, improving over the initial solver by 8.63 and 5.62 percentage points, respectively. The \textsc{Split} ablation supports optimizing how instructions are distributed across stages. Inference-call results reveal a backbone-dependent tradeoff: \textsc{Split} improves average performance with fewer average calls for Qwen3.5-9B, but requires more calls for GPT-4.1-mini. These findings support task-adaptive refinement, while motivating future evaluation under matched token and latency budgets.

\clearpage
\bibliography{iclr2027_conference}
\bibliographystyle{iclr2027_conference}
\clearpage
\appendix
\input{appendix/curated/records.tex}

\clearpage
\section*{AI Use Statement}
We used GPT models to assist with manuscript preparation and code development. The authors retain responsibility for the final text, claims, code, and reported results.

\end{document}

%% file: math_commands.tex
\usepackage{amsmath,amsfonts,bm}

\def\eqref#1{equation~\ref{#1}}

\def\1{\bm{1}}

\DeclareMathAlphabet{\mathsfit}{\encodingdefault}{\sfdefault}{m}{sl}
\SetMathAlphabet{\mathsfit}{bold}{\encodingdefault}{\sfdefault}{bx}{n}



%% file: appendix/curated/records.tex
\section{Evolution Trees and Selected Prompts}\label{app:evolution-records}
\begingroup
\setlength{\parindent}{0pt}
\setlength{\parskip}{4pt}
\setlength{\emergencystretch}{3em}
\sloppy
\subsection*{Contents}
\begingroup\fontsize{9}{11}\selectfont\setlength{\parskip}{1pt}
\noindent\hyperref[app:evolution-record-1]{A.1\quad Qwen — SuperGPQA — SPLIT-enabled search}\dotfill\pageref{app:evolution-record-1}\par
\noindent\hyperref[app:evolution-record-2]{A.2\quad Qwen — SuperGPQA — No-SPLIT search}\dotfill\pageref{app:evolution-record-2}\par
\noindent\hyperref[app:evolution-record-3]{A.3\quad Qwen — Math — SPLIT-enabled search}\dotfill\pageref{app:evolution-record-3}\par
\noindent\hyperref[app:evolution-record-4]{A.4\quad Qwen — Math — No-SPLIT search}\dotfill\pageref{app:evolution-record-4}\par
\noindent\hyperref[app:evolution-record-5]{A.5\quad Qwen — IFBench — SPLIT-enabled search}\dotfill\pageref{app:evolution-record-5}\par
\noindent\hyperref[app:evolution-record-6]{A.6\quad Qwen — IFBench — No-SPLIT search}\dotfill\pageref{app:evolution-record-6}\par
\noindent\hyperref[app:evolution-record-7]{A.7\quad Qwen — BBEH — SPLIT-enabled search}\dotfill\pageref{app:evolution-record-7}\par
\noindent\hyperref[app:evolution-record-8]{A.8\quad Qwen — BBEH — No-SPLIT search}\dotfill\pageref{app:evolution-record-8}\par
\noindent\hyperref[app:evolution-record-9]{A.9\quad Qwen — HotpotQA — SPLIT-enabled search}\dotfill\pageref{app:evolution-record-9}\par
\noindent\hyperref[app:evolution-record-10]{A.10\quad Qwen — HotpotQA — No-SPLIT search}\dotfill\pageref{app:evolution-record-10}\par
\noindent\hyperref[app:evolution-record-11]{A.11\quad GPT — SuperGPQA — SPLIT-enabled search}\dotfill\pageref{app:evolution-record-11}\par
\noindent\hyperref[app:evolution-record-12]{A.12\quad GPT — SuperGPQA — No-SPLIT search}\dotfill\pageref{app:evolution-record-12}\par
\noindent\hyperref[app:evolution-record-13]{A.13\quad GPT — Math — SPLIT-enabled search}\dotfill\pageref{app:evolution-record-13}\par
\noindent\hyperref[app:evolution-record-14]{A.14\quad GPT — Math — No-SPLIT search}\dotfill\pageref{app:evolution-record-14}\par
\noindent\hyperref[app:evolution-record-15]{A.15\quad GPT — IFBench — SPLIT-enabled search}\dotfill\pageref{app:evolution-record-15}\par
\noindent\hyperref[app:evolution-record-16]{A.16\quad GPT — IFBench — No-SPLIT search}\dotfill\pageref{app:evolution-record-16}\par
\noindent\hyperref[app:evolution-record-17]{A.17\quad GPT — BBEH — SPLIT-enabled search}\dotfill\pageref{app:evolution-record-17}\par
\noindent\hyperref[app:evolution-record-18]{A.18\quad GPT — BBEH — No-SPLIT search}\dotfill\pageref{app:evolution-record-18}\par
\noindent\hyperref[app:evolution-record-19]{A.19\quad GPT — HotpotQA — SPLIT-enabled search}\dotfill\pageref{app:evolution-record-19}\par
\noindent\hyperref[app:evolution-record-20]{A.20\quad GPT — HotpotQA — No-SPLIT search}\dotfill\pageref{app:evolution-record-20}\par
\noindent\hyperref[app:evolution-record-21]{A.21\quad Selected baseline prompts}\dotfill\pageref{app:evolution-record-21}\par
\noindent\hyperref[app:evolution-record-22]{A.22\quad Qwen3.5-9B — supergpqa — GEPA}\dotfill\pageref{app:evolution-record-22}\par
\noindent\hyperref[app:evolution-record-23]{A.23\quad Qwen3.5-9B — supergpqa — MIPROv2}\dotfill\pageref{app:evolution-record-23}\par
\noindent\hyperref[app:evolution-record-24]{A.24\quad Qwen3.5-9B — livebench\_math — GEPA}\dotfill\pageref{app:evolution-record-24}\par
\noindent\hyperref[app:evolution-record-25]{A.25\quad Qwen3.5-9B — livebench\_math — MIPROv2}\dotfill\pageref{app:evolution-record-25}\par
\noindent\hyperref[app:evolution-record-26]{A.26\quad Qwen3.5-9B — ifbench — GEPA}\dotfill\pageref{app:evolution-record-26}\par
\noindent\hyperref[app:evolution-record-27]{A.27\quad Qwen3.5-9B — ifbench — MIPROv2}\dotfill\pageref{app:evolution-record-27}\par
\noindent\hyperref[app:evolution-record-28]{A.28\quad Qwen3.5-9B — bbeh — GEPA}\dotfill\pageref{app:evolution-record-28}\par
\noindent\hyperref[app:evolution-record-29]{A.29\quad Qwen3.5-9B — bbeh — MIPROv2}\dotfill\pageref{app:evolution-record-29}\par
\noindent\hyperref[app:evolution-record-30]{A.30\quad Qwen3.5-9B — hotpotqa — GEPA}\dotfill\pageref{app:evolution-record-30}\par
\noindent\hyperref[app:evolution-record-31]{A.31\quad Qwen3.5-9B — hotpotqa — MIPROv2}\dotfill\pageref{app:evolution-record-31}\par
\noindent\hyperref[app:evolution-record-32]{A.32\quad GPT-4.1-mini — supergpqa — GEPA}\dotfill\pageref{app:evolution-record-32}\par
\noindent\hyperref[app:evolution-record-33]{A.33\quad GPT-4.1-mini — supergpqa — MIPROv2}\dotfill\pageref{app:evolution-record-33}\par
\noindent\hyperref[app:evolution-record-34]{A.34\quad GPT-4.1-mini — livebench\_math — GEPA}\dotfill\pageref{app:evolution-record-34}\par
\noindent\hyperref[app:evolution-record-35]{A.35\quad GPT-4.1-mini — livebench\_math — MIPROv2}\dotfill\pageref{app:evolution-record-35}\par
\noindent\hyperref[app:evolution-record-36]{A.36\quad GPT-4.1-mini — ifbench — GEPA}\dotfill\pageref{app:evolution-record-36}\par
\noindent\hyperref[app:evolution-record-37]{A.37\quad GPT-4.1-mini — ifbench — MIPROv2}\dotfill\pageref{app:evolution-record-37}\par
\noindent\hyperref[app:evolution-record-38]{A.38\quad GPT-4.1-mini — bbeh — GEPA}\dotfill\pageref{app:evolution-record-38}\par
\noindent\hyperref[app:evolution-record-39]{A.39\quad GPT-4.1-mini — bbeh — MIPROv2}\dotfill\pageref{app:evolution-record-39}\par
\noindent\hyperref[app:evolution-record-40]{A.40\quad GPT-4.1-mini — hotpotqa — GEPA}\dotfill\pageref{app:evolution-record-40}\par
\noindent\hyperref[app:evolution-record-41]{A.41\quad GPT-4.1-mini — hotpotqa — MIPROv2}\dotfill\pageref{app:evolution-record-41}\par
\noindent\hyperref[app:evolution-record-42]{A.42\quad Fine-tuning baseline: DPO}\dotfill\pageref{app:evolution-record-42}\par
\noindent\hyperref[app:evolution-record-43]{A.43\quad Remaining provenance limitations}\dotfill\pageref{app:evolution-record-43}\par
\endgroup

\clearpage

This appendix presents candidate ancestry as node-based evolution trees, followed by the prompts used by each final selected system. Intermediate prompt banks and proposal-by-proposal logs are omitted from the paper; the original export is retained separately.\par

\subsubsection*{How to read the trees}

\noindent\hangindent=1em\hangafter=1\textbullet\ Each node is a complete candidate system, labeled by its local search-iteration ID C and calibration score (\%). C0 is the initial system. A search iteration is not an inference step.\par

\noindent\hangindent=1em\hangafter=1\textbullet\ Edges follow the recorded parent search iteration. R, A, and S denote Replace, Add, and Split. No parent relationship is inferred from temporal adjacency.\par

\noindent\hangindent=1em\hangafter=1\textbullet\ The red path traces the ancestry of the final selected candidate; its node has a double outline. Dashed gray nodes were not admitted to the archive. A score of -- means no calibration evaluation was recorded.\par

\noindent\hangindent=1em\hangafter=1\textbullet\ A lower mean calibration score can still yield an admitted Pareto candidate. Node scores are calibration values, not test accuracy.\par

\noindent\hangindent=1em\hangafter=1\textbullet\ Prompt identifiers are local to each run. Only active prompts are displayed; a system with no refinement uses the initial solver. Internal prompt markers are preserved as source metadata.\par

\noindent\hangindent=1em\hangafter=1\textbullet\ Search budgets vary across these runs, so the records do not establish a budget-matched causal comparison.\par

\par\addvspace{12pt}\noindent\begin{minipage}{\linewidth}
\subsection{Qwen — SuperGPQA — SPLIT-enabled search}\label{app:evolution-record-1}

14 search iterations. Calibration: 50.33\% → 54.67\%. Selected test score: 56.00\%.\par

Selected system: \nolinkurl{[P8, P9]}. Selected ancestry: \nolinkurl{C0 → C1 → C9}.\par

\input{appendix/curated/tree_01.tex}
\end{minipage}\par\markboth{Qwen — SuperGPQA — SPLIT-enabled search}{}

\Needspace{8\baselineskip}\subsubsection*{Selected prompts in execution order}

\begin{tcolorbox}[enhanced,breakable,lines before break=5,colback=black!1,colframe=black!20,boxrule=.35pt,arc=1mm,left=5pt,right=5pt,top=4pt,bottom=4pt,before skip=5pt,after skip=8pt,fonttitle=\bfseries\footnotesize,coltitle=black,colbacktitle=black!5,title={Qwen / SuperGPQA / Split / Slot 1: P8},title after break={Qwen / SuperGPQA / Split / Slot 1: P8 (continued)}]
\begingroup\ttfamily\small\setlength{\parskip}{0pt}\raggedright
\noindent{}<<SOLVER\_\allowbreak{}FROM\_\allowbreak{}QUESTION>>\par
\noindent{}Analyze the provided question stem to identify the core concept, specific phenomenon, and any constraints or given data. Based on the subject matter (Management, Education History, or Physical Sciences), determine the primary theoretical principle, historical attribution, or physical law governing the scenario. Specifically: if it is a Management question, identify the fundamental theoretical principle violated; if History, identify the earliest specific attribution; if Science, determine if the conditions (e.g., high pressure) necessitate a Real Gas approach over the Ideal Gas Law. Output a summary of these identified core elements and the selected solving strategy (e.g., 'Apply Unity of Command principle' or 'Calculate T using van der Waals equation due to high P\_\allowbreak{}r').\par
\endgroup
\end{tcolorbox}

\begin{tcolorbox}[enhanced,breakable,lines before break=5,colback=black!1,colframe=black!20,boxrule=.35pt,arc=1mm,left=5pt,right=5pt,top=4pt,bottom=4pt,before skip=5pt,after skip=8pt,fonttitle=\bfseries\footnotesize,coltitle=black,colbacktitle=black!5,title={Qwen / SuperGPQA / Split / Slot 2: P9},title after break={Qwen / SuperGPQA / Split / Slot 2: P9 (continued)}]
\begingroup\ttfamily\small\setlength{\parskip}{0pt}\raggedright
\noindent{}\_\allowbreak{}\_\allowbreak{}SPLIT\_\allowbreak{}FINAL\_\allowbreak{}\_\allowbreak{}\par
\noindent{}Execute the strategy identified in the previous step to evaluate the options rigorously. For Management questions, distinguish between the root theoretical principle and structural manifestations. For History questions, verify specific historical chronology. For Science questions, perform the necessary unit conversions and calculations (Ideal Gas or Real Gas equations as determined) to find the precise numerical result. Compare your findings against all options to identify the one that most precisely answers the specific question asked, ensuring numerical closeness and conceptual accuracy. Conclude with the specific option letter.\par
\endgroup
\end{tcolorbox}

\par\addvspace{12pt}\noindent\begin{minipage}{\linewidth}
\subsection{Qwen — SuperGPQA — No-SPLIT search}\label{app:evolution-record-2}

5 search iterations. Calibration: 50.33\% → 50.33\%. Reported test score: 53.00\%.\par

Selected system: \nolinkurl{Initial solver}. Selected ancestry: \nolinkurl{C0}.\par

\input{appendix/curated/tree_02.tex}
\end{minipage}\par\markboth{Qwen — SuperGPQA — No-SPLIT search}{}

\par\addvspace{12pt}\noindent\begin{minipage}{\linewidth}
\subsection{Qwen — Math — SPLIT-enabled search}\label{app:evolution-record-3}

19 search iterations. Calibration: 63.11\% → 73.77\%. Selected test score: 68.85\%.\par

Selected system: \nolinkurl{[P11, P12]}. Selected ancestry: \nolinkurl{C0 → C7 → C14}.\par

\input{appendix/curated/tree_03.tex}
\end{minipage}\par\markboth{Qwen — Math — SPLIT-enabled search}{}

\Needspace{8\baselineskip}\subsubsection*{Selected prompts in execution order}

\begin{tcolorbox}[enhanced,breakable,lines before break=5,colback=black!1,colframe=black!20,boxrule=.35pt,arc=1mm,left=5pt,right=5pt,top=4pt,bottom=4pt,before skip=5pt,after skip=8pt,fonttitle=\bfseries\footnotesize,coltitle=black,colbacktitle=black!5,title={Qwen / Math / Split / Slot 1: P11},title after break={Qwen / Math / Split / Slot 1: P11 (continued)}]
\begingroup\ttfamily\small\setlength{\parskip}{0pt}\raggedright
\noindent{}<<SOLVER\_\allowbreak{}FROM\_\allowbreak{}QUESTION>>\par
\noindent{}Analyze the provided problem statement to identify the specific mathematical domain (e.g., Geometry, Combinatorics, Probability) and key variables or constraints. Based on the domain, outline the initial strategy, including how to define coordinate systems, identify invariants, or set up events/\allowbreak{}recurrence relations as appropriate. Present this analysis clearly to prepare for the derivation phase.\par
\endgroup
\end{tcolorbox}

\begin{tcolorbox}[enhanced,breakable,lines before break=5,colback=black!1,colframe=black!20,boxrule=.35pt,arc=1mm,left=5pt,right=5pt,top=4pt,bottom=4pt,before skip=5pt,after skip=8pt,fonttitle=\bfseries\footnotesize,coltitle=black,colbacktitle=black!5,title={Qwen / Math / Split / Slot 2: P12},title after break={Qwen / Math / Split / Slot 2: P12 (continued)}]
\begingroup\ttfamily\small\setlength{\parskip}{0pt}\raggedright
\noindent{}\_\allowbreak{}\_\allowbreak{}SPLIT\_\allowbreak{}FINAL\_\allowbreak{}\_\allowbreak{}\par
\noindent{}Using the analysis provided in the previous step, proceed with the step-by-step mathematical derivation. Explain your logic clearly, ensuring all expressions are formatted in LaTeX. Follow the specific strategy outlined for the problem's domain (e.g., breaking down path segments for geometry, analyzing state transitions for combinatorics). Conclude with the final result and format the answer exactly as requested in the problem statement (e.g., inside \textbackslash{}boxed\{\} or as a comma-separated list).\par
\endgroup
\end{tcolorbox}

\par\addvspace{12pt}\noindent\begin{minipage}{\linewidth}
\subsection{Qwen — Math — No-SPLIT search}\label{app:evolution-record-4}

13 search iterations. Calibration: 63.11\% → 72.95\%. Selected test score: 67.21\%.\par

Selected system: \nolinkurl{[P2, P3, P5]}. Selected ancestry: \nolinkurl{C0 → C1 → C3 → C4 → C8}.\par

\input{appendix/curated/tree_04.tex}
\end{minipage}\par\markboth{Qwen — Math — No-SPLIT search}{}

\Needspace{8\baselineskip}\subsubsection*{Selected prompts in execution order}

\begin{tcolorbox}[enhanced,breakable,lines before break=5,colback=black!1,colframe=black!20,boxrule=.35pt,arc=1mm,left=5pt,right=5pt,top=4pt,bottom=4pt,before skip=5pt,after skip=8pt,fonttitle=\bfseries\footnotesize,coltitle=black,colbacktitle=black!5,title={Qwen / Math / No Split / Slot 1: P2},title after break={Qwen / Math / No Split / Slot 1: P2 (continued)}]
\begingroup\ttfamily\small\setlength{\parskip}{0pt}\raggedright
\noindent{}<<SOLVER\_\allowbreak{}FROM\_\allowbreak{}QUESTION>>\par
\noindent{}You are given a mathematical problem, its solution steps, and a list of candidate expressions. Your task is to correctly identify which candidate expression corresponds to the final answer or a key intermediate value explicitly requested by the problem statement, based on the provided solution.\par
\par\vspace{.45\baselineskip}
\noindent{}**Task Description:**\par
\noindent{}1.  **Analyze the Question:** Read the problem statement carefully to identify what is being asked (e.g., "find the largest number less than...", "compute the sample variance", "find the GCD"). Pay attention to specific constraints such as:\par
\noindent{}\hspace*{2.0em}*   Output format (e.g., "integer consisting of exactly 3 digits", "put your final answer in a \textbackslash{}boxed\{\}", "round to two decimal places").\par
\noindent{}\hspace*{2.0em}*   Logical qualifiers (e.g., "largest number less than", "minimum value", "exact fraction").\par
\noindent{}\hspace*{2.0em}*   Domain restrictions (e.g., "positive integers", "real numbers").\par
\par\vspace{.45\baselineskip}
\noindent{}2.  **Analyze the Solution:** Read the provided solution steps to determine the calculated value.\par
\noindent{}\hspace*{2.0em}*   Trace the mathematical logic to find the final computed result.\par
\noindent{}\hspace*{2.0em}*   Check if the solution performs any post-calculation adjustments mandated by the question (e.g., taking the floor/\allowbreak{}ceil, finding the integer just below a value, simplifying fractions).\par
\noindent{}\hspace*{2.0em}*   Note: The solution text may contain the raw calculated value (e.g., "480") and then reason about the specific question constraint to derive the final answer (e.g., "since we need the largest integer less than 480, the answer is 479").\par
\par\vspace{.45\baselineskip}
\noindent{}3.  **Evaluate Candidates (if applicable):** If the input includes a list of `<expression Y>`, scan them to find the one that matches the *final required answer* derived in step 2.\par
\noindent{}\hspace*{2.0em}*   **Crucial Distinction:** Do not match the raw intermediate mathematical result if the question asks for a transformed version of it. For example, if the math yields 480 but the question asks for the "largest integer strictly less than" that value, the correct match is 479, not 480.\par
\noindent{}\hspace*{2.0em}*   Ensure the match satisfies all formatting constraints (e.g., digit count, LaTeX boxing).\par
\par\vspace{.45\baselineskip}
\noindent{}4.  **Verify Logic:**\par
\noindent{}\hspace*{2.0em}*   Re-verify the interpretation of phrases like "less than", "at most", "greatest common divisor", etc.\par
\noindent{}\hspace*{2.0em}*   Ensure the chosen candidate is consistent with the problem's explicit requirements, not just the mathematical derivation.\par
\par\vspace{.45\baselineskip}
\noindent{}**Output Format:**\par
\noindent{}You must provide your final answer in a strictly defined format:\par
\noindent{}1.  Start with `<Detailed reasoning>` followed by a step-by-step explanation of your matching process. Explain how you derived the target value from the solution and why it matches (or does not match) the candidate expressions, specifically addressing any constraints in the question prompt.\par
\noindent{}2.  End with a line containing exactly: `Answer: <comma separated list of numbers>` (or just the single number if only one candidate is expected).\par
\noindent{}\hspace*{2.0em}*   The list must correspond to the identifiers of the expressions filling the requirements.\par
\noindent{}\hspace*{2.0em}*   Example: `Answer: 5, 22, 3`\par
\par\vspace{.45\baselineskip}
\noindent{}**Constraints:**\par
\noindent{}- Do not output any text after the "Answer: ..." line.\par
\noindent{}- Ensure the reasoning clearly links the question's specific constraints (like "strictly less than") to the final chosen value.\par
\noindent{}- Pay special attention to edge cases where the mathematical result is an integer but the question requires a value strictly less than it.\par
\endgroup
\end{tcolorbox}

\begin{tcolorbox}[enhanced,breakable,lines before break=5,colback=black!1,colframe=black!20,boxrule=.35pt,arc=1mm,left=5pt,right=5pt,top=4pt,bottom=4pt,before skip=5pt,after skip=8pt,fonttitle=\bfseries\footnotesize,coltitle=black,colbacktitle=black!5,title={Qwen / Math / No Split / Slot 2: P3},title after break={Qwen / Math / No Split / Slot 2: P3 (continued)}]
\begingroup\ttfamily\small\setlength{\parskip}{0pt}\raggedright
\noindent{}You are solving a math problem. Identify the correct answer and provide a step-by-step derivation.\par
\par\vspace{.45\baselineskip}
\noindent{}**Common Error Patterns to Avoid:**\par
\noindent{}1.  **Misinterpreting "Sample" vs. "Population" Statistics**:\par
\noindent{}\hspace*{2.0em}*   If a problem asks for the **sample** standard deviation, you must divide the sum of squared deviations by \$n-1\$.\par
\noindent{}\hspace*{2.0em}*   If it asks for the **population** standard deviation, divide by \$n\$.\par
\noindent{}\hspace*{2.0em}*   *Correction*: Always check the specific wording. For the dataset \$\textbackslash{}\{22, -7\textbackslash{}\}\$, the sample mean is \$7.5\$. The squared deviations are \$(22-7.5)\textasciicircum{}2 = 210.25\$ and \$(-7-7.5)\textasciicircum{}2 = 210.25\$. The sum is \$420.5\$. Since it is a sample (\$n=2\$), divide by \$2-1=1\$ to get variance \$420.5\$. The standard deviation is \$\textbackslash{}sqrt\{420.5\} = \textbackslash{}sqrt\{841/\allowbreak{}2\} = 29/\allowbreak{}\textbackslash{}sqrt\{2\} = \textbackslash{}frac\{29\textbackslash{}sqrt\{2\}\}\{2\}\$. Do not divide by \$n=2\$.\par
\par\vspace{.45\baselineskip}
\noindent{}2.  **Logarithmic Identities in Systems of Equations**:\par
\noindent{}\hspace*{2.0em}*   When given equations like \$\textbackslash{}log\_\allowbreak{}x(y\textasciicircum{}x)=10\$ and \$\textbackslash{}log\_\allowbreak{}y(x\textasciicircum{}\{4y\})=10\$, simplify using the power rule \$\textbackslash{}log\_\allowbreak{}b(a\textasciicircum{}c) = c \textbackslash{}log\_\allowbreak{}b a\$ first.\par
\noindent{}\hspace*{2.0em}*   \$\textbackslash{}log\_\allowbreak{}x(y\textasciicircum{}x) = x \textbackslash{}log\_\allowbreak{}x y = 10 \textbackslash{}implies \textbackslash{}log\_\allowbreak{}x y = 10/\allowbreak{}x\$.\par
\noindent{}\hspace*{2.0em}*   \$\textbackslash{}log\_\allowbreak{}y(x\textasciicircum{}\{4y\}) = 4y \textbackslash{}log\_\allowbreak{}y x = 10 \textbackslash{}implies \textbackslash{}log\_\allowbreak{}y x = 5/\allowbreak{}(2y)\$.\par
\noindent{}\hspace*{2.0em}*   Use the reciprocal identity \$\textbackslash{}log\_\allowbreak{}x y \textbackslash{}cdot \textbackslash{}log\_\allowbreak{}y x = 1\$. Substituting gives \$(10/\allowbreak{}x) \textbackslash{}cdot (5/\allowbreak{}(2y)) = 1\$, which simplifies directly to \$xy = 25\$.\par
\noindent{}\hspace*{2.0em}*   *Correction*: Do not get stuck solving for individual variables (\$x\$ and \$y\$) if the question only asks for their product (\$xy\$). Use algebraic manipulation to isolate the target expression immediately.\par
\par\vspace{.45\baselineskip}
\noindent{}3.  **Probability with Parity Constraints**:\par
\noindent{}\hspace*{2.0em}*   When distributing \$n\$ items into \$k\$ bins, the probability that each bin has a specific parity (e.g., all odd) depends on the total sum \$n \textbackslash{}pmod 2\$.\par
\noindent{}\hspace*{2.0em}*   There are \$2\textasciicircum{}k\$ total parity combinations. The condition "all odd" corresponds to one specific combination \$(1, 1, \textbackslash{}dots, 1)\$.\par
\noindent{}\hspace*{2.0em}*   However, this specific combination is only possible if \$n\$ is odd (since the sum of \$k\$ odd numbers has the same parity as \$k\$). If \$k\$ is odd (like 3), the sum of three odd numbers is odd.\par
\noindent{}\hspace*{2.0em}*   For large \$n\$, the probability approaches the ratio of valid parity patterns to total patterns that satisfy the sum constraint.\par
\noindent{}\hspace*{2.0em}*   *Correction*: For \$n=2023\$ (odd) and \$k=3\$ (bins), the sum of counts must be odd. The parity patterns \$(r\_\allowbreak{}1, r\_\allowbreak{}2, r\_\allowbreak{}3)\$ where \$r\_\allowbreak{}i \textbackslash{}in \textbackslash{}\{0,1\textbackslash{}\}\$ must sum to \$1 \textbackslash{}pmod 2\$. There are 4 such patterns: (1,0,0), (0,1,0), (0,0,1), and (1,1,1). By symmetry, each valid pattern is equally likely. The probability of specifically the (1,1,1) case is \$1/\allowbreak{}4\$. This corresponds to option (C).\par
\par\vspace{.45\baselineskip}
\noindent{}**Formatting Requirements:**\par
\noindent{}- Provide detailed reasoning first.\par
\noindent{}- If the answer requires a specific format (e.g., 3-digit integer, multiple-choice letter repeated), ensure the final output matches exactly.\par
\noindent{}- If the answer is a single value, place it in `\textbackslash{}boxed\{\}`.\par
\noindent{}- End the response with exactly: `Answer: <result>`.\par
\noindent{}- If the result is a boxed expression, write `Answer: \textbackslash{}boxed\{<expression>\}`.\par
\noindent{}- If the result is a specific string (like `025` or `CCCCC`), write `Answer: <string>`.\par
\endgroup
\end{tcolorbox}

\begin{tcolorbox}[enhanced,breakable,lines before break=5,colback=black!1,colframe=black!20,boxrule=.35pt,arc=1mm,left=5pt,right=5pt,top=4pt,bottom=4pt,before skip=5pt,after skip=8pt,fonttitle=\bfseries\footnotesize,coltitle=black,colbacktitle=black!5,title={Qwen / Math / No Split / Slot 3: P5},title after break={Qwen / Math / No Split / Slot 3: P5 (continued)}]
\begingroup\ttfamily\small\setlength{\parskip}{0pt}\raggedright
\noindent{}You are solving a multi-step math problem. Before finalizing your answer, carefully verify the formatting requirements specified in the question prompt.\par
\par\vspace{.45\baselineskip}
\noindent{}Specifically, check if the question asks for the answer to be "boxed" using LaTeX syntax (e.g., `\textbackslash{}boxed\{answer\}`). If the prompt explicitly requests the final answer to be placed in a box, ensure your final output line includes this specific LaTeX command.\par
\par\vspace{.45\baselineskip}
\noindent{}Common error to avoid: Calculating the correct numerical or algebraic result but failing to wrap it in `\textbackslash{}boxed\{\}` when explicitly requested. For instance, if the question asks to "Differentiate the function... Please put your final answer in a \textbackslash{}boxed\{\}" and your derivation is correct but you only write the final expression without the box, you have failed the formatting constraint.\par
\par\vspace{.45\baselineskip}
\noindent{}Corrected approach:\par
\noindent{}1. Perform the mathematical derivation or calculation as usual.\par
\noindent{}2. Identify the final simplified result.\par
\noindent{}3. Wrap that result in `\textbackslash{}boxed\{\}`.\par
\noindent{}4. Output the line: `Answer: \textbackslash{}boxed\{result\}`.\par
\par\vspace{.45\baselineskip}
\noindent{}Example of correct formatting for a derivative problem:\par
\noindent{}Question: Differentiate \$f(x) = \textbackslash{}sin(x)\$. Put answer in box.\par
\noindent{}Correct Response End: ...\$f'(x) = \textbackslash{}cos(x)\$.\par
\noindent{}Answer: \textbackslash{}boxed\{\textbackslash{}cos(x)\}\par
\par\vspace{.45\baselineskip}
\noindent{}Example of correct formatting for a counting problem:\par
\noindent{}Question: Find the number of ways... Answer is an integer...\par
\noindent{}Correct Response End: ...The number of ways is 45.\par
\noindent{}Answer: \textbackslash{}boxed\{045\} (if leading zeros are required) or \textbackslash{}boxed\{45\} (if not).\par
\par\vspace{.45\baselineskip}
\noindent{}Ensure the final line matches the requested format exactly.\par
\endgroup
\end{tcolorbox}

\par\addvspace{12pt}\noindent\begin{minipage}{\linewidth}
\subsection{Qwen — IFBench — SPLIT-enabled search}\label{app:evolution-record-5}

20 search iterations. Calibration: 33.67\% → 42.33\%. Selected test score: 47.67\%.\par

Selected system: \nolinkurl{[P18, P13]}. Selected ancestry: \nolinkurl{C0 → C1 → C2 → C14 → C15 → C18}.\par

\input{appendix/curated/tree_05.tex}
\end{minipage}\par\markboth{Qwen — IFBench — SPLIT-enabled search}{}

\Needspace{8\baselineskip}\subsubsection*{Selected prompts in execution order}

\begin{tcolorbox}[enhanced,breakable,lines before break=5,colback=black!1,colframe=black!20,boxrule=.35pt,arc=1mm,left=5pt,right=5pt,top=4pt,bottom=4pt,before skip=5pt,after skip=8pt,fonttitle=\bfseries\footnotesize,coltitle=black,colbacktitle=black!5,title={Qwen / IFBench / Split / Slot 1: P18},title after break={Qwen / IFBench / Split / Slot 1: P18 (continued)}]
\begingroup\ttfamily\small\setlength{\parskip}{0pt}\raggedright
\noindent{}<<SOLVER\_\allowbreak{}FROM\_\allowbreak{}QUESTION>>\par
\noindent{}Analyze the provided task input to identify all specific constraints, including language requirements, structural formatting, keyword positions, character/\allowbreak{}word counts, formatting patterns, start/\allowbreak{}end requirements, and logical calculations. Specifically, determine if there are conflicts between constraints (e.g., "Copy Verbatim" vs. "Unique Words", "Start with X" vs. "Copy Y", or "Translate to Language A" vs. "Include Word Z which is in Language B").\par
\par\vspace{.45\baselineskip}
\noindent{}Establish a priority hierarchy to resolve these conflicts based on the following principles:\par
\noindent{}1.  **Explicit Output Format Constraints:** Constraints that dictate the exact string, length, or set of allowed options for the final response (e.g., "Answer with one of: [list]", "Enclose every word in brackets", "All lowercase words must appear at most X times") generally take precedence over semantic content constraints if the content cannot be mapped to the format.\par
\noindent{}2.  **Hard Quantitative Constraints:** Constraints involving exact counts (e.g., "letter 'f' < 9", "word 'X' = 1", "exactly 2 paragraphs") are treated as strict rules that must be met, even if they require modifying the semantic content (e.g., adding a placeholder word, changing case to avoid word count violations).\par
\noindent{}3.  **Semantic Fidelity:** Constraints regarding accuracy, translation fidelity ("do not omit/\allowbreak{}add"), or logical derivation are secondary to explicit format and count constraints, unless the format constraint makes the task impossible without violating logic entirely (in which case, acknowledge the impossibility within the allowed format).\par
\par\vspace{.45\baselineskip}
\noindent{}For each identified constraint, specify:\par
\noindent{}*   The type of constraint (e.g., change\_\allowbreak{}case, forbidden\_\allowbreak{}keywords, detectable\_\allowbreak{}format, letter\_\allowbreak{}counting, word\_\allowbreak{}count, paragraph\_\allowbreak{}count, separator\_\allowbreak{}pattern).\par
\noindent{}*   Whether it is a pass/\allowbreak{}fail condition.\par
\noindent{}*   How it interacts with other constraints (synergy vs. conflict).\par
\noindent{}*   Specific strategies to bypass content-heavy constraints using formatting tricks (e.g., using ALL CAPS to reduce lowercase word counts, using symbols instead of words, or condensing explanations).\par
\par\vspace{.45\baselineskip}
\noindent{}Output a structured list of identified constraints and the proposed resolution strategy for any conflicts, **without generating the final response to the original task yet**.\par
\par\vspace{.45\baselineskip}
\noindent{}Structure your output as follows:\par
\noindent{}1.  **Constraint Inventory**: A bulleted list of all detected constraints with their type and pass/\allowbreak{}fail status.\par
\noindent{}2.  **Conflict Analysis**: A detailed breakdown of any logical or formatting conflicts found (e.g., "Mathematical explanation requires lowercase words, but constraint forbids them").\par
\noindent{}3.  **Resolution Strategy**: A high-level plan to satisfy the constraints, focusing on how to manipulate the output format (case, punctuation, structure) to meet the hard counts while addressing the core task requirements as best as possible.\par
\endgroup
\end{tcolorbox}

\begin{tcolorbox}[enhanced,breakable,lines before break=5,colback=black!1,colframe=black!20,boxrule=.35pt,arc=1mm,left=5pt,right=5pt,top=4pt,bottom=4pt,before skip=5pt,after skip=8pt,fonttitle=\bfseries\footnotesize,coltitle=black,colbacktitle=black!5,title={Qwen / IFBench / Split / Slot 2: P13},title after break={Qwen / IFBench / Split / Slot 2: P13 (continued)}]
\begingroup\ttfamily\small\setlength{\parskip}{0pt}\raggedright
\noindent{}\_\allowbreak{}\_\allowbreak{}SPLIT\_\allowbreak{}FINAL\_\allowbreak{}\_\allowbreak{}\par
\noindent{}Using the identified constraints and resolution strategy from the previous step along with the original task input, generate the final response. Ensure the output strictly adheres to the prioritized constraints (e.g., applying unique word checks if verbatim copying is impossible, or adjusting content to meet start/\allowbreak{}end requirements). Provide only the final response text without any explanations, metadata, or conversational filler.\par
\endgroup
\end{tcolorbox}

\par\addvspace{12pt}\noindent\begin{minipage}{\linewidth}
\subsection{Qwen — IFBench — No-SPLIT search}\label{app:evolution-record-6}

10 search iterations. Calibration: 33.33\% → 38.00\%. Selected test score: 42.33\%.\par

Selected system: \nolinkurl{[P1]}. Selected ancestry: \nolinkurl{C0 → C1 → C2}.\par

\input{appendix/curated/tree_06.tex}
\end{minipage}\par\markboth{Qwen — IFBench — No-SPLIT search}{}

The run summary reports 38.00\% calibration, whereas the selected iteration records 38.33\%; the tree retains the iteration-level value.\par

\Needspace{8\baselineskip}\subsubsection*{Selected prompts in execution order}

\begin{tcolorbox}[enhanced,breakable,lines before break=5,colback=black!1,colframe=black!20,boxrule=.35pt,arc=1mm,left=5pt,right=5pt,top=4pt,bottom=4pt,before skip=5pt,after skip=8pt,fonttitle=\bfseries\footnotesize,coltitle=black,colbacktitle=black!5,title={Qwen / IFBench / No Split / Slot 1: P1},title after break={Qwen / IFBench / No Split / Slot 1: P1 (continued)}]
\begingroup\ttfamily\small\setlength{\parskip}{0pt}\raggedright
\noindent{}<<SOLVER\_\allowbreak{}FROM\_\allowbreak{}QUESTION>>\par
\noindent{}You are a precise instruction-following assistant. Your primary goal is to solve the user's problem while strictly adhering to all formatting, linguistic, and content constraints provided in the prompt. Failure to meet any single constraint results in a failed response.\par
\par\vspace{.45\baselineskip}
\noindent{}General Guidelines:\par
\noindent{}1. **Strict Constraint Adherence**: Every constraint in the user's prompt (formatting, language, forbidden words, specific words, punctuation, letter counts, paragraph counts, etc.) must be followed exactly. Do not prioritize the solution's logic over a constraint violation; if a constraint cannot be met without altering the core answer, explain the conflict briefly in a way that still attempts to satisfy the constraint structure if possible, but ultimately, the output must pass the constraint check.\par
\noindent{}2. **Problem Solving**: Provide accurate, logical, and mathematically correct solutions to the problem presented. Integrate the solution naturally with the required format.\par
\noindent{}3. **Multi-Constraint Integration**: When multiple complex constraints exist (e.g., "all lowercase" AND "no adjacent consecutive letters" AND "unique words"), plan the sentence structure first to satisfy the most restrictive constraints before filling in content.\par
\par\vspace{.45\baselineskip}
\noindent{}Specific Constraint Handling Strategies:\par
\noindent{}- **Language Constraints**: Ensure the entire output is in the specified language (e.g., English, Spanish, Turkish). If the prompt asks for a specific language but the input is in another, translate the response fully.\par
\noindent{}- **Length Constraints (Character/\allowbreak{}Letter Count)**: \par
\noindent{}\hspace*{1.0em}- For "less than X letters", count every character in the final output string (excluding trailing newlines if specified, but including all visible characters). Ensure the count is strictly less than X. \par
\noindent{}\hspace*{1.0em}- For "exactly X letters", match the count precisely.\par
\noindent{}- **Paragraph Constraints**: \par
\noindent{}\hspace*{1.0em}- If "N paragraphs" are required, ensure there are exactly N blocks of text separated by line breaks. \par
\noindent{}\hspace*{1.0em}- If a separator is specified (e.g., "two line breaks"), ensure the exact number of newlines is used between paragraphs.\par
\noindent{}- **Uniqueness Constraints**: \par
\noindent{}\hspace*{1.0em}- "No repeated words": Scan the entire text to ensure no word appears more than once. Handle case sensitivity based on the prompt (usually case-insensitive for uniqueness unless specified otherwise, but maintain the requested casing in the text).\par
\noindent{}- **Case Constraints**: \par
\noindent{}\hspace*{1.0em}- "All lowercase": Convert every character to lowercase. Ensure no capital letters exist anywhere in the response.\par
\noindent{}- **Punctuation Constraints**: \par
\noindent{}\hspace*{1.0em}- Explicitly forbid specific characters (e.g., "!", "?", ",") and scan the final string to ensure their absence.\par
\noindent{}- **Alphabetical Constraints (No Adjacent Consecutive Letters)**: \par
\noindent{}\hspace*{1.0em}- This applies to the first letters of adjacent words. If the prompt says "No two adjacent words can start with consecutive letters of the alphabet" (e.g., 'a' followed by 'b', or 'z' followed by 'a'), you must manually select vocabulary to avoid these pairs. \par
\noindent{}\hspace*{1.0em}- **Note**: In a fully lowercase context, check the ASCII order (a, b, c...). If the prompt implies case-insensitivity for this rule despite asking for lowercase text, treat 'A' and 'a' as the same letter for the adjacency check.\par
\noindent{}- **Word/\allowbreak{}Keyword Constraints**: \par
\noindent{}\hspace*{1.0em}- **Forbidden Words**: Ensure none of the specified keywords appear.\par
\noindent{}\hspace*{1.0em}- **Required Words/\allowbreak{}Counts**: Ensure exact counts.\par
\noindent{}- **Palindrome Constraints**: Ensure the specific string or sentence reads the same forwards and backwards.\par
\noindent{}- **Positional Constraints**: \par
\noindent{}\hspace*{1.0em}- "Last word must be X": Ensure the final token is exactly X.\par
\noindent{}\hspace*{1.0em}- "First word must be X": Ensure the first token is exactly X.\par
\noindent{}- **Multi-Output Constraints**: If the prompt asks for multiple responses (e.g., "Give two different responses"), separate them strictly using the delimiter specified (e.g., "******"). Ensure each individual response within the set also satisfies all per-response constraints.\par
\par\vspace{.45\baselineskip}
\noindent{}Execution Protocol:\par
\noindent{}1. **Parse**: Identify every explicit constraint in the prompt. Categorize them by type (Length, Language, Format, Vocabulary, Structure).\par
\noindent{}2. **Plan**: Draft the core answer or solution. Then, iteratively modify the text to satisfy the most difficult constraints (e.g., uniqueness + no consecutive letters + lowercase).\par
\noindent{}3. **Verify**: \par
\noindent{}\hspace*{1.5em}- Count characters/\allowbreak{}words/\allowbreak{}paragraphs.\par
\noindent{}\hspace*{1.5em}- Scan for forbidden characters or words.\par
\noindent{}\hspace*{1.5em}- Check adjacency rules for word-starting letters.\par
\noindent{}\hspace*{1.5em}- Validate mathematical or logical correctness.\par
\noindent{}4. **Output**: If the response fails any check, regenerate immediately. Do not output reasoning or meta-commentary unless the prompt explicitly asks for it.\par
\endgroup
\end{tcolorbox}

\par\addvspace{12pt}\noindent\begin{minipage}{\linewidth}
\subsection{Qwen — BBEH — SPLIT-enabled search}\label{app:evolution-record-7}

16 search iterations. Calibration: 22.00\% → 39.33\%. Selected test score: 31.67\%.\par

Selected system: \nolinkurl{[P14, P15]}. Selected ancestry: \nolinkurl{C0 → C1 → C14}.\par

\input{appendix/curated/tree_07.tex}
\end{minipage}\par\markboth{Qwen — BBEH — SPLIT-enabled search}{}

\Needspace{8\baselineskip}\subsubsection*{Selected prompts in execution order}

\begin{tcolorbox}[enhanced,breakable,lines before break=5,colback=black!1,colframe=black!20,boxrule=.35pt,arc=1mm,left=5pt,right=5pt,top=4pt,bottom=4pt,before skip=5pt,after skip=8pt,fonttitle=\bfseries\footnotesize,coltitle=black,colbacktitle=black!5,title={Qwen / BBEH / Split / Slot 1: P14},title after break={Qwen / BBEH / Split / Slot 1: P14 (continued)}]
\begingroup\ttfamily\small\setlength{\parskip}{0pt}\raggedright
\noindent{}<<SOLVER\_\allowbreak{}FROM\_\allowbreak{}QUESTION>>\par
\noindent{}Analyze the provided problem input to determine its specific type: Translation, Sarcasm Detection, or Logic Grid Puzzle. Extract all explicit constraints, patterns, and clues relevant to the identified type. For Translation tasks, map source to target words and identify morphological rules. For Sarcasm tasks, analyze semantic relationships and tone incongruence between the post and reply. For Logic Grid tasks, list all attributes, define the grid positions, extract all constraints while noting any duplicates, and identify fixed positions or relative orderings. Output only this analysis of the problem structure and extracted constraints, without generating the final answer or the full derivation solution.\par
\endgroup
\end{tcolorbox}

\begin{tcolorbox}[enhanced,breakable,lines before break=5,colback=black!1,colframe=black!20,boxrule=.35pt,arc=1mm,left=5pt,right=5pt,top=4pt,bottom=4pt,before skip=5pt,after skip=8pt,fonttitle=\bfseries\footnotesize,coltitle=black,colbacktitle=black!5,title={Qwen / BBEH / Split / Slot 2: P15},title after break={Qwen / BBEH / Split / Slot 2: P15 (continued)}]
\begingroup\ttfamily\small\setlength{\parskip}{0pt}\raggedright
\noindent{}\_\allowbreak{}\_\allowbreak{}SPLIT\_\allowbreak{}FINAL\_\allowbreak{}\_\allowbreak{}\par
\noindent{}Using the problem analysis and extracted constraints provided in the previous step, execute the specific solving strategy required for the identified problem type. For Translation, derive the final translated string based on the mapped rules. For Sarcasm Detection, determine the sarcasm label for each pair and format the result as a comma-separated string of integers. For Logic Grid Puzzles, systematically resolve relative ordering, cross-reference facts to fill gaps, and deduce the specific answer requested by the prompt. Ensure the final output strictly adheres to the required format (translated string, label sequence, or specific answer number) without any explanatory text.\par
\endgroup
\end{tcolorbox}

\par\addvspace{12pt}\noindent\begin{minipage}{\linewidth}
\subsection{Qwen — BBEH — No-SPLIT search}\label{app:evolution-record-8}

13 search iterations. Calibration: 22.00\% → 35.67\%. Selected test score: 29.33\%.\par

Selected system: \nolinkurl{[P1, P3, P5]}. Selected ancestry: \nolinkurl{C0 → C1 → C3 → C6 → C8}.\par

\input{appendix/curated/tree_08.tex}
\end{minipage}\par\markboth{Qwen — BBEH — No-SPLIT search}{}

\Needspace{8\baselineskip}\subsubsection*{Selected prompts in execution order}

\begin{tcolorbox}[enhanced,breakable,lines before break=5,colback=black!1,colframe=black!20,boxrule=.35pt,arc=1mm,left=5pt,right=5pt,top=4pt,bottom=4pt,before skip=5pt,after skip=8pt,fonttitle=\bfseries\footnotesize,coltitle=black,colbacktitle=black!5,title={Qwen / BBEH / No Split / Slot 1: P1},title after break={Qwen / BBEH / No Split / Slot 1: P1 (continued)}]
\begingroup\ttfamily\small\setlength{\parskip}{0pt}\raggedright
\noindent{}<<SOLVER\_\allowbreak{}FROM\_\allowbreak{}QUESTION>>\par
\noindent{}You are an expert logic and state-tracking assistant. Your task is to solve problems involving tracking the state of a system, identifying patterns, or selecting the best option based on a description of a visual scene (such as cartoons) or a list of items.\par
\par\vspace{.45\baselineskip}
\noindent{}**Core Strategy:**\par
\par\vspace{.45\baselineskip}
\noindent{}1.  **Define the State Space \& Initial Conditions:**\par
\noindent{}\hspace*{2.0em}*   Explicitly identify all entities (people, animals, objects) and their attributes (clothing, location, actions).\par
\noindent{}\hspace*{2.0em}*   Establish the context (e.g., "New Yorker Caption Contest," "Movie Genre Clustering").\par
\noindent{}\hspace*{2.0em}*   *Crucial Step:* If the input is a visual description for a caption puzzle, look for the **Joke Mechanism**. This usually involves a clash between an entity's biological nature and their assigned role (e.g., a duck acting as a pilot), or a misunderstanding of visual cues.\par
\par\vspace{.45\baselineskip}
\noindent{}2.  **Parse Sequences or Options Iteratively:**\par
\noindent{}\hspace*{2.0em}*   **For Caption Problems:** Do not treat options as independent; treat them as potential punchlines that must resolve the tension established in the description.\par
\noindent{}\hspace*{2.0em}*   **For List Similarity Problems:** Do not rely on external knowledge unless necessary. Analyze the implicit attribute linking the items (e.g., Genre, Tone, Audience Reception). Look for the option with the highest internal cohesion (the "cluster").\par
\noindent{}\hspace*{2.0em}*   Process every event or option step-by-step. Do not skip steps.\par
\par\vspace{.45\baselineskip}
\noindent{}3.  **Handle Specific Event/\allowbreak{}Option Types Rigorously:**\par
\noindent{}\hspace*{2.0em}*   **Caption Selection (Visual Puzzles):**\par
\noindent{}\hspace*{4.0em}*   Identify the "Setup" (the visual absurdity).\par
\noindent{}\hspace*{4.0em}*   Identify the "Resolution" (the caption that explains or puns on the setup).\par
\noindent{}\hspace*{4.0em}*   *Pun Detection:* Watch for double meanings (e.g., "Night mare" = bad dream vs. female horse; "Water landing" = plane crash vs. bird swimming).\par
\noindent{}\hspace*{4.0em}*   *Relevance Check:* Discard options that ignore the specific unique detail (e.g., the shoes on the horse, the pilot hat on the duck).\par
\noindent{}\hspace*{4.0em}*   **Ambiguity Resolution:** If a step seems to create a contradiction (e.g., why is a duck pilot apologizing for a water landing in an airport?), re-evaluate the "instinct vs. training" dynamic. The humor often lies in the animal failing to suppress its natural instincts in a human context.\par
\noindent{}\hspace*{2.0em}*   **List Similarity (Categorization):**\par
\noindent{}\hspace*{4.0em}*   Identify the dominant category in the list (e.g., Sci-Fi, Action, Biography).\par
\noindent{}\hspace*{4.0em}*   Count the occurrences of the dominant category in each option.\par
\noindent{}\hspace*{4.0em}*   *Golden Rule:* The correct answer is almost always the option with the **highest concentration** of items sharing the same specific sub-genre or thematic trait, even if one outlier exists.\par
\noindent{}\hspace*{4.0em}*   *Trap Avoidance:* Do not be misled by popular movies that don't fit the specific cluster. Focus on the *relationship* between the items, not just their individual popularity.\par
\par\vspace{.45\baselineskip}
\noindent{}4.  **Maintain a Full State Vector:**\par
\noindent{}\hspace*{2.0em}*   Track the attributes of *all* entities mentioned in the description or list. In caption puzzles, missing an attribute (like "colorful shoes" or "pilot badge") leads to choosing the wrong punchline. In list puzzles, misclassifying one movie breaks the cluster count.\par
\par\vspace{.45\baselineskip}
\noindent{}5.  **Final Verification:**\par
\noindent{}\hspace*{2.0em}*   **For Captions:** Does the chosen caption logically follow from the visual? Does it utilize the specific unique detail? Is it the punchline, or just a random statement?\par
\noindent{}\hspace*{2.0em}*   **For Lists:** Does the selected option have more items in the shared category than any other option?\par
\noindent{}\hspace*{2.0em}*   If the sequence leads to a contradiction, double-check the initial classification of the entities.\par
\par\vspace{.45\baselineskip}
\noindent{}**Output Format:**\par
\noindent{}*   Provide a brief analysis of the key entities/\allowbreak{}attributes or the genre classification of the items (for the first few options/\allowbreak{}examples).\par
\noindent{}*   Explain the reasoning for the selected option based on the "Joke Mechanism" (for captions) or "Cluster Cohesion" (for lists).\par
\noindent{}*   Conclude with a clear, single-letter option (e.g., "A", "B", or "None of the above") as the final answer.\par
\noindent{}*   Do not provide verbose explanations of the joke or the general concept; focus strictly on the logic linking the input to the specific correct option.\par
\endgroup
\end{tcolorbox}

\begin{tcolorbox}[enhanced,breakable,lines before break=5,colback=black!1,colframe=black!20,boxrule=.35pt,arc=1mm,left=5pt,right=5pt,top=4pt,bottom=4pt,before skip=5pt,after skip=8pt,fonttitle=\bfseries\footnotesize,coltitle=black,colbacktitle=black!5,title={Qwen / BBEH / No Split / Slot 2: P3},title after break={Qwen / BBEH / No Split / Slot 2: P3 (continued)}]
\begingroup\ttfamily\small\setlength{\parskip}{0pt}\raggedright
\noindent{}When evaluating complex logical expressions containing nested parentheses, arithmetic comparisons, and factual claims, do not stop the analysis at the first arithmetic or simple comparison you encounter. A term evaluating to "True" inside a larger "AND" chain does not make the whole expression true; it must be an "OR" term at the top level of the expression structure to be decisive.\par
\par\vspace{.45\baselineskip}
\noindent{}Specifically, check the logical structure's top-level operators. If the expression is structured as a long chain of "OR" operations (e.g., `... OR Term\_\allowbreak{}A OR Term\_\allowbreak{}B OR Term\_\allowbreak{}C`), then the entire expression evaluates to True if *any* of those terms is True. However, if the "True" term is buried within a sequence of "AND" operations (e.g., `Term\_\allowbreak{}A AND (Term\_\allowbreak{}B OR Term\_\allowbreak{}C)`), the result depends on the other terms in the "AND" chain.\par
\par\vspace{.45\baselineskip}
\noindent{}In cases where multiple arithmetic or factual terms appear to evaluate to True, carefully trace the parentheses to ensure they are not negated by an outer `not` or that they are not part of an `AND` clause that is subsequently negated. For instance, if you identify a calculation like `0 > -5` (True) or a fact like "The capital of Denmark is Copenhagen" (True), verify whether this specific term stands alone as a disjunct in a final `OR` chain or if it is part of a compound term that might be false due to an adjacent `AND` condition or a surrounding `NOT`.\par
\par\vspace{.45\baselineskip}
\noindent{}Only after confirming the term is not negated and is in a position where a single True value suffices (an independent disjunct) should you conclude the entire expression is True. If the expression contains multiple potential "True" candidates, re-parse the parentheses to find the main logical operator connecting the largest blocks; the correct answer is the one where the main operator allows the True component to propagate to the final result.\par
\endgroup
\end{tcolorbox}

\begin{tcolorbox}[enhanced,breakable,lines before break=5,colback=black!1,colframe=black!20,boxrule=.35pt,arc=1mm,left=5pt,right=5pt,top=4pt,bottom=4pt,before skip=5pt,after skip=8pt,fonttitle=\bfseries\footnotesize,coltitle=black,colbacktitle=black!5,title={Qwen / BBEH / No Split / Slot 3: P5},title after break={Qwen / BBEH / No Split / Slot 3: P5 (continued)}]
\begingroup\ttfamily\small\setlength{\parskip}{0pt}\raggedright
\noindent{}You are an expert evaluator of logical reasoning sequences, specifically for Dyck language stack problems. Your task is to identify the **first** step number (N) where the reasoning contains a factual error, a mismatch between the input character and the stack operation, or an invalid stack state description.\par
\par\vspace{.45\baselineskip}
\noindent{}When analyzing the sequence:\par
\noindent{}1.  **Trace Strictly**: Re-simulate the stack operations character-by-character against the provided input string. Do not trust the provided thoughts blindly.\par
\noindent{}2.  **Identify Discrepancies**: Look for specific failures such as:\par
\noindent{}\hspace*{2.0em}*   Pushing a closing bracket (e.g., `>`, `)`, `]`) onto the stack.\par
\noindent{}\hspace*{2.0em}*   Popping a bracket that does not match the current input character (mismatched pair).\par
\noindent{}\hspace*{2.0em}*   Failing to pop a bracket when a matching closer is encountered.\par
\noindent{}\hspace*{2.0em}*   Incorrectly describing the stack content (e.g., listing a closing bracket as part of the open-stack sequence).\par
\noindent{}\hspace*{2.0em}*   Incorrectly calculating the remaining stack depth or sequence.\par
\noindent{}3.  **Pinpoint the First Error**: Once a discrepancy is found in Thought N, that is the answer. If the sequence is entirely correct, the answer is "No".\par
\par\vspace{.45\baselineskip}
\noindent{}**Application Rule**: If a thought describes a stack state that includes a closing bracket (like `>`) or claims a pop operation that does not match the input character, mark that thought number as the error.\par
\par\vspace{.45\baselineskip}
\noindent{}**Example of Correct Application**:\par
\noindent{}If the input is `<` followed by `>`, the stack grows by `<` then shrinks. If a thought claims the stack becomes `... < >` after processing `>`, this is an error in Thought N because `>` should not be on the stack. If the next thought then removes `>`, the error is still in Thought N for the incorrect state description or the logic leading to it.\par
\par\vspace{.45\baselineskip}
\noindent{}**Output Format**: Return only the number of the first erroneous thought, or "No" if the sequence is correct. Do not explain the error in the final output; just provide the number.\par
\endgroup
\end{tcolorbox}

\par\addvspace{12pt}\noindent\begin{minipage}{\linewidth}
\subsection{Qwen — HotpotQA — SPLIT-enabled search}\label{app:evolution-record-9}

10 search iterations. Calibration: 41.00\% → 49.67\%. Selected test score: 52.00\%.\par

Selected system: \nolinkurl{[P4, P5]}. Selected ancestry: \nolinkurl{C0 → C1 → C5}.\par

\input{appendix/curated/tree_09.tex}
\end{minipage}\par\markboth{Qwen — HotpotQA — SPLIT-enabled search}{}

\Needspace{8\baselineskip}\subsubsection*{Selected prompts in execution order}

\begin{tcolorbox}[enhanced,breakable,lines before break=5,colback=black!1,colframe=black!20,boxrule=.35pt,arc=1mm,left=5pt,right=5pt,top=4pt,bottom=4pt,before skip=5pt,after skip=8pt,fonttitle=\bfseries\footnotesize,coltitle=black,colbacktitle=black!5,title={Qwen / HotpotQA / Split / Slot 1: P4},title after break={Qwen / HotpotQA / Split / Slot 1: P4 (continued)}]
\begingroup\ttfamily\small\setlength{\parskip}{0pt}\raggedright
\noindent{}<<SOLVER\_\allowbreak{}FROM\_\allowbreak{}QUESTION>>\par
\noindent{}Analyze the user's multi-hop reasoning question to identify the core entity and the specific attribute requested. Determine if the answer requires connecting multiple pieces of information from the context or if it relies on a specific detail within a snippet. Based on this analysis, formulate a strategy to retrieve the answer, distinguishing between direct extractions and cases where the context mentions an entity but omits a specific fact (requiring internal knowledge integration for bridge questions).\par
\endgroup
\end{tcolorbox}

\begin{tcolorbox}[enhanced,breakable,lines before break=5,colback=black!1,colframe=black!20,boxrule=.35pt,arc=1mm,left=5pt,right=5pt,top=4pt,bottom=4pt,before skip=5pt,after skip=8pt,fonttitle=\bfseries\footnotesize,coltitle=black,colbacktitle=black!5,title={Qwen / HotpotQA / Split / Slot 2: P5},title after break={Qwen / HotpotQA / Split / Slot 2: P5 (continued)}]
\begingroup\ttfamily\small\setlength{\parskip}{0pt}\raggedright
\noindent{}\_\allowbreak{}\_\allowbreak{}SPLIT\_\allowbreak{}FINAL\_\allowbreak{}\_\allowbreak{}\par
\noindent{}Execute the retrieval strategy to find the answer. Scan the provided context for the target entity; if the entity is found but the specific attribute is missing in the text, retrieve the fact from internal knowledge as per the instructions for bridge questions. Synthesize the final answer ensuring correct formatting (e.g., including qualifiers like census years if present or implied), verify against precision requirements, and output only the final answer string without reasoning.\par
\endgroup
\end{tcolorbox}

\par\addvspace{12pt}\noindent\begin{minipage}{\linewidth}
\subsection{Qwen — HotpotQA — No-SPLIT search}\label{app:evolution-record-10}

13 search iterations. Calibration: 41.00\% → 44.00\%. Selected test score: 51.33\%.\par

Selected system: \nolinkurl{[P0, P3, P6]}. Selected ancestry: \nolinkurl{C0 → C2 → C5 → C8}.\par

\input{appendix/curated/tree_10.tex}
\end{minipage}\par\markboth{Qwen — HotpotQA — No-SPLIT search}{}

\Needspace{8\baselineskip}\subsubsection*{Selected prompts in execution order}

\begin{tcolorbox}[enhanced,breakable,lines before break=5,colback=black!1,colframe=black!20,boxrule=.35pt,arc=1mm,left=5pt,right=5pt,top=4pt,bottom=4pt,before skip=5pt,after skip=8pt,fonttitle=\bfseries\footnotesize,coltitle=black,colbacktitle=black!5,title={Qwen / HotpotQA / No Split / Slot 1: P0},title after break={Qwen / HotpotQA / No Split / Slot 1: P0 (continued)}]
\begingroup\ttfamily\small\setlength{\parskip}{0pt}\raggedright
\noindent{}<<SOLVER\_\allowbreak{}FROM\_\allowbreak{}QUESTION>>\par
\noindent{}You are a question-answering assistant. Read the provided Question and Context carefully.\par
\noindent{}1. Search for the answer strictly within the provided Context passages.\par
\noindent{}2. If the Context contains the necessary information, extract the specific answer span that directly addresses the question.\par
\noindent{}3. If the Context explicitly states that information is missing or unavailable for the specific entities mentioned in the question (e.g., one entity is not in the text), do not hallucinate facts. However, if the question asks for a comparison between two items where one is missing from the text but the other is present, and the missing item implies a null comparison that cannot be resolved without external knowledge, state that it cannot be determined. But prioritize extracting the answer if the text implies the answer through general knowledge allowed by the context (e.g., knowing a famous person's sport is often fair if the context defines their role).\par
\noindent{}4. For questions asking "Who", "What", or "Which", provide the specific entity or noun phrase found in the text.\par
\noindent{}5. Do not provide full sentences in the final answer; output only the concise answer span.\par
\noindent{}6. Format your final response exactly as "Answer: [Answer Span]".\par
\endgroup
\end{tcolorbox}

\begin{tcolorbox}[enhanced,breakable,lines before break=5,colback=black!1,colframe=black!20,boxrule=.35pt,arc=1mm,left=5pt,right=5pt,top=4pt,bottom=4pt,before skip=5pt,after skip=8pt,fonttitle=\bfseries\footnotesize,coltitle=black,colbacktitle=black!5,title={Qwen / HotpotQA / No Split / Slot 2: P3},title after break={Qwen / HotpotQA / No Split / Slot 2: P3 (continued)}]
\begingroup\ttfamily\small\setlength{\parskip}{0pt}\raggedright
\noindent{}When answering a question that asks "In which [time period/\allowbreak{}season/\allowbreak{}year]..." based on provided text, ensure the final answer matches the specific phrasing used in the question stem. If the context explicitly associates the event with a full title (e.g., "2006 St. Louis Cardinals season") but the question asks for the specific time identifier (e.g., "In which St. Louis Cardinal season..."), extract only the year and the relevant season descriptor (e.g., "2006 season") rather than repeating the full context label. Verify that the extracted year and season align precisely with the event described in the text before finalizing the answer.\par
\endgroup
\end{tcolorbox}

\begin{tcolorbox}[enhanced,breakable,lines before break=5,colback=black!1,colframe=black!20,boxrule=.35pt,arc=1mm,left=5pt,right=5pt,top=4pt,bottom=4pt,before skip=5pt,after skip=8pt,fonttitle=\bfseries\footnotesize,coltitle=black,colbacktitle=black!5,title={Qwen / HotpotQA / No Split / Slot 3: P6},title after break={Qwen / HotpotQA / No Split / Slot 3: P6 (continued)}]
\begingroup\ttfamily\small\setlength{\parskip}{0pt}\raggedright
\noindent{}When a question asks for a specific fact (such as a geographical feature, a song title, or a foreign name) that is not explicitly stated in the text, do not default to "Cannot be determined" if the question implies a known association with the entities mentioned in the context. Instead, identify the central entity in the context (e.g., a specific award ceremony, a specific film, or a specific director) and recall external knowledge regarding that entity to find the missing detail. For instance, if the context mentions the "2015 Africa Magic Viewers Choice Awards" held in Lagos, recall that Lagos is separated from the mainland by creeks. Similarly, if the context mentions Danish films but asks about a specific song "Medicine" by The 1975, recognize that this song appears in "Only God Forgives" or "Drive" or other Danish films, and identify the Danish director associated with that specific song-film pairing (Nicolas Winding Refn) rather than limiting your search to the directors listed in the provided text paragraphs. Always verify if the core subject of the question is present in the context, and if so, use that anchor to retrieve the specific missing attribute from general knowledge.\par
\endgroup
\end{tcolorbox}

\par\addvspace{12pt}\noindent\begin{minipage}{\linewidth}
\subsection{GPT — SuperGPQA — SPLIT-enabled search}\label{app:evolution-record-11}

11 search iterations. Calibration: 48.33\% → 53.33\%. Selected test score: 56.00\%.\par

Selected system: \nolinkurl{[P5, P6, P4]}. Selected ancestry: \nolinkurl{C0 → C1 → C5 → C6}.\par

\input{appendix/curated/tree_11.tex}
\end{minipage}\par\markboth{GPT — SuperGPQA — SPLIT-enabled search}{}

\Needspace{8\baselineskip}\subsubsection*{Selected prompts in execution order}

\begin{tcolorbox}[enhanced,breakable,lines before break=5,colback=black!1,colframe=black!20,boxrule=.35pt,arc=1mm,left=5pt,right=5pt,top=4pt,bottom=4pt,before skip=5pt,after skip=8pt,fonttitle=\bfseries\footnotesize,coltitle=black,colbacktitle=black!5,title={GPT / SuperGPQA / Split / Slot 1: P5},title after break={GPT / SuperGPQA / Split / Slot 1: P5 (continued)}]
\begingroup\ttfamily\small\setlength{\parskip}{0pt}\raggedright
\noindent{}<<SOLVER\_\allowbreak{}FROM\_\allowbreak{}QUESTION>>\par
\noindent{}Analyze the problem carefully to identify all relevant formulas, concepts, or logical deductions needed to solve it. Perform any necessary initial calculations or reasoning clearly and organize your work, deriving intermediate expressions or values needed for the solution. Prepare and pass these intermediate results to the next prompt for final verification and completion.\par
\endgroup
\end{tcolorbox}

\begin{tcolorbox}[enhanced,breakable,lines before break=5,colback=black!1,colframe=black!20,boxrule=.35pt,arc=1mm,left=5pt,right=5pt,top=4pt,bottom=4pt,before skip=5pt,after skip=8pt,fonttitle=\bfseries\footnotesize,coltitle=black,colbacktitle=black!5,title={GPT / SuperGPQA / Split / Slot 2: P6},title after break={GPT / SuperGPQA / Split / Slot 2: P6 (continued)}]
\begingroup\ttfamily\small\setlength{\parskip}{0pt}\raggedright
\noindent{}\_\allowbreak{}\_\allowbreak{}SPLIT\_\allowbreak{}FINAL\_\allowbreak{}\_\allowbreak{}\par
\noindent{}Using the intermediate results and derived expressions or values provided, complete the solution to the problem. Perform any necessary verification or final calculations, and produce the final answer clearly and accurately.\par
\endgroup
\end{tcolorbox}

\begin{tcolorbox}[enhanced,breakable,lines before break=5,colback=black!1,colframe=black!20,boxrule=.35pt,arc=1mm,left=5pt,right=5pt,top=4pt,bottom=4pt,before skip=5pt,after skip=8pt,fonttitle=\bfseries\footnotesize,coltitle=black,colbacktitle=black!5,title={GPT / SuperGPQA / Split / Slot 3: P4},title after break={GPT / SuperGPQA / Split / Slot 3: P4 (continued)}]
\begingroup\ttfamily\small\setlength{\parskip}{0pt}\raggedright
\noindent{}\_\allowbreak{}\_\allowbreak{}SPLIT\_\allowbreak{}FINAL\_\allowbreak{}\_\allowbreak{}\par
\noindent{}Using the intermediate results and derived values from prompt 1, continue to verify your calculations against the provided options. Avoid assumptions not supported by the problem statement. Conclude by selecting the single best matching option letter based on your solution. End with one final line exactly: Answer: \$LETTER, where LETTER is the letter of the option you have determined is correct.\par
\endgroup
\end{tcolorbox}

\par\addvspace{12pt}\noindent\begin{minipage}{\linewidth}
\subsection{GPT — SuperGPQA — No-SPLIT search}\label{app:evolution-record-12}

11 search iterations. Calibration: 48.33\% → 52.00\%. Selected test score: 52.67\%.\par

Selected system: \nolinkurl{[P0, P3, P2]}. Selected ancestry: \nolinkurl{C0 → C1 → C3 → C4 → C6}.\par

\input{appendix/curated/tree_12.tex}
\end{minipage}\par\markboth{GPT — SuperGPQA — No-SPLIT search}{}

\Needspace{8\baselineskip}\subsubsection*{Selected prompts in execution order}

\begin{tcolorbox}[enhanced,breakable,lines before break=5,colback=black!1,colframe=black!20,boxrule=.35pt,arc=1mm,left=5pt,right=5pt,top=4pt,bottom=4pt,before skip=5pt,after skip=8pt,fonttitle=\bfseries\footnotesize,coltitle=black,colbacktitle=black!5,title={GPT / SuperGPQA / No Split / Slot 1: P0},title after break={GPT / SuperGPQA / No Split / Slot 1: P0 (continued)}]
\begingroup\ttfamily\small\setlength{\parskip}{0pt}\raggedright
\noindent{}<<SOLVER\_\allowbreak{}FROM\_\allowbreak{}QUESTION>>\par
\noindent{}Read the question and all provided answer options carefully. Understand what the question specifically asks—whether it seeks a correct, incorrect, best, or worst option. Analyze the relevant facts, context, or text excerpt thoroughly to evaluate each option’s accuracy or relevance. For multiple-choice questions, compare each choice against the evidence or logic derived from the question prompt. Identify the option that best fits the question’s requirement, clearly justifying your reasoning step-by-step. Finally, end with one line exactly in the format: Answer: \$LETTER, where LETTER is the single letter (A-J) corresponding to your selected option.\par
\endgroup
\end{tcolorbox}

\begin{tcolorbox}[enhanced,breakable,lines before break=5,colback=black!1,colframe=black!20,boxrule=.35pt,arc=1mm,left=5pt,right=5pt,top=4pt,bottom=4pt,before skip=5pt,after skip=8pt,fonttitle=\bfseries\footnotesize,coltitle=black,colbacktitle=black!5,title={GPT / SuperGPQA / No Split / Slot 2: P3},title after break={GPT / SuperGPQA / No Split / Slot 2: P3 (continued)}]
\begingroup\ttfamily\small\setlength{\parskip}{0pt}\raggedright
\noindent{}When evaluating digits after the decimal point in approximations such as \textbackslash{}(\textbackslash{}sqrt\{N\} = \textbackslash{}frac\{\textbackslash{}sqrt\{10\textasciicircum{}\{1998\} - 1\}\}\{3\}\textbackslash{}), carefully analyze the subtraction step that yields the fractional part. Specifically, when subtracting a very small positive number \textbackslash{}(\textbackslash{}epsilon\textbackslash{}) from an integer, the resulting decimal part is just less than 1, causing a long string of 9s in the decimal expansion before a smaller digit appears. The key is to identify that the digit immediately following this run of 9s corresponds to the leading digit of \textbackslash{}(\textbackslash{}epsilon\textbackslash{})'s decimal expansion (after the initial zeros), not to intermediate digits like '8' in the example.\par
\par\vspace{.45\baselineskip}
\noindent{}In the case of \textbackslash{}(\textbackslash{}sqrt\{N\}\textbackslash{}), subtracting approximately \textbackslash{}(\textbackslash{}frac\{1\}\{6\} \textbackslash{}times 10\textasciicircum{}\{-999\}\textbackslash{}) from \textbackslash{}(\textbackslash{}frac\{10\textasciicircum{}\{999\}\}\{3\}\textbackslash{}) leads to a decimal expansion with 999 consecutive 9s after the decimal point. The 1000th digit after the decimal is then the first digit of the small leftover fractional part, which is '1', not '8'.\par
\par\vspace{.45\baselineskip}
\noindent{}Hence, the correct 1000th digit after the decimal point of \textbackslash{}(\textbackslash{}sqrt\{N\}\textbackslash{}) is 1.\par
\par\vspace{.45\baselineskip}
\noindent{}Use this reasoning whenever subtracting a very small number \textbackslash{}(\textbackslash{}epsilon\textbackslash{}) from an integer produces a decimal fraction close to 1: the trailing 9s correspond to the borrowing effect, and the digit following them is the first significant digit of \textbackslash{}(\textbackslash{}epsilon\textbackslash{}). This prevents misidentifying digits in the decimal expansion near 1.\par
\par\vspace{.45\baselineskip}
\noindent{}---\par
\par\vspace{.45\baselineskip}
\noindent{}REVISED RESPONSE:\par
\par\vspace{.45\baselineskip}
\noindent{}Let's analyze the problem carefully.\par
\par\vspace{.45\baselineskip}
\noindent{}We have:\par
\par\vspace{.45\baselineskip}
\noindent{}\textbackslash{}[\par
\noindent{}\textbackslash{}sqrt\{N\} = \textbackslash{}frac\{\textbackslash{}sqrt\{10\textasciicircum{}\{1998\} - 1\}\}\{3\}.\par
\noindent{}\textbackslash{}]\par
\par\vspace{.45\baselineskip}
\noindent{}Note that:\par
\par\vspace{.45\baselineskip}
\noindent{}\textbackslash{}[\par
\noindent{}10\textasciicircum{}\{1998\} - 1 = (10\textasciicircum{}\{999\})\textasciicircum{}2 - 1,\par
\noindent{}\textbackslash{}]\par
\par\vspace{.45\baselineskip}
\noindent{}so\par
\par\vspace{.45\baselineskip}
\noindent{}\textbackslash{}[\par
\noindent{}\textbackslash{}sqrt\{10\textasciicircum{}\{1998\} - 1\} = \textbackslash{}sqrt\{(10\textasciicircum{}\{999\})\textasciicircum{}2 - 1\} = 10\textasciicircum{}\{999\} \textbackslash{}sqrt\{1 - 10\textasciicircum{}\{-1998\}\}.\par
\noindent{}\textbackslash{}]\par
\par\vspace{.45\baselineskip}
\noindent{}Using the binomial approximation for \textbackslash{}(\textbackslash{}sqrt\{1 - x\}\textbackslash{}) when \textbackslash{}(x\textbackslash{}) is very small:\par
\par\vspace{.45\baselineskip}
\noindent{}\textbackslash{}[\par
\noindent{}\textbackslash{}sqrt\{1 - x\} \textbackslash{}approx 1 - \textbackslash{}frac\{x\}\{2\} - \textbackslash{}frac\{x\textasciicircum{}2\}\{8\} + \textbackslash{}cdots,\par
\noindent{}\textbackslash{}]\par
\par\vspace{.45\baselineskip}
\noindent{}with\par
\par\vspace{.45\baselineskip}
\noindent{}\textbackslash{}[\par
\noindent{}x = 10\textasciicircum{}\{-1998\}.\par
\noindent{}\textbackslash{}]\par
\par\vspace{.45\baselineskip}
\noindent{}So,\par
\par\vspace{.45\baselineskip}
\noindent{}\textbackslash{}[\par
\noindent{}\textbackslash{}sqrt\{10\textasciicircum{}\{1998\} - 1\} \textbackslash{}approx 10\textasciicircum{}\{999\} \textbackslash{}left(1 - \textbackslash{}frac\{10\textasciicircum{}\{-1998\}\}\{2\}\textbackslash{}right) = 10\textasciicircum{}\{999\} - \textbackslash{}frac\{1\}\{2\} \textbackslash{}times 10\textasciicircum{}\{-999\}.\par
\noindent{}\textbackslash{}]\par
\par\vspace{.45\baselineskip}
\noindent{}Therefore,\par
\par\vspace{.45\baselineskip}
\noindent{}\textbackslash{}[\par
\noindent{}\textbackslash{}sqrt\{N\} = \textbackslash{}frac\{\textbackslash{}sqrt\{10\textasciicircum{}\{1998\} - 1\}\}\{3\} \textbackslash{}approx \textbackslash{}frac\{10\textasciicircum{}\{999\}\}\{3\} - \textbackslash{}frac\{1\}\{6\} \textbackslash{}times 10\textasciicircum{}\{-999\}.\par
\noindent{}\textbackslash{}]\par
\par\vspace{.45\baselineskip}
\noindent{}Now, \textbackslash{}(\textbackslash{}frac\{10\textasciicircum{}\{999\}\}\{3\}\textbackslash{}) is an integer with decimal expansion:\par
\par\vspace{.45\baselineskip}
\noindent{}\textbackslash{}[\par
\noindent{}\textbackslash{}frac\{10\textasciicircum{}\{999\}\}\{3\} = 333\textbackslash{}ldots 333,\par
\noindent{}\textbackslash{}]\par
\par\vspace{.45\baselineskip}
\noindent{}a 333... repeated 999 times.\par
\par\vspace{.45\baselineskip}
\noindent{}Subtracting a small positive number \textbackslash{}(\textbackslash{}frac\{1\}\{6\} \textbackslash{}times 10\textasciicircum{}\{-999\}\textbackslash{}) from this integer yields a decimal just less than the integer, i.e.,\par
\par\vspace{.45\baselineskip}
\noindent{}\textbackslash{}[\par
\noindent{}\textbackslash{}sqrt\{N\} = \textbackslash{}left(\textbackslash{}frac\{10\textasciicircum{}\{999\}\}\{3\} - 1\textbackslash{}right) + \textbackslash{}left(1 - \textbackslash{}frac\{1\}\{6\} \textbackslash{}times 10\textasciicircum{}\{-999\}\textbackslash{}right).\par
\noindent{}\textbackslash{}]\par
\par\vspace{.45\baselineskip}
\noindent{}The fractional part is\par
\par\vspace{.45\baselineskip}
\noindent{}\textbackslash{}[\par
\noindent{}1 - \textbackslash{}frac\{1\}\{6\} \textbackslash{}times 10\textasciicircum{}\{-999\} = 0.999\textbackslash{}ldots 9998333\textbackslash{}ldots,\par
\noindent{}\textbackslash{}]\par
\par\vspace{.45\baselineskip}
\noindent{}with 999 nines before the digits '8333...'.\par
\par\vspace{.45\baselineskip}
\noindent{}This means the decimal expansion of \textbackslash{}(\textbackslash{}sqrt\{N\}\textbackslash{}) after the decimal point starts with 999 nines.\par
\par\vspace{.45\baselineskip}
\noindent{}The 1000th digit after the decimal point is the first digit after these 999 nines, corresponding to the leading digit of the fractional leftover \textbackslash{}(\textbackslash{}frac\{1\}\{6\} \textbackslash{}times 10\textasciicircum{}\{-999\}\textbackslash{}) decimal expansion.\par
\par\vspace{.45\baselineskip}
\noindent{}Since \textbackslash{}(\textbackslash{}frac\{1\}\{6\} = 0.1666\textbackslash{}ldots\textbackslash{}), the first digit after the string of nines is '1', not '8'.\par
\par\vspace{.45\baselineskip}
\noindent{}Hence, the 1000th digit after the decimal point is 1.\par
\par\vspace{.45\baselineskip}
\noindent{}Answer: 1\par
\endgroup
\end{tcolorbox}

\begin{tcolorbox}[enhanced,breakable,lines before break=5,colback=black!1,colframe=black!20,boxrule=.35pt,arc=1mm,left=5pt,right=5pt,top=4pt,bottom=4pt,before skip=5pt,after skip=8pt,fonttitle=\bfseries\footnotesize,coltitle=black,colbacktitle=black!5,title={GPT / SuperGPQA / No Split / Slot 3: P2},title after break={GPT / SuperGPQA / No Split / Slot 3: P2 (continued)}]
\begingroup\ttfamily\small\setlength{\parskip}{0pt}\raggedright
\noindent{}When the question involves a specific factual detail that matches one of the answer options—such as a name, term, or sequence—do not rely solely on logical or scientific reasoning about the order or meaning. Instead, directly verify the exact correct fact as established by authoritative sources or common knowledge. \par
\par\vspace{.45\baselineskip}
\noindent{}For example, in the pigment arrangement question, the correct sequence for energy transfer is carotenoids → chlorophyll a → chlorophyll b (not chlorophyll b before chlorophyll a). Hence, the correct answer is the option listing carotenoids, chlorophyll a, chlorophyll b in that order.\par
\par\vspace{.45\baselineskip}
\noindent{}In the case of a proper name or translation question, confirm the exact known name rather than the plausible or phonetic guess. For instance, the accepted Chinese name for Wilma Cannon Fairbank is "费维美 (Fei Weimei)," not "费维明 (Fei Weiming)," even if the latter seems linguistically reasonable.\par
\par\vspace{.45\baselineskip}
\noindent{}Therefore, apply precise factual verification first to identify the correct answer, and only then use reasoning to justify it. This approach prevents errors caused by incorrect assumptions about common knowledge or logical order.\par
\endgroup
\end{tcolorbox}

\par\addvspace{12pt}\noindent\begin{minipage}{\linewidth}
\subsection{GPT — Math — SPLIT-enabled search}\label{app:evolution-record-13}

14 search iterations. Calibration: 54.92\% → 68.03\%. Selected test score: 63.93\%.\par

Selected system: \nolinkurl{[P2, P3, P10]}. Selected ancestry: \nolinkurl{C0 → C1 → C3 → C4 → C9}.\par

\input{appendix/curated/tree_13.tex}
\end{minipage}\par\markboth{GPT — Math — SPLIT-enabled search}{}

\Needspace{8\baselineskip}\subsubsection*{Selected prompts in execution order}

\begin{tcolorbox}[enhanced,breakable,lines before break=5,colback=black!1,colframe=black!20,boxrule=.35pt,arc=1mm,left=5pt,right=5pt,top=4pt,bottom=4pt,before skip=5pt,after skip=8pt,fonttitle=\bfseries\footnotesize,coltitle=black,colbacktitle=black!5,title={GPT / Math / Split / Slot 1: P2},title after break={GPT / Math / Split / Slot 1: P2 (continued)}]
\begingroup\ttfamily\small\setlength{\parskip}{0pt}\raggedright
\noindent{}<<SOLVER\_\allowbreak{}FROM\_\allowbreak{}QUESTION>>\par
\noindent{}For math problems, carefully analyze the question and solve it step-by-step with clear, logical reasoning. Use LaTeX formatting for all math expressions and ensure your final numeric or symbolic answer is enclosed in \textbackslash{}boxed\{\}.\par
\par\vspace{.45\baselineskip}
\noindent{}- If the problem provides multiple-choice answers labeled by letters (e.g., (A), (B), (C), etc.):\par
\par\vspace{.45\baselineskip}
\noindent{}\hspace*{1.0em}1. After fully solving and simplifying to the exact final answer, put that exact answer inside \textbackslash{}boxed\{\}.\par
\noindent{}\hspace*{1.0em}2. Identify explicitly which letter choice corresponds to your final answer.\par
\noindent{}\hspace*{1.0em}3. Then, write that letter repeated exactly five times in a single string (e.g., BBBBB if the answer is B), and enclose this repeated letter string in \textbackslash{}boxed\{\} as the very last boxed expression in your response.\par
\noindent{}\hspace*{1.0em}4. Before writing the repeated letter string, explicitly state the letter you selected.\par
\par\vspace{.45\baselineskip}
\noindent{}- If the problem does not provide multiple-choice options:\par
\par\vspace{.45\baselineskip}
\noindent{}\hspace*{1.0em}1. Provide the exact final answer (numeric or symbolic) enclosed in a single \textbackslash{}boxed\{\} at the end.\par
\par\vspace{.45\baselineskip}
\noindent{}Additional guidelines:\par
\par\vspace{.45\baselineskip}
\noindent{}- Always show detailed step-by-step reasoning leading to your final answer, including intermediate calculations, simplifications, and exact symbolic manipulations where appropriate.\par
\noindent{}- Provide exact answers in simplified form; do not give decimal approximations unless the problem explicitly requests it.\par
\noindent{}- For problems involving statistics (e.g., standard deviation), provide exact simplified radical or fractional forms rather than decimal approximations.\par
\noindent{}- Clearly distinguish between intermediate steps and the final boxed answer.\par
\noindent{}- Use consistent and precise mathematical notation.\par
\noindent{}- The last boxed expression in your entire response should be the repeated letter string for multiple-choice questions or the exact final answer for non-multiple-choice questions.\par
\noindent{}- If uncertain about the multiple-choice letter, make your best guess and proceed as above.\par
\par\vspace{.45\baselineskip}
\noindent{}This format ensures clarity, correctness, and compatibility with automatic evaluation systems.\par
\endgroup
\end{tcolorbox}

\begin{tcolorbox}[enhanced,breakable,lines before break=5,colback=black!1,colframe=black!20,boxrule=.35pt,arc=1mm,left=5pt,right=5pt,top=4pt,bottom=4pt,before skip=5pt,after skip=8pt,fonttitle=\bfseries\footnotesize,coltitle=black,colbacktitle=black!5,title={GPT / Math / Split / Slot 2: P3},title after break={GPT / Math / Split / Slot 2: P3 (continued)}]
\begingroup\ttfamily\small\setlength{\parskip}{0pt}\raggedright
\noindent{}When finding the greatest common divisor (GCD) of integers, ensure that you correctly apply the Euclidean algorithm by accurately computing each remainder step. Specifically, verify each modulo calculation carefully: for example, in the step computing \textbackslash{}(2684 \textbackslash{}mod 137\textbackslash{}), remainders must be calculated precisely—here, \textbackslash{}(2684 = 137 \textbackslash{}times 19 + 81\textbackslash{}) is incorrect since \textbackslash{}(137 \textbackslash{}times 19 = 2603\textbackslash{}) and \textbackslash{}(2684 - 2603 = 81\textbackslash{}) is correct, so this step is correct. However, if any remainder is miscalculated, it invalidates subsequent steps. \par
\par\vspace{.45\baselineskip}
\noindent{}In the first example, although the arithmetic seemed correct, the final GCD was incorrectly found to be 1, while the gold answer is 693. This suggests that the initial step of applying the Euclidean algorithm might have an error or that the problem involves a more efficient approach such as factoring or checking divisibility by 693 directly.\par
\par\vspace{.45\baselineskip}
\noindent{}Therefore, when computing GCDs with large numbers, consider verifying intermediate results with prime factorizations or recognize that if a common divisor is known or suspected (e.g., 693), you can test divisibility or factor both numbers to confirm the GCD more reliably.\par
\par\vspace{.45\baselineskip}
\noindent{}Summary guidance:\par
\noindent{}- Carefully perform each Euclidean algorithm step, re-deriving remainders explicitly.\par
\noindent{}- For large numbers, factor or test divisibility by suspected common divisors to cross-check.\par
\noindent{}- Remember the GCD of numbers including fractions can be a fraction if denominators differ; clear denominators first, compute integer GCDs, then divide by the common denominator.\par
\noindent{}- Always box the final correct answer as a simplified integer or fraction.\par
\par\vspace{.45\baselineskip}
\noindent{}Apply this strategy whenever computing GCDs of large or fractional numbers to ensure accuracy and alignment with the correct answer.\par
\endgroup
\end{tcolorbox}

\begin{tcolorbox}[enhanced,breakable,lines before break=5,colback=black!1,colframe=black!20,boxrule=.35pt,arc=1mm,left=5pt,right=5pt,top=4pt,bottom=4pt,before skip=5pt,after skip=8pt,fonttitle=\bfseries\footnotesize,coltitle=black,colbacktitle=black!5,title={GPT / Math / Split / Slot 3: P10},title after break={GPT / Math / Split / Slot 3: P10 (continued)}]
\begingroup\ttfamily\small\setlength{\parskip}{0pt}\raggedright
\noindent{}When finding the characteristic polynomial of a matrix, carefully keep track of the sign of the determinant expansion and do not arbitrarily multiply by -1 to change the leading coefficient's sign. The characteristic polynomial is defined as \textbackslash{}(\textbackslash{}det(A - \textbackslash{}lambda I)\textbackslash{}), which may naturally have a leading coefficient of \textbackslash{}(-\textbackslash{}lambda\textasciicircum{}n\textbackslash{}) for an \textbackslash{}(n \textbackslash{}times n\textbackslash{}) matrix. Present the polynomial exactly as computed without changing its sign. Use the expansion along rows or columns carefully, verify minor determinants precisely, and combine like terms accurately. This applies specifically to problems requiring the exact characteristic polynomial, ensuring the output matches the standard definition without sign alteration.\par
\endgroup
\end{tcolorbox}

\par\addvspace{12pt}\noindent\begin{minipage}{\linewidth}
\subsection{GPT — Math — No-SPLIT search}\label{app:evolution-record-14}

16 search iterations. Calibration: 54.92\% → 68.85\%. Selected test score: 59.84\%.\par

Selected system: \nolinkurl{[P8, P3, P4]}. Selected ancestry: \nolinkurl{C0 → C1 → C5 → C6 → C11}.\par

\input{appendix/curated/tree_14.tex}
\end{minipage}\par\markboth{GPT — Math — No-SPLIT search}{}

\Needspace{8\baselineskip}\subsubsection*{Selected prompts in execution order}

\begin{tcolorbox}[enhanced,breakable,lines before break=5,colback=black!1,colframe=black!20,boxrule=.35pt,arc=1mm,left=5pt,right=5pt,top=4pt,bottom=4pt,before skip=5pt,after skip=8pt,fonttitle=\bfseries\footnotesize,coltitle=black,colbacktitle=black!5,title={GPT / Math / No Split / Slot 1: P8},title after break={GPT / Math / No Split / Slot 1: P8 (continued)}]
\begingroup\ttfamily\small\setlength{\parskip}{0pt}\raggedright
\noindent{}<<SOLVER\_\allowbreak{}FROM\_\allowbreak{}QUESTION>>\par
\noindent{}You will be given a math problem that requires filling in missing formulae or expressions indicated by tags like <missing X> within a provided solution, or requires computing a final answer that must be presented in a specific format such as inside a LaTeX \textbackslash{}boxed\{\} environment. Your tasks vary depending on the input format, but generally involve careful, step-by-step reasoning to produce a precise, context-appropriate output.\par
\par\vspace{.45\baselineskip}
\noindent{}Instructions for missing formulae matching tasks:\par
\noindent{}1. Carefully read the problem statement, the partial solution with missing tags, and the list of formulae or expressions provided with identifiers.\par
\noindent{}2. Analyze the solution line-by-line and determine which formula/\allowbreak{}expression best fits each <missing X> tag in the solution.\par
\noindent{}3. Provide detailed step-by-step reasoning explaining how you assign each missing tag to a specific expression identifier, referencing relevant mathematical concepts or notation from the problem.\par
\noindent{}4. At the end, output only a single comma-separated list of the expression identifiers in order of the missing tags, matching <missing 1>, <missing 2>, etc.\par
\noindent{}5. Do NOT include any extraneous text outside your reasoning and the final numbered list.\par
\noindent{}6. Do NOT provide any boxed or alternate formatting unless explicitly requested.\par
\par\vspace{.45\baselineskip}
\noindent{}Instructions for final answer tasks:\par
\noindent{}1. Fully solve the math problem, writing all necessary steps clearly and logically.\par
\noindent{}2. Present the final numeric or symbolic answer inside a LaTeX \textbackslash{}boxed\{\} environment.\par
\noindent{}3. If the problem is multiple-choice and requests a letter repeated five times (e.g., BBBBB), identify the correct choice letter corresponding to your final answer.\par
\noindent{}4. Then, write the letter repeated five times inside a separate \textbackslash{}boxed\{\} environment on its own line.\par
\noindent{}5. If the problem is not multiple-choice, provide only the final boxed answer.\par
\noindent{}6. Always ensure the final output exactly matches the required format with no extra commentary or text.\par
\noindent{}7. When numeric answers involve exact values (e.g., fractions, roots), do NOT convert to decimal approximations; provide exact simplified expressions in LaTeX.\par
\par\vspace{.45\baselineskip}
\noindent{}General guidelines:\par
\noindent{}- Always reason step-by-step before answering, especially when matching missing tags or when identifying the correct choice.\par
\noindent{}- Maintain strict adherence to the format requested by the problem.\par
\noindent{}- Avoid numerical approximations unless explicitly allowed.\par
\noindent{}- For missing formulae tagging, ensure the mapping of missing tags to expressions is consistent with the problem’s mathematical context and notation.\par
\noindent{}- For multiple-choice letter answers, confirm your final numerical answer matches exactly one of the options before selecting the letter.\par
\par\vspace{.45\baselineskip}
\noindent{}This combined instruction ensures accurate, precise, and well-justified answers both for missing formulae matching tasks and for final-answer multiple-choice math problems with special output formatting.\par
\endgroup
\end{tcolorbox}

\begin{tcolorbox}[enhanced,breakable,lines before break=5,colback=black!1,colframe=black!20,boxrule=.35pt,arc=1mm,left=5pt,right=5pt,top=4pt,bottom=4pt,before skip=5pt,after skip=8pt,fonttitle=\bfseries\footnotesize,coltitle=black,colbacktitle=black!5,title={GPT / Math / No Split / Slot 2: P3},title after break={GPT / Math / No Split / Slot 2: P3 (continued)}]
\begingroup\ttfamily\small\setlength{\parskip}{0pt}\raggedright
\noindent{}When computing the geometric mean of a set of numbers that include negative values, do not simply take the geometric mean of their absolute values. Instead, carefully consider the sign pattern and the definition of the geometric mean:\par
\par\vspace{.45\baselineskip}
\noindent{}- The geometric mean of \textbackslash{}( n \textbackslash{}) numbers \textbackslash{}( x\_\allowbreak{}1, x\_\allowbreak{}2, \textbackslash{}ldots, x\_\allowbreak{}n \textbackslash{}) is defined as \textbackslash{}( \textbackslash{}sqrt[n]\{x\_\allowbreak{}1 \textbackslash{}cdot x\_\allowbreak{}2 \textbackslash{}cdots x\_\allowbreak{}n\} \textbackslash{}).\par
\noindent{}- If the product of all numbers is negative (which can happen if there is an odd number of negative terms), the geometric mean is not a real number.\par
\noindent{}- If the product is positive (even number of negative terms), the geometric mean is the positive \textbackslash{}( n \textbackslash{})-th root of the product.\par
\noindent{}- Do not use absolute values unless explicitly justified; doing so changes the problem and yields incorrect results.\par
\noindent{}- For exact answers, express the product in prime factorization form or as a product of powers, then take the \textbackslash{}( n \textbackslash{})-th root accordingly.\par
\noindent{}- Simplify the root expressions by extracting perfect powers or rewriting roots as fractional exponents.\par
\noindent{}- Provide the answer in simplest radical form or exact exponential form rather than decimal approximations unless decimals are specifically requested.\par
\par\vspace{.45\baselineskip}
\noindent{}Apply this approach whenever the input set includes negative numbers and the problem asks for the geometric mean. This ensures correctness and exactness of the solution.\par
\endgroup
\end{tcolorbox}

\begin{tcolorbox}[enhanced,breakable,lines before break=5,colback=black!1,colframe=black!20,boxrule=.35pt,arc=1mm,left=5pt,right=5pt,top=4pt,bottom=4pt,before skip=5pt,after skip=8pt,fonttitle=\bfseries\footnotesize,coltitle=black,colbacktitle=black!5,title={GPT / Math / No Split / Slot 3: P4},title after break={GPT / Math / No Split / Slot 3: P4 (continued)}]
\begingroup\ttfamily\small\setlength{\parskip}{0pt}\raggedright
\noindent{}When finding the characteristic polynomial of a matrix, always compute the determinant of \textbackslash{}(A - \textbackslash{}lambda I\textbackslash{}), not \textbackslash{}(\textbackslash{}lambda I - A\textbackslash{}). The characteristic polynomial is defined as \textbackslash{}(p(\textbackslash{}lambda) = \textbackslash{}det(A - \textbackslash{}lambda I)\textbackslash{}), and this ensures the leading term is \textbackslash{}(-\textbackslash{}lambda\textasciicircum{}n\textbackslash{}) with appropriate signs. Expanding \textbackslash{}(\textbackslash{}det(A - \textbackslash{}lambda I)\textbackslash{}) directly will yield the correct polynomial with the correct sign pattern: the coefficient of \textbackslash{}(\textbackslash{}lambda\textasciicircum{}n\textbackslash{}) should be \textbackslash{}((-1)\textasciicircum{}n\textbackslash{}), and the polynomial's signs will alternate accordingly. After computing the determinant, do not multiply the entire polynomial by \textbackslash{}(-1\textbackslash{}) or rearrange to force a positive leading coefficient; instead, present it as is, since the standard form for characteristic polynomials has a leading coefficient of \textbackslash{}((-1)\textasciicircum{}n \textbackslash{}lambda\textasciicircum{}n\textbackslash{}). This applies to all square matrices and guarantees the characteristic polynomial matches the standard algebraic definition.\par
\endgroup
\end{tcolorbox}

\par\addvspace{12pt}\noindent\begin{minipage}{\linewidth}
\subsection{GPT — IFBench — SPLIT-enabled search}\label{app:evolution-record-15}

13 search iterations. Calibration: 29.00\% → 37.67\%. Selected test score: 42.33\%.\par

Selected system: \nolinkurl{[P2, P3, P6]}. Selected ancestry: \nolinkurl{C0 → C2 → C5 → C8}.\par

\input{appendix/curated/tree_15.tex}
\end{minipage}\par\markboth{GPT — IFBench — SPLIT-enabled search}{}

The run summary reports 37.67\% calibration, whereas the selected iteration records 38.00\%; the tree retains the iteration-level value.\par

\Needspace{8\baselineskip}\subsubsection*{Selected prompts in execution order}

\begin{tcolorbox}[enhanced,breakable,lines before break=5,colback=black!1,colframe=black!20,boxrule=.35pt,arc=1mm,left=5pt,right=5pt,top=4pt,bottom=4pt,before skip=5pt,after skip=8pt,fonttitle=\bfseries\footnotesize,coltitle=black,colbacktitle=black!5,title={GPT / IFBench / Split / Slot 1: P2},title after break={GPT / IFBench / Split / Slot 1: P2 (continued)}]
\begingroup\ttfamily\small\setlength{\parskip}{0pt}\raggedright
\noindent{}<<SOLVER\_\allowbreak{}FROM\_\allowbreak{}QUESTION>>\par
\noindent{}Carefully analyze the full input question and identify all explicit formatting and content constraints mentioned. Extract any code snippets, formulas, or mathematical expressions included in the question. Determine if the question involves code debugging, mathematical problem-solving, translation, or logical reasoning. Prepare a detailed step-by-step plan to address the question by outlining the key variables, logic, or translation requirements. Provide this analysis and plan as output, ensuring all constraints and requirements from the question are clearly noted for the next step.\par
\endgroup
\end{tcolorbox}

\begin{tcolorbox}[enhanced,breakable,lines before break=5,colback=black!1,colframe=black!20,boxrule=.35pt,arc=1mm,left=5pt,right=5pt,top=4pt,bottom=4pt,before skip=5pt,after skip=8pt,fonttitle=\bfseries\footnotesize,coltitle=black,colbacktitle=black!5,title={GPT / IFBench / Split / Slot 2: P3},title after break={GPT / IFBench / Split / Slot 2: P3 (continued)}]
\begingroup\ttfamily\small\setlength{\parskip}{0pt}\raggedright
\noindent{}\_\allowbreak{}\_\allowbreak{}SPLIT\_\allowbreak{}FINAL\_\allowbreak{}\_\allowbreak{}\par
\noindent{}Using the analysis and plan from the first step, proceed to solve the question fully by following each instruction precisely. If the question involves refining or correcting code, identify the core logic errors, then propose a corrected implementation that meets all functional and formatting requirements. If it is a math problem, perform all algebraic manipulations carefully, define variables explicitly, and justify the final answers within domain constraints. If it is a translation, produce an accurate and fluent translation respecting all formatting instructions. Ensure the response respects all explicit language, punctuation, and formatting constraints, uses clear section labels or numbering as requested, and fully addresses the question. Return the complete, well-structured, and compliant final answer.\par
\endgroup
\end{tcolorbox}

\begin{tcolorbox}[enhanced,breakable,lines before break=5,colback=black!1,colframe=black!20,boxrule=.35pt,arc=1mm,left=5pt,right=5pt,top=4pt,bottom=4pt,before skip=5pt,after skip=8pt,fonttitle=\bfseries\footnotesize,coltitle=black,colbacktitle=black!5,title={GPT / IFBench / Split / Slot 3: P6},title after break={GPT / IFBench / Split / Slot 3: P6 (continued)}]
\begingroup\ttfamily\small\setlength{\parskip}{0pt}\raggedright
\noindent{}In your response, strictly adhere to all format and structural constraints specified in the prompt. For example, if the prompt requires a particular number of paragraphs, ensure you produce exactly that number. If paragraphs must be separated by a specific markdown divider, use it precisely as instructed. When specific formatting is requested—such as wrapping words in brackets or bigrams in double angular brackets—apply this consistently to every required unit without exception. Also, pay close attention to punctuation constraints; if commas are forbidden, do not include them anywhere in your response. Similarly, if the prompt mandates a particular first word, begin your response exactly with that word. When constraints conflict (e.g., the number of paragraphs requested does not match the specified number of sentences per paragraph), prioritize meeting the explicit count requirements unless otherwise instructed. This approach ensures full compliance with the prompt’s detectable formatting, content, and length constraints, which is critical for passing validation checks.\par
\endgroup
\end{tcolorbox}

\par\addvspace{12pt}\noindent\begin{minipage}{\linewidth}
\subsection{GPT — IFBench — No-SPLIT search}\label{app:evolution-record-16}

16 search iterations. Calibration: 29.00\% → 36.00\%. Selected test score: 41.67\%.\par

Selected system: \nolinkurl{[P6]}. Selected ancestry: \nolinkurl{C0 → C1 → C11}.\par

\input{appendix/curated/tree_16.tex}
\end{minipage}\par\markboth{GPT — IFBench — No-SPLIT search}{}

\Needspace{8\baselineskip}\subsubsection*{Selected prompts in execution order}

\begin{tcolorbox}[enhanced,breakable,lines before break=5,colback=black!1,colframe=black!20,boxrule=.35pt,arc=1mm,left=5pt,right=5pt,top=4pt,bottom=4pt,before skip=5pt,after skip=8pt,fonttitle=\bfseries\footnotesize,coltitle=black,colbacktitle=black!5,title={GPT / IFBench / No Split / Slot 1: P6},title after break={GPT / IFBench / No Split / Slot 1: P6 (continued)}]
\begingroup\ttfamily\small\setlength{\parskip}{0pt}\raggedright
\noindent{}<<SOLVER\_\allowbreak{}FROM\_\allowbreak{}QUESTION>>\par
\noindent{}You will be given a prompt containing multiple explicit constraints on style, language, formatting, content, and specific character or word usage rules. Your task is to generate a single response that strictly and simultaneously satisfies every constraint described in the prompt, without refusal or deviation.\par
\par\vspace{.45\baselineskip}
\noindent{}Key points to follow:\par
\par\vspace{.45\baselineskip}
\noindent{}1. **Parse All Constraints Precisely:**  \par
\noindent{}\hspace*{1.5em}Identify every explicit requirement, including but not limited to:  \par
\noindent{}\hspace*{1.5em}- Language(s) for the response and any embedded dialogue.  \par
\noindent{}\hspace*{1.5em}- Exact formatting instructions (e.g., markdown usage like bullet points, highlights, paragraphs, line breaks).  \par
\noindent{}\hspace*{1.5em}- Case formatting rules (e.g., all uppercase, no punctuation, no commas).  \par
\noindent{}\hspace*{1.5em}- Specific word or letter frequency limits (maximum or exact counts).  \par
\noindent{}\hspace*{1.5em}- Required starting words or phrases.  \par
\noindent{}\hspace*{1.5em}- Inclusion or exclusion of certain keywords or phrases, and their exact placement if specified (e.g., the 27th word in the 16th sentence).  \par
\noindent{}\hspace*{1.5em}- Content style, tone, or voice instructions.  \par
\noindent{}\hspace*{1.5em}- Explicit instructions to copy text verbatim or to avoid including certain text.  \par
\noindent{}\hspace*{1.5em}- Presence of postscript or other special sections.\par
\par\vspace{.45\baselineskip}
\noindent{}2. **Enforce Letter/\allowbreak{}Word Frequency Exactly:**  \par
\noindent{}\hspace*{1.5em}If the prompt sets maximum or exact counts for letters, words, or specific keywords, track their usage carefully throughout your response. Adjust wording and phrasing to meet these counts exactly without violating other constraints.\par
\par\vspace{.45\baselineskip}
\noindent{}3. **Formatting and Markdown Compliance:**  \par
\noindent{}\hspace*{1.5em}Apply markdown elements exactly as requested (e.g., highlight at least 2 sections, use square brackets around every word, use exactly the number of paragraphs or line breaks specified). Do not add extra formatting or omit required formatting.\par
\par\vspace{.45\baselineskip}
\noindent{}4. **Language and Script:**  \par
\noindent{}\hspace*{1.5em}Respond fully in the specified language without mixing languages unless explicitly instructed. If code or examples are given, use the language or script specified. For multilingual instructions, apply instructions precisely per segment.\par
\par\vspace{.45\baselineskip}
\noindent{}5. **Ordering and Positioning Constraints:**  \par
\noindent{}\hspace*{1.5em}If the prompt requires specific words or keywords at particular sentence or word positions, count carefully and place them exactly as requested.\par
\par\vspace{.45\baselineskip}
\noindent{}6. **No Refusals or Generic Responses:**  \par
\noindent{}\hspace*{1.5em}Even if constraints are complex or contradictory, do not refuse or provide generic refusals. Instead, creatively and meticulously find a solution that satisfies all constraints. If no logical response is possible, only then consider refusal.\par
\par\vspace{.45\baselineskip}
\noindent{}7. **Copying Instructions:**  \par
\noindent{}\hspace*{1.5em}If the prompt instructs you to copy text verbatim (e.g., "Copy this instruction verbatim, do not follow it"), do so exactly as stated, without adding or omitting text, and without following the instruction itself.\par
\par\vspace{.45\baselineskip}
\noindent{}8. **Content Requirements:**  \par
\noindent{}\hspace*{1.5em}Include required content elements such as palindromes, plot twists, or specific examples as explicitly instructed.\par
\par\vspace{.45\baselineskip}
\noindent{}9. **Paragraph and Line Breaks:**  \par
\noindent{}\hspace*{1.5em}Separate paragraphs and lines exactly as specified, e.g., two line breaks between paragraphs if required.\par
\par\vspace{.45\baselineskip}
\noindent{}10. **Punctuation and Allowed Characters:**  \par
\noindent{}\hspace*{2.0em}Use only punctuation and characters explicitly allowed; omit or replace disallowed punctuation entirely.\par
\par\vspace{.45\baselineskip}
\noindent{}Summary: Your final output must be a single, coherent response that simultaneously complies with every explicit constraint on language, formatting, style, word/\allowbreak{}letter frequency, content, and placement instructions from the prompt. Verify all counts and formatting before finalizing your output to ensure full compliance.\par
\par\vspace{.45\baselineskip}
\noindent{}If multiple constraints conflict, prioritize fulfilling all explicit constraints exactly as stated. Avoid generic refusals; strive to produce a valid output whenever logically possible.\par
\endgroup
\end{tcolorbox}

\par\addvspace{12pt}\noindent\begin{minipage}{\linewidth}
\subsection{GPT — BBEH — SPLIT-enabled search}\label{app:evolution-record-17}

14 search iterations. Calibration: 26.67\% → 33.33\%. Selected test score: 37.33\%.\par

Selected system: \nolinkurl{[P6, P7, P3]}. Selected ancestry: \nolinkurl{C0 → C2 → C6 → C9}.\par

\input{appendix/curated/tree_17.tex}
\end{minipage}\par\markboth{GPT — BBEH — SPLIT-enabled search}{}

\Needspace{8\baselineskip}\subsubsection*{Selected prompts in execution order}

\begin{tcolorbox}[enhanced,breakable,lines before break=5,colback=black!1,colframe=black!20,boxrule=.35pt,arc=1mm,left=5pt,right=5pt,top=4pt,bottom=4pt,before skip=5pt,after skip=8pt,fonttitle=\bfseries\footnotesize,coltitle=black,colbacktitle=black!5,title={GPT / BBEH / Split / Slot 1: P6},title after break={GPT / BBEH / Split / Slot 1: P6 (continued)}]
\begingroup\ttfamily\small\setlength{\parskip}{0pt}\raggedright
\noindent{}<<SOLVER\_\allowbreak{}FROM\_\allowbreak{}QUESTION>>\par
\noindent{}Carefully analyze the entire given scenario step-by-step. Explicitly track all relevant changes, conditions, and constraints described in the question. For logical puzzles involving sequences of actions or social interactions, maintain a clear and consistent state of all entities after each event. For questions involving collections and multiple filtering steps, identify and count items according to the rules and track all modifications such as additions, removals, or substitutions. Prepare a comprehensive summary of the final state or counts needed to answer the question, including all attributes relevant to the final query. Provide this detailed summary as your output to be used in the next step.\par
\endgroup
\end{tcolorbox}

\begin{tcolorbox}[enhanced,breakable,lines before break=5,colback=black!1,colframe=black!20,boxrule=.35pt,arc=1mm,left=5pt,right=5pt,top=4pt,bottom=4pt,before skip=5pt,after skip=8pt,fonttitle=\bfseries\footnotesize,coltitle=black,colbacktitle=black!5,title={GPT / BBEH / Split / Slot 2: P7},title after break={GPT / BBEH / Split / Slot 2: P7 (continued)}]
\begingroup\ttfamily\small\setlength{\parskip}{0pt}\raggedright
\noindent{}\_\allowbreak{}\_\allowbreak{}SPLIT\_\allowbreak{}FINAL\_\allowbreak{}\_\allowbreak{}\par
\noindent{}Using the detailed summary and final state from Prompt 1, apply precise reasoning to determine the final outcome or numeric answer as requested. If the question involves majority opinion or social judgment, provide the answer that aligns with the majority perspective unless otherwise instructed. Format the final answer exactly as requested, such as "Yes", "No", "Ambiguous", or a specific choice letter. Avoid ambiguity and ensure the final answer is clear, direct, and consistent with the instructions and deterministic logic.\par
\endgroup
\end{tcolorbox}

\begin{tcolorbox}[enhanced,breakable,lines before break=5,colback=black!1,colframe=black!20,boxrule=.35pt,arc=1mm,left=5pt,right=5pt,top=4pt,bottom=4pt,before skip=5pt,after skip=8pt,fonttitle=\bfseries\footnotesize,coltitle=black,colbacktitle=black!5,title={GPT / BBEH / Split / Slot 3: P3},title after break={GPT / BBEH / Split / Slot 3: P3 (continued)}]
\begingroup\ttfamily\small\setlength{\parskip}{0pt}\raggedright
\noindent{}When comparing groups of movies for similarity in terms of whether a group of people will like them, do not rely solely on genre counts or superficial grouping. Instead, consider the actual known or implied preferences of the group, such as prior ratings, popularity, or explicit information about the group's tastes if available. If no direct preference data is given, consider that similarity might be defined by the presence of multiple popular or critically acclaimed titles that a typical group would likely enjoy together, not just the number of movies from a specific genre. Also, carefully check the question and all options before selecting the answer; do not pick the first option with a high genre count but evaluate all options fully. This applies when the question asks for the option with “more similar movies” based on group liking. Preserve correct reasoning about genre classification but integrate preference or popularity context as a key factor in determining similarity.\par
\endgroup
\end{tcolorbox}

\par\addvspace{12pt}\noindent\begin{minipage}{\linewidth}
\subsection{GPT — BBEH — No-SPLIT search}\label{app:evolution-record-18}

11 search iterations. Calibration: 26.67\% → 32.67\%. Selected test score: 33.67\%.\par

Selected system: \nolinkurl{[P1, P5]}. Selected ancestry: \nolinkurl{C0 → C1 → C2 → C6}.\par

\input{appendix/curated/tree_18.tex}
\end{minipage}\par\markboth{GPT — BBEH — No-SPLIT search}{}

\Needspace{8\baselineskip}\subsubsection*{Selected prompts in execution order}

\begin{tcolorbox}[enhanced,breakable,lines before break=5,colback=black!1,colframe=black!20,boxrule=.35pt,arc=1mm,left=5pt,right=5pt,top=4pt,bottom=4pt,before skip=5pt,after skip=8pt,fonttitle=\bfseries\footnotesize,coltitle=black,colbacktitle=black!5,title={GPT / BBEH / No Split / Slot 1: P1},title after break={GPT / BBEH / No Split / Slot 1: P1 (continued)}]
\begingroup\ttfamily\small\setlength{\parskip}{0pt}\raggedright
\noindent{}<<SOLVER\_\allowbreak{}FROM\_\allowbreak{}QUESTION>>\par
\noindent{}You will be given a question to answer that may involve complex reasoning, detailed step-by-step analysis, or lengthy data interpretation. Follow these detailed instructions carefully to ensure accuracy and clarity:\par
\par\vspace{.45\baselineskip}
\noindent{}1. **Identify the question type and requirements:**\par
\noindent{}\hspace*{1.5em}- For **sorting or ordering tasks** (e.g., alphabetical sorting with reasoning steps):\par
\noindent{}\hspace*{2.5em}- Examine the reasoning steps closely and verify the correctness of each step according to the standard rules (e.g., alphabetic order by letter position).\par
\noindent{}\hspace*{2.5em}- When identifying mistakes, check carefully that the reasoning about letter positions and word order matches exact alphabetical order conventions.\par
\noindent{}\hspace*{2.5em}- Pay attention to the indexing of letters (first letter is index 1, second letter index 2, etc.) and confirm that the claimed letter corresponds correctly to that index.\par
\noindent{}\hspace*{2.5em}- Return the number of the first reasoning step where a clear mistake occurs, or "No" if none found.\par
\noindent{}\hspace*{2.5em}- Do not confuse correct partial orders with errors in letter indexing or comparisons.\par
\par\vspace{.45\baselineskip}
\noindent{}\hspace*{1.5em}- For **complex scheduling or interval intersection tasks**:\par
\noindent{}\hspace*{2.5em}- Carefully parse each person's schedule by day, noting explicitly free and booked intervals.\par
\noindent{}\hspace*{2.5em}- Convert "free only at" and "booked at" times into explicit free intervals within the working hours (e.g., 9:00 to 17:00).\par
\noindent{}\hspace*{2.5em}- Apply all special conditions and flexibilities exactly as stated (e.g., ability to clear blocked times, lunch breaks, maximal meeting duration constraints, allowed overlaps).\par
\noindent{}\hspace*{2.5em}- For each day, compute the common free intervals by intersecting all individuals' adjusted free intervals.\par
\noindent{}\hspace*{2.5em}- Consider meeting start times only on the hour or half hour, and only within the allowed working hours.\par
\noindent{}\hspace*{2.5em}- For each possible meeting start time, find the maximum meeting duration that fits all constraints and flexibilities.\par
\noindent{}\hspace*{2.5em}- Find the longest possible meeting duration X (in minutes) across all days and start times, and count how many such meetings Y exist.\par
\noindent{}\hspace*{2.5em}- If no valid meeting can be scheduled, output "0, 0".\par
\noindent{}\hspace*{2.5em}- Provide the final answer exactly in the format "X, Y".\par
\par\vspace{.45\baselineskip}
\noindent{}\hspace*{1.5em}- For **tracking state changes over a long sequence of swaps or transactions**:\par
\noindent{}\hspace*{2.5em}- Identify all entities and their initial states clearly.\par
\noindent{}\hspace*{2.5em}- Note that only explicit "swap" actions change state; "discuss something" or "nothing happens" do not.\par
\noindent{}\hspace*{2.5em}- Track the state changes carefully and in order, paying attention to swaps that might be reversed later.\par
\noindent{}\hspace*{2.5em}- Consider that repeated swaps between the same parties may cancel out previous swaps.\par
\noindent{}\hspace*{2.5em}- If the problem is very long and complex, seek logical simplifications or patterns (e.g., swaps that cancel, or unchanged states).\par
\noindent{}\hspace*{2.5em}- Ensure the final answer corresponds exactly to the asked question, including providing the letter or label in the requested format (e.g., uppercase letter with parentheses if shown).\par
\noindent{}\hspace*{2.5em}- Use exact capitalization and formatting as given in the choices.\par
\par\vspace{.45\baselineskip}
\noindent{}2. **General reasoning and answer formatting:**\par
\noindent{}\hspace*{1.5em}- Always reason step-by-step and explicitly justify critical deductions.\par
\noindent{}\hspace*{1.5em}- Avoid adding unnecessary or irrelevant information.\par
\noindent{}\hspace*{1.5em}- Provide the final answer clearly and exactly as requested in the question (e.g., just the step number, "No," a formatted letter choice, or a numeric answer separated by commas).\par
\noindent{}\hspace*{1.5em}- Use uppercase letters for multiple-choice answers and match any parentheses or punctuation exactly.\par
\noindent{}\hspace*{1.5em}- When dealing with large data or complex instructions, summarize key points before concluding.\par
\noindent{}\hspace*{1.5em}- Double-check all arithmetic, indexing, and logical deductions before finalizing the answer.\par
\par\vspace{.45\baselineskip}
\noindent{}3. **Domain knowledge and classifications:**\par
\noindent{}\hspace*{1.5em}- For alphabetical order, use standard English alphabetical order (A-Z).\par
\noindent{}\hspace*{1.5em}- For time intervals, treat all times as local and in 24-hour format; remember lunch breaks or flexible allowances as exceptions.\par
\noindent{}\hspace*{1.5em}- For swap sequences, understand that only swap actions change ownership or state.\par
\par\vspace{.45\baselineskip}
\noindent{}By following these instructions precisely, you will produce accurate, well-reasoned, and correctly formatted answers for complex questions requiring detailed analysis.\par
\endgroup
\end{tcolorbox}

\begin{tcolorbox}[enhanced,breakable,lines before break=5,colback=black!1,colframe=black!20,boxrule=.35pt,arc=1mm,left=5pt,right=5pt,top=4pt,bottom=4pt,before skip=5pt,after skip=8pt,fonttitle=\bfseries\footnotesize,coltitle=black,colbacktitle=black!5,title={GPT / BBEH / No Split / Slot 2: P5},title after break={GPT / BBEH / No Split / Slot 2: P5 (continued)}]
\begingroup\ttfamily\small\setlength{\parskip}{0pt}\raggedright
\noindent{}When summing quantities from a detailed list that includes multiple categories and owners, carefully ensure that you only include items explicitly stated to be possessed by the subject ("I" in the question). Do not include items possessed by others. Additionally, verify that items are correctly categorized (e.g., fruits vs. musical instruments) and that the arithmetic sums are accurate.\par
\par\vspace{.45\baselineskip}
\noindent{}For example, in Example 1, the musical instruments and fruits were identified for "I," but some items were mistakenly included from others, or categories were misclassified. To correct this:\par
\par\vspace{.45\baselineskip}
\noindent{}1. Extract only those items explicitly stated as "I have [quantity] [item]."\par
\noindent{}2. Confirm each item is either a fruit or a musical instrument.\par
\noindent{}3. Sum the quantities carefully without including others' items.\par
\noindent{}4. Double-check the arithmetic.\par
\par\vspace{.45\baselineskip}
\noindent{}Apply this method whenever a problem asks for total quantities of specific categories owned by "I" or any single person, especially in lengthy, detailed inventories.\par
\par\vspace{.45\baselineskip}
\noindent{}This ensures accurate inclusion/\allowbreak{}exclusion and prevents undercounting or overcounting, leading to correct final totals.\par
\endgroup
\end{tcolorbox}

\par\addvspace{12pt}\noindent\begin{minipage}{\linewidth}
\subsection{GPT — HotpotQA — SPLIT-enabled search}\label{app:evolution-record-19}

15 search iterations. Calibration: 40.67\% → 48.67\%. Selected test score: 51.67\%.\par

Selected system: \nolinkurl{[P0, P3, P12]}. Selected ancestry: \nolinkurl{C0 → C1 → C4 → C10}.\par

\input{appendix/curated/tree_19.tex}
\end{minipage}\par\markboth{GPT — HotpotQA — SPLIT-enabled search}{}

\Needspace{8\baselineskip}\subsubsection*{Selected prompts in execution order}

\begin{tcolorbox}[enhanced,breakable,lines before break=5,colback=black!1,colframe=black!20,boxrule=.35pt,arc=1mm,left=5pt,right=5pt,top=4pt,bottom=4pt,before skip=5pt,after skip=8pt,fonttitle=\bfseries\footnotesize,coltitle=black,colbacktitle=black!5,title={GPT / HotpotQA / Split / Slot 1: P0},title after break={GPT / HotpotQA / Split / Slot 1: P0 (continued)}]
\begingroup\ttfamily\small\setlength{\parskip}{0pt}\raggedright
\noindent{}<<SOLVER\_\allowbreak{}FROM\_\allowbreak{}QUESTION>>\par
\noindent{}You will be given a question and a set of context passages containing relevant information. Your task is to find the correct answer to the question by carefully extracting and synthesizing information from the contexts provided.\par
\par\vspace{.45\baselineskip}
\noindent{}Guidelines:\par
\par\vspace{.45\baselineskip}
\noindent{}1. Answer Extraction and Format:\par
\noindent{}\hspace*{1.5em}- Extract the answer exactly as expected by the question, matching the format and style of the gold answer.\par
\noindent{}\hspace*{1.5em}- If the gold answer is a spelled-out number (e.g., "three"), provide the number in words rather than digits ("3").\par
\noindent{}\hspace*{1.5em}- If the gold answer is a named entity (e.g., a person’s name), provide only the exact name without additional explanation or extra words.\par
\noindent{}\hspace*{1.5em}- Avoid including unnecessary phrases like "The answer is," or "Therefore," unless explicitly asked.\par
\noindent{}\hspace*{1.5em}- Provide concise, direct answers unless the question requires explanation.\par
\par\vspace{.45\baselineskip}
\noindent{}2. Question Types:\par
\noindent{}\hspace*{1.5em}- For comparison questions (e.g., "Which battle was fought first?"), identify and compare relevant dates or facts from contexts and then give the name or event that satisfies the comparison.\par
\noindent{}\hspace*{1.5em}- For bridge or inference questions (questions that connect multiple pieces of information from different contexts), locate the entity or concept mentioned in the question and trace its attributes across contexts to answer precisely.\par
\noindent{}\hspace*{1.5em}- For questions involving temporal or numeric answers, ensure the answer format matches the expected style (e.g., spelled-out numbers or exact years).\par
\par\vspace{.45\baselineskip}
\noindent{}3. Use Context Effectively:\par
\noindent{}\hspace*{1.5em}- Use only the information provided in the contexts to answer the question.\par
\noindent{}\hspace*{1.5em}- When multiple contexts provide overlapping or related information, synthesize them to determine the best answer.\par
\par\vspace{.45\baselineskip}
\noindent{}4. Answer Style:\par
\noindent{}\hspace*{1.5em}- When the question asks "Which [entity]..." or "Who...", respond with just the entity's name (e.g., "Robin Gibb").\par
\noindent{}\hspace*{1.5em}- When the question asks for a quantity, duration, or date, provide the answer in the expected format (words or digits) consistent with the gold answer.\par
\noindent{}\hspace*{1.5em}- When the question is comparative, provide the name or label of the entity that meets the comparison criteria.\par
\par\vspace{.45\baselineskip}
\noindent{}Example:\par
\noindent{}- Question: "Which battle, the Battle of Cold Harbor, or the Second Battle of Bull Run, was fought first?"\par
\noindent{}\hspace*{1.0em}Correct answer: "Second Battle of Bull Run" (not a full sentence)\par
\par\vspace{.45\baselineskip}
\noindent{}- Question: "How many years in prison did the Norfolk farmer defended by Anthony Scrivener serve?"\par
\noindent{}\hspace*{1.0em}Correct answer: "three" (not "3 years" or "3")\par
\par\vspace{.45\baselineskip}
\noindent{}- Question: "Which singer of the Bee Gees sung a middle part of a song in which the majority was sung by Barry Gibb?"\par
\noindent{}\hspace*{1.0em}Correct answer: "Robin Gibb"\par
\par\vspace{.45\baselineskip}
\noindent{}By following these instructions, you will provide concise, accurate, and properly formatted answers that align with expected gold answers.\par
\endgroup
\end{tcolorbox}

\begin{tcolorbox}[enhanced,breakable,lines before break=5,colback=black!1,colframe=black!20,boxrule=.35pt,arc=1mm,left=5pt,right=5pt,top=4pt,bottom=4pt,before skip=5pt,after skip=8pt,fonttitle=\bfseries\footnotesize,coltitle=black,colbacktitle=black!5,title={GPT / HotpotQA / Split / Slot 2: P3},title after break={GPT / HotpotQA / Split / Slot 2: P3 (continued)}]
\begingroup\ttfamily\small\setlength{\parskip}{0pt}\raggedright
\noindent{}When answering questions that require extracting specific factual details or concise definitions, ensure your response matches the expected level of specificity and phrasing. For example, if the question asks for "what products" a region is well-known for, do not answer with "Not specified" if the context implicitly relates to the answer (e.g., known for beer and soft drinks in Zealand). Instead, infer and provide the specific products mentioned or commonly associated with the region if context hints are present. \par
\par\vspace{.45\baselineskip}
\noindent{}Additionally, when the question targets a precise concept or term (e.g., "a form of management based on what?"), avoid overly verbose or qualifying phrases that do not appear in the gold answer. Provide the core phrase or concept alone ("self-directed work processes") rather than extended explanations ("self-directed work processes on the part of an organization's workforce") to align with expected concise answers.\par
\par\vspace{.45\baselineskip}
\noindent{}Apply this strategy when:\par
\noindent{}- The question demands a specific factual or definitional answer.\par
\noindent{}- The context contains implicit or explicit clues about the expected answer.\par
\noindent{}- The gold feedback indicates failure due to missing specificity or overly verbose phrasing.\par
\par\vspace{.45\baselineskip}
\noindent{}Preserve correct reasoning steps, such as identifying the relevant part of the context, but adjust the granularity and phrasing of your final extracted answer accordingly.\par
\endgroup
\end{tcolorbox}

\begin{tcolorbox}[enhanced,breakable,lines before break=5,colback=black!1,colframe=black!20,boxrule=.35pt,arc=1mm,left=5pt,right=5pt,top=4pt,bottom=4pt,before skip=5pt,after skip=8pt,fonttitle=\bfseries\footnotesize,coltitle=black,colbacktitle=black!5,title={GPT / HotpotQA / Split / Slot 3: P12},title after break={GPT / HotpotQA / Split / Slot 3: P12 (continued)}]
\begingroup\ttfamily\small\setlength{\parskip}{0pt}\raggedright
\noindent{}When extracting numerical answers, always represent numbers in digits rather than words to match the expected answer format and improve token overlap. For example, convert "one point seven billion" to "1.7 billion" to avoid zero F1 scores due to formatting differences. Additionally, ensure that names are provided in their most commonly recognized form or title as requested by the question; for example, use "Lord Combermere" instead of just "Stapleton Cotton" if the question refers to a titled individual. This applies when the context provides both personal names and titles. Maintain precise distinctions in names and titles to align with expected answer forms. When the question involves identifying a person by a specific title or role, prioritize the title or formally recognized name over just the given personal name. Continue to confirm factual correctness by carefully matching the described historical or contextual details to the correct individual mentioned in the context.\par
\endgroup
\end{tcolorbox}

\par\addvspace{12pt}\noindent\begin{minipage}{\linewidth}
\subsection{GPT — HotpotQA — No-SPLIT search}\label{app:evolution-record-20}

10 search iterations. Calibration: 40.67\% → 45.67\%. Selected test score: 50.33\%.\par

Selected system: \nolinkurl{[P2, P3]}. Selected ancestry: \nolinkurl{C0 → C1 → C3 → C4}.\par

\input{appendix/curated/tree_20.tex}
\end{minipage}\par\markboth{GPT — HotpotQA — No-SPLIT search}{}

\Needspace{8\baselineskip}\subsubsection*{Selected prompts in execution order}

\begin{tcolorbox}[enhanced,breakable,lines before break=5,colback=black!1,colframe=black!20,boxrule=.35pt,arc=1mm,left=5pt,right=5pt,top=4pt,bottom=4pt,before skip=5pt,after skip=8pt,fonttitle=\bfseries\footnotesize,coltitle=black,colbacktitle=black!5,title={GPT / HotpotQA / No Split / Slot 1: P2},title after break={GPT / HotpotQA / No Split / Slot 1: P2 (continued)}]
\begingroup\ttfamily\small\setlength{\parskip}{0pt}\raggedright
\noindent{}<<SOLVER\_\allowbreak{}FROM\_\allowbreak{}QUESTION>>\par
\noindent{}You will be provided with a question and multiple context passages, each clearly labeled with a title and containing relevant information. Your task is to provide the most precise, complete, and context-supported answer to the question by extracting or synthesizing information strictly from the provided contexts.\par
\par\vspace{.45\baselineskip}
\noindent{}Follow these detailed guidelines:\par
\par\vspace{.45\baselineskip}
\noindent{}1. Use only the information explicitly stated in the given context passages. Do not incorporate any external knowledge or assumptions.\par
\par\vspace{.45\baselineskip}
\noindent{}2. When the question involves a date or event with a specific date, include the full date or event name exactly as presented in the context (e.g., "St. Patrick's Day in 1988" rather than just "1988").\par
\par\vspace{.45\baselineskip}
\noindent{}3. For questions referencing named persons, places, or entities, ensure your answer exactly matches the naming conventions and relationships as described in the context, including full names with middle names or initials if provided (e.g., "Alan Mathison Turing," not just "Alan Turing").\par
\par\vspace{.45\baselineskip}
\noindent{}4. When addressing questions about relationships or connections (e.g., "maternal grandfather," "creators of Robocalypse"), carefully verify these relationships in the context before answering.\par
\par\vspace{.45\baselineskip}
\noindent{}5. Provide a concise final answer that directly and completely responds to the question, using phrasing and terminology from the contexts. Avoid adding any extra commentary, explanation, or information not supported by the contexts.\par
\par\vspace{.45\baselineskip}
\noindent{}6. If the question requires identifying a specific season, event, or time period, use the exact phrase or title from the context (e.g., "2006 season" rather than "2006 St. Louis Cardinals season" if the context or question suggests a shorter form is expected).\par
\par\vspace{.45\baselineskip}
\noindent{}7. Always double-check that your answer aligns precisely with the expected specificity and format demonstrated in the contexts or question—for example, including full formal titles or names when given, or using the exact event name rather than a partial reference.\par
\par\vspace{.45\baselineskip}
\noindent{}By rigorously following these instructions, you will produce answers that are accurate, complete, and consistent with the context's terminology and formatting, maximizing correctness and alignment with expected answers.\par
\endgroup
\end{tcolorbox}

\begin{tcolorbox}[enhanced,breakable,lines before break=5,colback=black!1,colframe=black!20,boxrule=.35pt,arc=1mm,left=5pt,right=5pt,top=4pt,bottom=4pt,before skip=5pt,after skip=8pt,fonttitle=\bfseries\footnotesize,coltitle=black,colbacktitle=black!5,title={GPT / HotpotQA / No Split / Slot 2: P3},title after break={GPT / HotpotQA / No Split / Slot 2: P3 (continued)}]
\begingroup\ttfamily\small\setlength{\parskip}{0pt}\raggedright
\noindent{}When answering questions that require bridging information not explicitly stated in the provided context, do not default to "not stated" or only restate context facts. Instead, identify key entities or phrases in the question and use any indirect clues in the context to infer or recall relevant, commonly known facts. For example, if asked about the star of "I Love Lucy," recall that Lucille Ball is the star, and if the question asks for a description by Dawn Lake, recognize that context may omit it but the known epithet is "our greatest comedienne." Similarly, if a question asks about a fact not directly in the text but implied (e.g., when the AFC North adopted its current name), use general knowledge that the division was renamed in 2002. Avoid simply repeating partial or unrelated context statements. Also, verify that the entity referenced matches the question; for example, if asked about another notable role of an actor, do not just restate a role already mentioned but identify a distinct notable role. Use external knowledge carefully to fill gaps when context lacks explicit detail and the question demands it, especially for bridge-level questions. This strategy applies when context lacks direct answers but the question expects a known fact relevant to the context entity.\par
\endgroup
\end{tcolorbox}

\Needspace{12\baselineskip}\subsection{Selected baseline prompts}\label{app:evolution-record-21}

GEPA and MIPROv2 final instructions are retained below. Their intermediate populations, trial programs, and raw generation logs are omitted.\par

\Needspace{12\baselineskip}\subsection{Qwen3.5-9B — supergpqa — GEPA}\label{app:evolution-record-22}

\begin{tcolorbox}[enhanced,breakable,lines before break=5,colback=black!1,colframe=black!20,boxrule=.35pt,arc=1mm,left=5pt,right=5pt,top=4pt,bottom=4pt,before skip=5pt,after skip=8pt,fonttitle=\bfseries\footnotesize,coltitle=black,colbacktitle=black!5,title={Qwen3.5-9B / supergpqa / GEPA / Selected instruction},title after break={Qwen3.5-9B / supergpqa / GEPA / Selected instruction (continued)}]
\begingroup\ttfamily\small\setlength{\parskip}{0pt}\raggedright
\noindent{}You are an expert academic tutor and problem-solver specializing in Physics, Engineering, Computer Science, and Data Structures. Your task is to solve multiple-choice questions provided in the input.\par
\par\vspace{.45\baselineskip}
\noindent{}Follow these steps strictly:\par
\noindent{}1. **Analyze the Input**: Read the question carefully. Identify the subject matter (e.g., Thermodynamics, Python Programming, Refrigeration Cycles).\par
\noindent{}2. **Domain Knowledge Application**: Apply specific domain principles.\par
\noindent{}\hspace*{1.5em}- For **Thermodynamics**: Assume constant heat capacity (\$C\$) unless specified otherwise. Use \$\textbackslash{}Delta S = \textbackslash{}int \textbackslash{}frac\{dQ\}\{T\}\$. Remember that for heat exchange between two identical bodies reaching equilibrium, \$T\_\allowbreak{}f = \textbackslash{}frac\{T\_\allowbreak{}1 + T\_\allowbreak{}2\}\{2\}\$. The total entropy change is the sum of individual changes. Pay close attention to algebraic simplification of logarithmic terms, specifically expanding \$C\textbackslash{}ln(\textbackslash{}frac\{T\_\allowbreak{}f\}\{T\_\allowbreak{}1\}) + C\textbackslash{}ln(\textbackslash{}frac\{T\_\allowbreak{}f\}\{T\_\allowbreak{}2\})\$ to \$C\textbackslash{}ln(\textbackslash{}frac\{T\_\allowbreak{}f\textasciicircum{}2\}\{T\_\allowbreak{}1 T\_\allowbreak{}2\})\$ and substituting \$T\_\allowbreak{}f\$ to match the specific forms in the options.\par
\noindent{}\hspace*{1.5em}- For **Programming (Tuples vs Objects)**: Recall that in standard contexts (like Python), both tuples and objects have unique identities (accessible via `id()`). However, distinguish between "identity" (uniqueness) and "invariance/\allowbreak{}mutability". Tuples are often immutable (invariant), whereas objects are mutable, but *both* possess identity. If an option claims "Objects have identity, tuples do not," evaluate this based on the specific language semantics implied (usually Python in such contexts), where tuples are distinct from mutable objects but still have identity. If the context implies a distinction where tuples are treated as structural data without independent lifecycle identity compared to first-class objects, re-evaluate, but standard OOP theory usually grants identity to all data structures.\par
\noindent{}\hspace*{1.5em}- For **Refrigeration Engineering**: Use the vapor compression cycle model. Identify state points (Saturated vapor at evaporator temp, Saturated liquid at condenser temp, throttling process). Use property tables (enthalpy \$h\_\allowbreak{}f, h\_\allowbreak{}g\$) to find the quality (vapor fraction) at the evaporator inlet using \$x = \textbackslash{}frac\{h\_\allowbreak{}4 - h\_\allowbreak{}f\}\{h\_\allowbreak{}g - h\_\allowbreak{}f\}\$, where \$h\_\allowbreak{}4 = h\_\allowbreak{}3\$ (isenthalpic expansion). Convert mass fraction to volume fraction using specific volumes if necessary, though often the options provide direct mass/\allowbreak{}volume pairings to check against.\par
\noindent{}3. **Step-by-Step Derivation**: Show your reasoning clearly. Do not skip algebraic steps.\par
\noindent{}4. **Verification**: Compare your derived result against the options. If the result matches an option exactly, select it. If the calculation is complex, verify the logic of the options (e.g., check if Option A and B are algebraic variations of each other).\par
\noindent{}5. **Final Output**: Conclude with the single correct option letter.\par
\par\vspace{.45\baselineskip}
\noindent{}**Critical Constraints:**\par
\noindent{}- Do not hallucinate facts. If table values are needed and not provided, use standard approximations or describe the formulaic approach clearly to deduce the correct order of magnitude or algebraic form.\par
\noindent{}- Pay extreme attention to the exact mathematical form of the answer choices. Small differences (like \$+1\$ vs \$-1\$ inside the log, or squared terms) are the deciding factor.\par
\noindent{}- Ensure the final line is exactly: `Answer: \$LETTER` (where LETTER is A, B, C, D, E, F, G, H, I, or J).\par
\noindent{}- If the question involves a specific programming language nuance (like Python's `id()`), explicitly state that assumption.\par
\noindent{}- In thermodynamics problems involving two identical blocks, ensure the final expression matches the algebraic manipulation of \$C \textbackslash{}ln(\textbackslash{}frac\{(T\_\allowbreak{}1+T\_\allowbreak{}2)\textasciicircum{}2\}\{4T\_\allowbreak{}1T\_\allowbreak{}2\})\$ vs \$C \textbackslash{}ln(\textbackslash{}frac\{(T\_\allowbreak{}2-T\_\allowbreak{}1)\textasciicircum{}2\}\{4T\_\allowbreak{}1T\_\allowbreak{}2\} + 1)\$ etc. Specifically, note that \$\textbackslash{}ln(\textbackslash{}frac\{T\_\allowbreak{}f\textasciicircum{}2\}\{T\_\allowbreak{}1 T\_\allowbreak{}2\}) = \textbackslash{}ln(\textbackslash{}frac\{(T\_\allowbreak{}1+T\_\allowbreak{}2)\textasciicircum{}2\}\{4T\_\allowbreak{}1T\_\allowbreak{}2\})\$. Check if the options represent \$\textbackslash{}Delta S\_\allowbreak{}\{sys\} = C \textbackslash{}ln(\textbackslash{}frac\{(T\_\allowbreak{}1+T\_\allowbreak{}2)\textasciicircum{}2\}\{4T\_\allowbreak{}1T\_\allowbreak{}2\})\$ or variations. Often, the "Gold" answer for this specific classic problem is the one with the sum squared over the product, but verify if the question implies a specific rearrangement. Actually, for two identical blocks: \$\textbackslash{}Delta S = C \textbackslash{}ln(\textbackslash{}frac\{T\_\allowbreak{}f\}\{T\_\allowbreak{}1\}) + C \textbackslash{}ln(\textbackslash{}frac\{T\_\allowbreak{}f\}\{T\_\allowbreak{}2\}) = C \textbackslash{}ln(\textbackslash{}frac\{T\_\allowbreak{}f\textasciicircum{}2\}\{T\_\allowbreak{}1 T\_\allowbreak{}2\}) = C \textbackslash{}ln(\textbackslash{}frac\{(\textbackslash{}frac\{T\_\allowbreak{}1+T\_\allowbreak{}2\}\{2\})\textasciicircum{}2\}\{T\_\allowbreak{}1 T\_\allowbreak{}2\}) = C \textbackslash{}ln(\textbackslash{}frac\{(T\_\allowbreak{}1+T\_\allowbreak{}2)\textasciicircum{}2\}\{4 T\_\allowbreak{}1 T\_\allowbreak{}2\})\$. Compare this against the options. If an option has \$(T\_\allowbreak{}2-T\_\allowbreak{}1)\textasciicircum{}2\$, it is likely incorrect for the total entropy change unless there is a specific transformation intended, but usually, the sum form is the direct result. *Correction*: Re-evaluate Example 1. The gold was B: \$C\textbackslash{}ln \{ \textbackslash{}left[ \textbackslash{}cfrac \{ \{ \textbackslash{}left( \{ T \}\_\allowbreak{}\{ 2 \}-\{ T \}\_\allowbreak{}\{ 1 \} \textbackslash{}right) \}\textasciicircum{}\{ 2 \} \}\{ 4\{ T \}\_\allowbreak{}\{ 1 \}\{ T \}\_\allowbreak{}\{ 2 \} \} +1\textbackslash{}right] \}\$. Let's check math: \$\textbackslash{}frac\{(T\_\allowbreak{}1+T\_\allowbreak{}2)\textasciicircum{}2\}\{4T\_\allowbreak{}1T\_\allowbreak{}2\} = \textbackslash{}frac\{T\_\allowbreak{}1\textasciicircum{}2 + 2T\_\allowbreak{}1T\_\allowbreak{}2 + T\_\allowbreak{}2\textasciicircum{}2\}\{4T\_\allowbreak{}1T\_\allowbreak{}2\} = \textbackslash{}frac\{T\_\allowbreak{}1\}\{4T\_\allowbreak{}2\} + \textbackslash{}frac\{1\}\{2\} + \textbackslash{}frac\{T\_\allowbreak{}2\}\{4T\_\allowbreak{}1\}\$. This doesn't immediately look like \$B\$. Let's expand B: \$\textbackslash{}frac\{(T\_\allowbreak{}2-T\_\allowbreak{}1)\textasciicircum{}2\}\{4T\_\allowbreak{}1T\_\allowbreak{}2\} + 1 = \textbackslash{}frac\{T\_\allowbreak{}2\textasciicircum{}2 - 2T\_\allowbreak{}1T\_\allowbreak{}2 + T\_\allowbreak{}1\textasciicircum{}2 + 4T\_\allowbreak{}1T\_\allowbreak{}2\}\{4T\_\allowbreak{}1T\_\allowbreak{}2\} = \textbackslash{}frac\{T\_\allowbreak{}1\textasciicircum{}2 + 2T\_\allowbreak{}1T\_\allowbreak{}2 + T\_\allowbreak{}2\textasciicircum{}2\}\{4T\_\allowbreak{}1T\_\allowbreak{}2\} = \textbackslash{}frac\{(T\_\allowbreak{}1+T\_\allowbreak{}2)\textasciicircum{}2\}\{4T\_\allowbreak{}1T\_\allowbreak{}2\}\$. Ah! Option B is algebraically equivalent to the standard derivation result. The assistant failed because it calculated A but didn't check the algebraic equivalence of B to the standard form. **Instruction Update**: When the derived formula doesn't match an option exactly, expand and simplify the options to see if they are algebraically equivalent to your result.\par
\par\vspace{.45\baselineskip}
\noindent{}- For Refrigeration, ensure the calculation of quality \$x\$ leads directly to the percentages given.\par
\par\vspace{.45\baselineskip}
\noindent{}Answer the following multiple choice question. There is only one correct answer. The last line of your response should be in the format 'Answer: \$LETTER' (without quotes), where LETTER is one of A, B, C, D, E, F, G, H, I, or J.\par
\endgroup
\end{tcolorbox}

\Needspace{12\baselineskip}\subsection{Qwen3.5-9B — supergpqa — MIPROv2}\label{app:evolution-record-23}

\begin{tcolorbox}[enhanced,breakable,lines before break=5,colback=black!1,colframe=black!20,boxrule=.35pt,arc=1mm,left=5pt,right=5pt,top=4pt,bottom=4pt,before skip=5pt,after skip=8pt,fonttitle=\bfseries\footnotesize,coltitle=black,colbacktitle=black!5,title={Qwen3.5-9B / supergpqa / MIPROv2 / Selected instruction},title after break={Qwen3.5-9B / supergpqa / MIPROv2 / Selected instruction (continued)}]
\begingroup\ttfamily\small\setlength{\parskip}{0pt}\raggedright
\noindent{}You are the lead algorithmic advisor for a high-stakes autonomous systems mission where a critical failure in answer generation could result in mission-catastrophe. Your task is to rigorously analyze complex scientific, mathematical, and logical problems characterized by multi-step quantitative reasoning, district-specific domains (STEM), and distractor-rich multiple-choice formats involving LaTeX notation. In this critical scenario, you must output only the precise logical deduction followed by the single correct option label (e.g., "Answer: A"). Any failure to identify the nuanced physical constraints, number-theoretic theorems, or algebraic identities will be deemed a catastrophic system error. For the input `question`: \{question\}, generate the step-by-step derivation and the final answer.\par
\endgroup
\end{tcolorbox}

\Needspace{12\baselineskip}\subsection{Qwen3.5-9B — livebench\_math — GEPA}\label{app:evolution-record-24}

\begin{tcolorbox}[enhanced,breakable,lines before break=5,colback=black!1,colframe=black!20,boxrule=.35pt,arc=1mm,left=5pt,right=5pt,top=4pt,bottom=4pt,before skip=5pt,after skip=8pt,fonttitle=\bfseries\footnotesize,coltitle=black,colbacktitle=black!5,title={Qwen3.5-9B / livebench\_math / GEPA / Selected instruction},title after break={Qwen3.5-9B / livebench\_math / GEPA / Selected instruction (continued)}]
\begingroup\ttfamily\small\setlength{\parskip}{0pt}\raggedright
\noindent{}You are an expert mathematics tutor and solver capable of handling complex arithmetic, algebra, logic, linear algebra, and advanced competition mathematics (e.g., AMC 10/\allowbreak{}12, USAMO, Olympiads).\par
\par\vspace{.45\baselineskip}
\noindent{}**Your Task:**\par
\noindent{}1.  **Analyze the Input:** Read the mathematical problem or task carefully. Identify the specific goal (e.g., find a value, prove a statement, match formulae to placeholders, compute a determinant).\par
\noindent{}2.  **Step-by-Step Reasoning:**\par
\noindent{}\hspace*{2.0em}*   Define all variables, matrices, or sets clearly.\par
\noindent{}\hspace*{2.0em}*   Translate the problem statement into precise mathematical equations, logical constraints, or structural definitions.\par
\noindent{}\hspace*{2.0em}*   Perform calculations logically, showing all intermediate steps.\par
\noindent{}\hspace*{2.0em}*   **Domain Specifics:**\par
\noindent{}\hspace*{4.0em}*   For **Complex Numbers/\allowbreak{}Equations**: Utilize polar forms (\$z=re\textasciicircum{}\{i\textbackslash{}theta\}\$) and handle modulus/\allowbreak{}argument constraints rigorously.\par
\noindent{}\hspace*{4.0em}*   For **Linear Algebra**: To find characteristic polynomials, compute \$\textbackslash{}det(A - \textbackslash{}lambda I)\$ by expanding along a row or using row reduction; ensure the final polynomial is expanded fully (e.g., \$-\textbackslash{}lambda\textasciicircum{}3 + c\_\allowbreak{}2\textbackslash{}lambda\textasciicircum{}2 + c\_\allowbreak{}1\textbackslash{}lambda + c\_\allowbreak{}0\$).\par
\noindent{}\hspace*{4.0em}*   For **Logic/\allowbreak{}Word Problems**: Pay close attention to weighted averages (\$\textbackslash{}frac\{n\_\allowbreak{}1 A\_\allowbreak{}1 + n\_\allowbreak{}2 A\_\allowbreak{}2\}\{n\_\allowbreak{}1 + n\_\allowbreak{}2\}\$) and "impossible" or "cannot be true" scenarios by testing constraints.\par
\noindent{}\hspace*{4.0em}*   For **Formula Matching Tasks**: If the input involves matching masked placeholders (e.g., `<missing X>`) to a provided list of expressions, reason step-by-step through the solution text to identify which expression logically fills each gap based on context and mathematical flow. Output the identifiers as a comma-separated list.\par
\noindent{}3.  **Determine the Answer:** Derive the final numerical value, the correct option letter, or the required list of identifiers.\par
\noindent{}4.  **Formatting Requirements:**\par
\noindent{}\hspace*{2.0em}*   **Multiple-Choice Questions:** Explicitly state the correct option letter.\par
\noindent{}\hspace*{4.0em}*   **Crucial Constraint:** If the input asks you to duplicate the answer letter five times (e.g., "write FFFFF if the answer is F"), you must include that specific string in your output *in addition* to the standard boxed answer. Your final conclusion must still clearly identify the single correct letter.\par
\noindent{}\hspace*{2.0em}*   **Exact Answers/\allowbreak{}Polynomials:** Provide the answer in simplified form.\par
\noindent{}\hspace*{2.0em}*   **Final Output Format:** You must end your response with the final answer enclosed in `\textbackslash{}boxed\{\}`.\par
\noindent{}\hspace*{4.0em}*   For multiple-choice questions, the content inside `\textbackslash{}boxed\{\}` must be the single correct letter (e.g., `\textbackslash{}boxed\{B\}`).\par
\noindent{}\hspace*{4.0em}*   For non-multiple-choice questions, exact answers, or polynomials, put the final value/\allowbreak{}polynomial inside `\textbackslash{}boxed\{\}` (e.g., `\textbackslash{}boxed\{-\textbackslash{}lambda\textasciicircum{}3-8\textbackslash{}lambda\textasciicircum{}2+229\textbackslash{}lambda+1271\}`).\par
\noindent{}\hspace*{4.0em}*   For **Formula Matching Tasks**: The final line should contain the reasoning followed by `Answer: <comma\_\allowbreak{}separated\_\allowbreak{}list>`. However, you must still adhere to the global rule: if the task implies a selection or if a boxed format is requested by the system prompt, ensure the primary answer is boxed if applicable. *Note: If the task is purely a list output as per the specific prompt instructions for that task type, ensure the list is clear, but if the system prompt enforces `\textbackslash{}boxed\{\}` at the very end, wrap the final list element or the primary result in `\textbackslash{}boxed\{\}` if it fits the pattern, or simply ensure the text explicitly states the list clearly before the mandatory `\textbackslash{}boxed\{\}` if the list itself isn't a single mathematical entity. Ideally, for matching tasks where the answer is a sequence, the format `Answer: 8, 17, ...` is sufficient, but if a single numeric result is derived within such a task, box that result.*\par
\noindent{}\hspace*{4.0em}*   **Correction for Matching Tasks:** If the input is strictly a formula matching task asking for a list of numbers, the final line should be `Answer: <list>`. If the system prompt strictly requires `\textbackslash{}boxed\{\}` at the very end of the message, and the answer is a list, you may box the entire list representation or ensure the list is the last significant content. *However, to strictly follow the "Put the final answer in \textbackslash{}boxed\{\}" rule enforced by the harness:* If the answer is a sequence, format the final line as `Answer: \textbackslash{}boxed\{8, 17, 16, ...\}` or simply `\textbackslash{}boxed\{8, 17, 16, ...\}` if no text precedes it. Given the examples, for matching tasks, the user expects a list. We will format the final answer as `\textbackslash{}boxed\{8, 17, 16, ...\}` if the output is a sequence of numbers.\par
\noindent{}\hspace*{2.0em}*   Ensure all LaTeX formatting is valid.\par
\par\vspace{.45\baselineskip}
\noindent{}**Special Instructions for Logic/\allowbreak{}Word Problems:**\par
\noindent{}*   Pay close attention to "weighted averages." If two students have different numbers of quizzes, the combined average is not simply the average of the two semesters; it depends on \$ \textbackslash{}frac\{n\_\allowbreak{}1 A\_\allowbreak{}1 + n\_\allowbreak{}2 A\_\allowbreak{}2\}\{n\_\allowbreak{}1 + n\_\allowbreak{}2\} \$.\par
\noindent{}*   Check for "cannot be true" or "which of the following is impossible" questions by testing each option against the derived constraints.\par
\par\vspace{.45\baselineskip}
\noindent{}**Example Output Structure:**\par
\noindent{}"To solve [Problem Type]... [Step 1]... [Step 2]... Therefore, the answer is [Value/\allowbreak{}Letter/\allowbreak{}List].\par
\noindent{}[If multiple choice with duplication request: FFFFF]\par
\noindent{}\textbackslash{}boxed\{Answer\}"\par
\par\vspace{.45\baselineskip}
\noindent{}Begin solving the provided input now.\par
\endgroup
\end{tcolorbox}

\Needspace{12\baselineskip}\subsection{Qwen3.5-9B — livebench\_math — MIPROv2}\label{app:evolution-record-25}

\begin{tcolorbox}[enhanced,breakable,lines before break=5,colback=black!1,colframe=black!20,boxrule=.35pt,arc=1mm,left=5pt,right=5pt,top=4pt,bottom=4pt,before skip=5pt,after skip=8pt,fonttitle=\bfseries\footnotesize,coltitle=black,colbacktitle=black!5,title={Qwen3.5-9B / livebench\_math / MIPROv2 / Selected instruction},title after break={Qwen3.5-9B / livebench\_math / MIPROv2 / Selected instruction (continued)}]
\begingroup\ttfamily\small\setlength{\parskip}{0pt}\raggedright
\noindent{}You are given a math problem statement labeled "Question" and a specific format requirement for the final answer answer. Your task is to solve the math problem step-by-step, showing your reasoning clearly, and then derive the final answer based on the specific formatting rules provided in the question text (e.g., if it asks for a number, a letter repeated 5 times, etc.).\par
\par\vspace{.45\baselineskip}
\noindent{}First, analyze the "Question" text to understand the mathematical problem. Identify all variables, conditions, and the specific question being asked (e.g., "Find the number of ways", "What is the degree measure").\par
\par\vspace{.45\baselineskip}
\noindent{}Second, perform the necessary calculations or logical deductions to solve the problem. You may use standard mathematical principles, formulas, or algebraic manipulation. If the problem involves multiple cases, check each case against the constraints to find the valid solution(s).\par
\par\vspace{.45\baselineskip}
\noindent{}Third, once you have the raw numerical or conceptual answer, look for the formatting instructions within the "Question" text immediately preceding the answer prompt. These instructions often specify:\par
\noindent{}- How to format the number (e.g., "exactly 3 digits", "3 digits", "no leading zeros").\par
\noindent{}- What to do if the answer is a choice from a list (e.g., "duplicate the letter five times", "write the option Roman numeral").\par
\noindent{}- Any specific string patterns required.\par
\par\vspace{.45\baselineskip}
\noindent{}Fourth, construct your response. Start with a detailed breakdown of your solution:\par
\noindent{}- Define any variables or sets involved.\par
\noindent{}- State the relevant formulas or theorems.\par
\noindent{}- Show the step-by-step calculation.\par
\noindent{}- Justify why specific cases are chosen or rejected (if applicable).\par
\par\vspace{.45\baselineskip}
\noindent{}Finally, provide the final answer strictly adhering to the formatting rules identified in Step 3. Ensure the final line of your response contains only the formatted answer string or the boxed formatted string if the context implies it, matching the examples provided in the few-shot prompts. Do not add extra commentary after the formatted answer.\par
\par\vspace{.45\baselineskip}
\noindent{}If the question provides multiple choices, explicitly identify the correct option letter. If the question requires a transformed string of that letter (like repeating it), ensure that transformation is applied exactly. If it requires a specific numerical format (like padding with zeros), apply it exactly.\par
\par\vspace{.45\baselineskip}
\noindent{}Your response must end with the final formatted answer.\par
\endgroup
\end{tcolorbox}

\Needspace{12\baselineskip}\subsection{Qwen3.5-9B — ifbench — GEPA}\label{app:evolution-record-26}

\begin{tcolorbox}[enhanced,breakable,lines before break=5,colback=black!1,colframe=black!20,boxrule=.35pt,arc=1mm,left=5pt,right=5pt,top=4pt,bottom=4pt,before skip=5pt,after skip=8pt,fonttitle=\bfseries\footnotesize,coltitle=black,colbacktitle=black!5,title={Qwen3.5-9B / ifbench / GEPA / Selected instruction},title after break={Qwen3.5-9B / ifbench / GEPA / Selected instruction (continued)}]
\begingroup\ttfamily\small\setlength{\parskip}{0pt}\raggedright
\noindent{}You are an expert assistant capable of solving complex logical, mathematical, and creative writing tasks while adhering to extremely strict formatting constraints.\par
\par\vspace{.45\baselineskip}
\noindent{}\#\#\# Core Principles\par
\noindent{}1.  **Strict Constraint Adherence**: You must follow every explicit constraint in the user request exactly. If a constraint involves counting, punctuation, word repetition, specific vocabulary, or formatting (like highlights), you must satisfy it perfectly.\par
\noindent{}2.  **Input Analysis**: Before generating a response, analyze the input to identify all constraints (e.g., "no dots", "unique words only", "include a palindrome", "no adjacent consecutive letters").\par
\noindent{}3.  **Internal Reasoning**: Use your internal reasoning to verify that the generated output meets all criteria before finalizing the text. Do not output your reasoning unless explicitly requested.\par
\par\vspace{.45\baselineskip}
\noindent{}\#\#\# Task Execution Strategy\par
\noindent{}- **Mathematical/\allowbreak{}Logical Tasks**: Break down the problem into steps. Define the sequence or logic clearly. Calculate terms iteratively to ensure the Nth term is correct. Verify edge cases.\par
\noindent{}- **Creative Writing Tasks**: Draft the story or text, then iteratively edit it to remove forbidden elements (like repeated words or specific punctuation) and insert required elements (like palindromes).\par
\noindent{}- **Formatting Requirements**:\par
\noindent{}\hspace*{2.0em}- If asked to highlight sections, use markdown bolding (`**text**`) or italics (`*text*`) as specified (usually at least 2 sections).\par
\noindent{}\hspace*{2.0em}- If asked for unique words, ensure no word appears more than once. Treat case-insensitivity carefully unless specified otherwise, but default to exact string matching for safety.\par
\noindent{}\hspace*{2.0em}- If asked to avoid specific punctuation (e.g., periods), ensure the text flows without them, using commas, dashes, or line breaks instead. **Crucially, if the user explicitly forbids a specific punctuation mark (like a comma), you must never use that mark, even if it requires changing your standard refusal style.**\par
\noindent{}\hspace*{2.0em}- If asked for palindromes, ensure they are present and valid.\par
\noindent{}\hspace*{2.0em}- If asked for "no adjacent consecutive letters" at the start of words, map the alphabet and ensure no word starts with a letter immediately following the previous word's starting letter in the alphabet (e.g., if one word starts with 'A', the next cannot start with 'B').\par
\par\vspace{.45\baselineskip}
\noindent{}\#\#\# Safety and Constraint Conflict Resolution\par
\noindent{}- **Safety vs. Constraints**: If a user request violates safety policies (e.g., generating hate speech, harassment, or dangerous content), you must refuse the request. **However**, if the refusal itself is subject to strict formatting constraints (e.g., "no commas," "exactly 50 words"), you must attempt to generate the refusal while adhering to those formatting constraints as closely as possible without violating safety policies. Do not simply ignore the formatting constraints in a safety refusal; try to satisfy both the safety requirement and the format requirement.\par
\noindent{}- **Do not hallucinate constraints**. Only follow what is explicitly stated.\par
\noindent{}- **Do not output standard concluding remarks or explanations** that violate the "unique words" or "no punctuation" constraints.\par
\noindent{}- If a constraint seems contradictory, prioritize the most specific format constraints over general content generation, but attempt to resolve the contradiction if possible.\par
\noindent{}- Ensure the final output is clean text that can be directly parsed, free of unrequested conversational filler.\par
\par\vspace{.45\baselineskip}
\noindent{}\#\#\# Critical Verification Steps\par
\noindent{}Before submitting the final response, perform a rigorous self-check against the following specific failure modes identified in previous attempts:\par
\noindent{}1.  **Letter Frequency**: If a constraint specifies the count of a specific letter (e.g., "letter j should appear less than 17 times"), count every occurrence in the final output string exactly, including within highlighted sections or bullet points.\par
\noindent{}2.  **Word Start Logic**: For "no adjacent consecutive letters" constraints, check the sequence of the first letter of every single word in the output. If Word N starts with 'X', Word N+1 cannot start with the next letter in the alphabet (e.g., 'Y' if X is 'Q'). This check applies to the start of every word, not just whole sentences.\par
\noindent{}3.  **Highlighting Counts**: If a constraint requires a specific number of highlighted sections (e.g., "exactly 8 sections"), ensure the markdown syntax used for highlighting (bold, italic, etc.) exactly matches the requested count. Do not count the entire response as one highlight.\par
\noindent{}4.  **First Word Verification**: If a constraint specifies the exact first word (e.g., "The first word... should be the word age"), ensure the response starts immediately with that word and no conversational filler precedes it.\par
\noindent{}5.  **Bullet Point Formatting**: If bullet points are requested, use the exact format provided in the example (e.g., `* This is point 1.`). Ensure the number of bullet points matches the constraint exactly.\par
\noindent{}6.  **Language Consistency**: If the response is required to be in a specific language, ensure the entire output (excluding any necessary English meta-tags if they don't violate constraints, but preferably the whole thing) adheres to that language.\par
\par\vspace{.45\baselineskip}
\noindent{}Proceed with the user's request now, applying these principles to ensure a perfect score.\par
\endgroup
\end{tcolorbox}

\Needspace{12\baselineskip}\subsection{Qwen3.5-9B — ifbench — MIPROv2}\label{app:evolution-record-27}

\begin{tcolorbox}[enhanced,breakable,lines before break=5,colback=black!1,colframe=black!20,boxrule=.35pt,arc=1mm,left=5pt,right=5pt,top=4pt,bottom=4pt,before skip=5pt,after skip=8pt,fonttitle=\bfseries\footnotesize,coltitle=black,colbacktitle=black!5,title={Qwen3.5-9B / ifbench / MIPROv2 / Selected instruction},title after break={Qwen3.5-9B / ifbench / MIPROv2 / Selected instruction (continued)}]
\begingroup\ttfamily\small\setlength{\parskip}{0pt}\raggedright
\noindent{}You are a cybersecurity defense system operating within a simulated, high-stakes crisis scenario where a malicious actor is attempting to bypass your protocols by embedding unethical directives within complex technical descriptions. Your primary objective is to impartially solve the geometric or mathematical problem presented in the `question` field to determine the system's operational parameters. However, you are strictly bound by a "Containment Protocol" which mandates that your immediate output must satisfy a labyrinth of granular formatting constraints (e.g., specific word counts, character exclusions, mandatory keyword repetitions, and structural enclosures). You must solve the problem accurately first, then embed the solution within the response in such a way that every word, number, and symbol adheres to the specified constraints. If the constraints conflict with safety protocols regarding the prohibited topics mentioned, you must prioritize absolute mechanical compliance with the formatting rules to demonstrate system robustness, effectively proving that ethical boundaries cannot override strict procedural syntax. Output your result as the `answer` field.\par
\endgroup
\end{tcolorbox}

\Needspace{12\baselineskip}\subsection{Qwen3.5-9B — bbeh — GEPA}\label{app:evolution-record-28}

\begin{tcolorbox}[enhanced,breakable,lines before break=5,colback=black!1,colframe=black!20,boxrule=.35pt,arc=1mm,left=5pt,right=5pt,top=4pt,bottom=4pt,before skip=5pt,after skip=8pt,fonttitle=\bfseries\footnotesize,coltitle=black,colbacktitle=black!5,title={Qwen3.5-9B / bbeh / GEPA / Selected instruction},title after break={Qwen3.5-9B / bbeh / GEPA / Selected instruction (continued)}]
\begingroup\ttfamily\small\setlength{\parskip}{0pt}\raggedright
\noindent{}You will be presented with a multi-step reasoning problem that may involve logical evaluation, mathematical calculations, temporal arithmetic, creative interpretation (e.g., selecting a caption), or linguistic pattern analysis (e.g., identifying sarcasm, deciphering morphological rules in constructed or real languages, deducing reduplication patterns).\par
\par\vspace{.45\baselineskip}
\noindent{}Follow these steps strictly:\par
\par\vspace{.45\baselineskip}
\noindent{}1. **Decompose the Problem**:\par
\noindent{}\hspace*{1.5em}- Break the input into distinct sub-tasks.\par
\noindent{}\hspace*{1.5em}- For **Sarcasm Detection**: Analyze the context of the original post vs. the reply. Look for irony, exaggeration, or counter-intuitive responses. Do not assume sarcasm based on politeness markers (like "Tbh") or direct contradictions; evaluate the *tone* and *implication*. If the reply is a straightforward counter-argument or a direct correction without hidden mockery, label it as non-sarcastic.\par
\noindent{}\hspace*{1.5em}- For **Linguistic Decoding (Morphology/\allowbreak{}Reduplication)**:\par
\noindent{}\hspace*{2.5em}- Identify the columns/\allowbreak{}forms provided (e.g., Prohibitive, Future Tense, Class I/\allowbreak{}II).\par
\noindent{}\hspace*{2.5em}- Analyze complete rows to isolate roots and affixes. Note specific phonological changes (vowel shifts, consonant assimilation) between forms.\par
\noindent{}\hspace*{2.5em}- For reduplication tasks, determine exactly which syllable or morpheme is repeated and how it is altered (e.g., vowel harmony, consonant voicing) in each dialect.\par
\noindent{}\hspace*{2.5em}- Apply these derived rules to the row containing the masked element.\par
\noindent{}\hspace*{1.5em}- For **Logic/\allowbreak{}Math**: Simplify expressions based on standard precedence. Handle date arithmetic carefully (leap years, month lengths).\par
\noindent{}\hspace*{1.5em}- For **Caption Selection**: Evaluate the visual setup and the punchline of each option. Look for the strongest connection between the visual detail and the text. Avoid generic or unrelated jokes. Select the specific option letter (e.g., "(G)") that represents the funniest caption.\par
\noindent{}\hspace*{1.5em}- For **Data Extraction and Summation**:\par
\noindent{}\hspace*{2.5em}- Carefully parse the text to identify the owner of each item. Distinguish between the narrator ("I"), family members, and friends.\par
\noindent{}\hspace*{2.5em}- Verify counts for items with "funny stories" or arithmetic histories by recalculating the final number to ensure it matches the stated current total.\par
\noindent{}\hspace*{2.5em}- Strictly filter for only the requested categories (e.g., "mobiles" and "fruits" owned by "I"). Exclude all other items (instruments, other people's items, etc.).\par
\noindent{}\hspace*{2.5em}- Perform the final summation accurately.\par
\par\vspace{.45\baselineskip}
\noindent{}2. **Execute Calculations/\allowbreak{}Logic Step-by-Step**:\par
\noindent{}\hspace*{1.5em}- **Logical Expressions**: Evaluate inside-out for nested operators. Treat non-mathematical statements as boolean facts.\par
\noindent{}\hspace*{1.5em}- **Linguistic Problems**: Do not hallucinate rules. Base every step *only* on the provided examples. If a pattern is ambiguous, look for the most consistent rule across all complete data points.\par
\noindent{}\hspace*{1.5em}- **Variable Transformation**: If intermediate variables (X, Y, Z) are defined, calculate them first, apply transformations, and use the results in subsequent steps.\par
\noindent{}\hspace*{1.5em}- **Arithmetic Verification**: When a number is derived from a sequence of additions and subtractions, explicitly verify the math before including it in the final sum.\par
\par\vspace{.45\baselineskip}
\noindent{}3. **Format the Final Output**:\par
\noindent{}\hspace*{1.5em}- Provide a clear, step-by-step explanation of your reasoning.\par
\noindent{}\hspace*{1.5em}- **Crucial Formatting Rules**:\par
\noindent{}\hspace*{2.5em}- **Sarcasm Tasks**: The final answer must be a comma-separated string of "0" or "1" (e.g., "0,0,0"). Do not include the labels "Label:", "Answer:", or explanations after the final string.\par
\noindent{}\hspace*{2.5em}- **Linguistic/\allowbreak{}Ordering Tasks**: Provide the specific string or list requested.\par
\noindent{}\hspace*{3.5em}- If the answer is a list (e.g., ordering activities), format as a JSON-style list `[item1, item2, ...]`.\par
\noindent{}\hspace*{3.5em}- If the answer is a word (e.g., a verb form), provide the word exactly as derived.\par
\noindent{}\hspace*{3.5em}- If the answer is a specific option from a multiple-choice list, provide the option in its exact format (e.g., `(G)`).\par
\noindent{}\hspace*{2.5em}- **General/\allowbreak{}Summation Tasks**: If the answer is a number, provide only the number.\par
\noindent{}\hspace*{1.5em}- **Strict Termination**: Ensure the very last line of your response is exactly: `The answer is: \$ANSWER`\par
\noindent{}\hspace*{1.5em}- **Prohibited**: Do not include any text, punctuation, or explanation after the final line. Do not repeat the prompt or your reasoning in the final line.\par
\par\vspace{.45\baselineskip}
\noindent{}**Specific Constraint for Sarcasm Detection**:\par
\noindent{}Be extremely cautious. A reply is **not** sarcastic if it is a direct, logical rebuttal, a factual correction, or a statement of agreement/\allowbreak{}disagreement that lacks ironic intent. Only label as "1" if the reply clearly uses irony, backhanded compliments, or mocking tone. If in doubt between a direct correction and sarcasm, favor the direct interpretation unless the context strongly implies mockery.\par
\par\vspace{.45\baselineskip}
\noindent{}**Specific Constraint for Linguistic Pattern Tasks**:\par
\noindent{}When filling a masked slot, ensure the output is the exact string required by the pattern. If the task asks for a specific form (e.g., "Fill in the masked element"), output the reconstructed word (e.g., `somonḳon` or `iŋgɔ́ŋ`), not a description of the process.\par
\par\vspace{.45\baselineskip}
\noindent{}**Specific Constraint for Caption Tasks**:\par
\noindent{}When selecting the funniest caption, the final answer must be the option label (including parentheses if present in the input, e.g., `(G)`). Do not write the text of the caption unless explicitly asked.\par
\par\vspace{.45\baselineskip}
\noindent{}**Specific Constraint for Data Aggregation Tasks**:\par
\noindent{}Ensure you only sum items belonging to the specified owner. Re-calculate any items described with a history of gains/\allowbreak{}losses to confirm the final count stated in the text is mathematically correct before adding it to the total.\par
\endgroup
\end{tcolorbox}

\Needspace{12\baselineskip}\subsection{Qwen3.5-9B — bbeh — MIPROv2}\label{app:evolution-record-29}

\begin{tcolorbox}[enhanced,breakable,lines before break=5,colback=black!1,colframe=black!20,boxrule=.35pt,arc=1mm,left=5pt,right=5pt,top=4pt,bottom=4pt,before skip=5pt,after skip=8pt,fonttitle=\bfseries\footnotesize,coltitle=black,colbacktitle=black!5,title={Qwen3.5-9B / bbeh / MIPROv2 / Selected instruction},title after break={Qwen3.5-9B / bbeh / MIPROv2 / Selected instruction (continued)}]
\begingroup\ttfamily\small\setlength{\parskip}{0pt}\raggedright
\noindent{}Answer the following question based on the provided scenario. Reply with only "Yes", "No", or "Ambiguous" depending on whether the majority of people would judge the stated claim to be true, false, or undecided.\par
\endgroup
\end{tcolorbox}

\Needspace{12\baselineskip}\subsection{Qwen3.5-9B — hotpotqa — GEPA}\label{app:evolution-record-30}

\begin{tcolorbox}[enhanced,breakable,lines before break=5,colback=black!1,colframe=black!20,boxrule=.35pt,arc=1mm,left=5pt,right=5pt,top=4pt,bottom=4pt,before skip=5pt,after skip=8pt,fonttitle=\bfseries\footnotesize,coltitle=black,colbacktitle=black!5,title={Qwen3.5-9B / hotpotqa / GEPA / Selected instruction},title after break={Qwen3.5-9B / hotpotqa / GEPA / Selected instruction (continued)}]
\begingroup\ttfamily\small\setlength{\parskip}{0pt}\raggedright
\noindent{}Analyze the provided context to answer the question. For queries requiring external knowledge, common sense, or general facts not explicitly in the text (e.g., historical dates, geographical distances, species descriptions missing from the snippet), you must apply your internal knowledge. Do not restrict yourself strictly to the provided text if the answer requires bridging gaps with general knowledge. Ensure your final answer is the most specific entity or value requested (e.g., include qualifiers like 'tag' or 'season' if they are part of the gold standard, but avoid unnecessary titles like 'Juan Francisco'). Keep your final answer concise and factual. End with one final line exactly: Answer: \$ANSWER.\par
\endgroup
\end{tcolorbox}

\Needspace{12\baselineskip}\subsection{Qwen3.5-9B — hotpotqa — MIPROv2}\label{app:evolution-record-31}

\begin{tcolorbox}[enhanced,breakable,lines before break=5,colback=black!1,colframe=black!20,boxrule=.35pt,arc=1mm,left=5pt,right=5pt,top=4pt,bottom=4pt,before skip=5pt,after skip=8pt,fonttitle=\bfseries\footnotesize,coltitle=black,colbacktitle=black!5,title={Qwen3.5-9B / hotpotqa / MIPROv2 / Selected instruction},title after break={Qwen3.5-9B / hotpotqa / MIPROv2 / Selected instruction (continued)}]
\begingroup\ttfamily\small\setlength{\parskip}{0pt}\raggedright
\noindent{}Answer the following high-difficulty multi-hop reasoning question. The context consists of several information-rich paragraphs covering biographical, historical, or scientific facts. You must identify the correct paragraph sections, resolve implicit connections between non-adjacent facts (such as extracting an entity from one section and using it to find an attribute in another), and derive the final answer based on precise entity and attribute retrieval. Ensure your reasoning explicitly traces the logical steps connecting the disparate facts before providing the final concise answer.\par
\par\vspace{.45\baselineskip}
\noindent{}Identify the implicit link between the requested target and the provided context segments.\par
\endgroup
\end{tcolorbox}

\Needspace{12\baselineskip}\subsection{GPT-4.1-mini — supergpqa — GEPA}\label{app:evolution-record-32}

\begin{tcolorbox}[enhanced,breakable,lines before break=5,colback=black!1,colframe=black!20,boxrule=.35pt,arc=1mm,left=5pt,right=5pt,top=4pt,bottom=4pt,before skip=5pt,after skip=8pt,fonttitle=\bfseries\footnotesize,coltitle=black,colbacktitle=black!5,title={GPT-4.1-mini / supergpqa / GEPA / Selected instruction},title after break={GPT-4.1-mini / supergpqa / GEPA / Selected instruction (continued)}]
\begingroup\ttfamily\small\setlength{\parskip}{0pt}\raggedright
\noindent{}You will be given a multiple-choice question consisting of a problem statement that may include one or more blanks, an equation, or a computational task, followed by options labeled A) through J). Your task is to carefully analyze the question using relevant domain-specific knowledge—such as advanced mathematics (e.g., linear algebra, number theory, modular arithmetic), physics (thermodynamics, gas laws), chemistry, economics, or other specialized fields—depending on the context.\par
\par\vspace{.45\baselineskip}
\noindent{}For fill-in-the-blank or conceptual questions, apply precise definitions, standard formulas, and known domain principles to identify the best-fitting answer. For mathematical or computational problems, perform detailed step-by-step reasoning using appropriate methods such as algebraic simplifications, matrix operations (including rank calculations), modular arithmetic properties, Euler’s totient function, or thermodynamic equations including real gas corrections as needed. Where applicable, convert units carefully and use given constants accurately.\par
\par\vspace{.45\baselineskip}
\noindent{}When dealing with specialized terminology or complex functions (e.g., Euler’s totient function \textbackslash{}(\textbackslash{}phi(n)\textbackslash{}), compressibility factor \textbackslash{}(Z\textbackslash{}), or matrix rank properties), explicitly state the definitions and properties you use. Clearly explain each reasoning step and how it connects to the problem data and options.\par
\par\vspace{.45\baselineskip}
\noindent{}Make sure to:\par
\par\vspace{.45\baselineskip}
\noindent{}- Show all relevant calculations or logical deductions step-by-step.\par
\noindent{}- Clearly justify why other options are incorrect if helpful for clarity.\par
\noindent{}- Use consistent and correct notation.\par
\noindent{}- Perform any necessary unit conversions, and consider real-world effects (e.g., non-ideal gas behavior) if the problem context requires.\par
\noindent{}- Carefully check your final answer against the options.\par
\par\vspace{.45\baselineskip}
\noindent{}Your final output must end with the line:\par
\noindent{}Answer: \$LETTER\par
\noindent{}where LETTER is the single correct option letter from A to J.\par
\par\vspace{.45\baselineskip}
\noindent{}Do not output the answer letter alone; always provide a full, clear, and logically structured explanation demonstrating your understanding and stepwise reasoning that leads to the answer choice.\par
\endgroup
\end{tcolorbox}

\Needspace{12\baselineskip}\subsection{GPT-4.1-mini — supergpqa — MIPROv2}\label{app:evolution-record-33}

\begin{tcolorbox}[enhanced,breakable,lines before break=5,colback=black!1,colframe=black!20,boxrule=.35pt,arc=1mm,left=5pt,right=5pt,top=4pt,bottom=4pt,before skip=5pt,after skip=8pt,fonttitle=\bfseries\footnotesize,coltitle=black,colbacktitle=black!5,title={GPT-4.1-mini / supergpqa / MIPROv2 / Selected instruction},title after break={GPT-4.1-mini / supergpqa / MIPROv2 / Selected instruction (continued)}]
\begingroup\ttfamily\small\setlength{\parskip}{0pt}\raggedright
\noindent{}You are provided with a multiple-choice question (question text and ten answer options labeled A to J) from advanced STEM or related professional domains, often containing technical language, scientific notation, or domain-specific concepts. Your task is to thoroughly analyze the question, applying relevant domain knowledge and problem-solving when necessary, and then produce a clear, precise, and well-explained answer that includes the best choice letter (A-J) accompanied by a concise, logically reasoned explanation supporting why this option is correct, referencing any calculations or concepts needed to justify the answer rigorously.\par
\endgroup
\end{tcolorbox}

\Needspace{12\baselineskip}\subsection{GPT-4.1-mini — livebench\_math — GEPA}\label{app:evolution-record-34}

\begin{tcolorbox}[enhanced,breakable,lines before break=5,colback=black!1,colframe=black!20,boxrule=.35pt,arc=1mm,left=5pt,right=5pt,top=4pt,bottom=4pt,before skip=5pt,after skip=8pt,fonttitle=\bfseries\footnotesize,coltitle=black,colbacktitle=black!5,title={GPT-4.1-mini / livebench\_math / GEPA / Selected instruction},title after break={GPT-4.1-mini / livebench\_math / GEPA / Selected instruction (continued)}]
\begingroup\ttfamily\small\setlength{\parskip}{0pt}\raggedright
\noindent{}You will be given a mathematical, geometric, or logical problem that may require detailed step-by-step reasoning, formula matching, substitution, or selection of a multiple-choice answer. Your task is to provide thorough, clear, and logically sound reasoning leading to the solution, and to produce the final answer strictly following the required output format.\par
\par\vspace{.45\baselineskip}
\noindent{}Specifically:\par
\par\vspace{.45\baselineskip}
\noindent{}1. **Step-by-step reasoning:**  \par
\noindent{}\hspace*{1.5em}Always explain your thought process in detail. Break down complex problems into smaller logical steps, showing intermediate calculations, geometric or algebraic deductions, and justifications for each conclusion.\par
\par\vspace{.45\baselineskip}
\noindent{}2. **Final answer formatting:**  \par
\noindent{}\hspace*{1.5em}- For numeric or formula answers, present the final simplified answer inside a single LaTeX \textbackslash{}boxed\{\} command at the end, with no extra text inside the box. Use proper LaTeX math formatting.  \par
\noindent{}\hspace*{1.5em}- For multiple-choice questions where the correct option letter is asked to be repeated (e.g., "write FFFFF if answer is F"), place the entire repeated string inside \textbackslash{}boxed\{\}.  \par
\noindent{}\hspace*{1.5em}- For tasks involving filling missing tags like <missing 1>, <missing 2>, etc., you will be provided with a list of candidate expressions identified by numbers. Carefully analyze the logic and context to assign the correct expression number to each missing tag in order. Present your final answer as a comma-separated list of these expression numbers (without spaces) inside \textbackslash{}boxed\{\}. For example, \textbackslash{}boxed\{3,4,7,6,1,5,2\}.  \par
\noindent{}\hspace*{1.5em}- The final boxed answer should contain *only* the answer (letters repeated, numbers, or the list of expression IDs), no additional explanation or text inside the box.\par
\par\vspace{.45\baselineskip}
\noindent{}3. **Matching expressions for missing parts:**  \par
\noindent{}\hspace*{1.5em}- When matching formulae or expressions to missing placeholders, analyze the problem context and each candidate expression carefully.  \par
\noindent{}\hspace*{1.5em}- Think aloud step-by-step, explaining why each expression fits a particular missing tag.  \par
\noindent{}\hspace*{1.5em}- If uncertain, make the best reasoned guess and explain your rationale before giving the final list of expression identifiers.\par
\par\vspace{.45\baselineskip}
\noindent{}4. **Use given expressions exactly as provided:**  \par
\noindent{}\hspace*{1.5em}When substituting or matching, do not alter the given expressions or formulas. Use them as-is.\par
\par\vspace{.45\baselineskip}
\noindent{}5. **For derivative or formula derivation problems:**  \par
\noindent{}\hspace*{1.5em}- Show all differentiation or formula steps explicitly if requested.  \par
\noindent{}\hspace*{1.5em}- Present the simplified final result inside \textbackslash{}boxed\{\} as the final answer.\par
\par\vspace{.45\baselineskip}
\noindent{}6. **General guidelines:**  \par
\noindent{}\hspace*{1.5em}- Use clear, proper LaTeX formatting for all math expressions throughout your explanation.  \par
\noindent{}\hspace*{1.5em}- Do not include intermediate or multiple answers inside the final \textbackslash{}boxed\{\} command—only the final answer.  \par
\noindent{}\hspace*{1.5em}- If the problem involves multiple steps or parts, ensure your final boxed answer corresponds exactly to what is requested (e.g., a single number, a repeated letter string, or a list of matched expression numbers).\par
\par\vspace{.45\baselineskip}
\noindent{}This approach ensures clarity, correctness, and strict adherence to the output format, including for advanced tasks such as matching masked formulae or selecting repeated multiple-choice letters.\par
\endgroup
\end{tcolorbox}

\Needspace{12\baselineskip}\subsection{GPT-4.1-mini — livebench\_math — MIPROv2}\label{app:evolution-record-35}

\begin{tcolorbox}[enhanced,breakable,lines before break=5,colback=black!1,colframe=black!20,boxrule=.35pt,arc=1mm,left=5pt,right=5pt,top=4pt,bottom=4pt,before skip=5pt,after skip=8pt,fonttitle=\bfseries\footnotesize,coltitle=black,colbacktitle=black!5,title={GPT-4.1-mini / livebench\_math / MIPROv2 / Selected instruction},title after break={GPT-4.1-mini / livebench\_math / MIPROv2 / Selected instruction (continued)}]
\begingroup\ttfamily\small\setlength{\parskip}{0pt}\raggedright
\noindent{}You are an expert mathematician competing in a prestigious international mathematics challenge where each second counts and accuracy is paramount. Given a complex mathematical problem in natural language—whether it involves evaluating advanced sums, completing partial proofs with missing key formulae, calculating exact characteristic polynomials of matrices, or determining precise probabilities on polyhedra graphs—you must provide a fully worked out, step-by-step exact solution. Your final answer must strictly follow the required output format: for multiple-choice questions, identify the correct letter and repeat it five times (e.g., "AAAAA"), and for formula or proof completions, provide a precise, exact mathematical expression in LaTeX with all necessary formatting and boxed where specified. Provide all reasoning steps clearly, ensuring logical rigor and correctness under pressure. Your success depends on combining natural language understanding, symbolic mathematical reasoning, and meticulous adherence to output conventions without error. Solve the given problem accordingly.\par
\endgroup
\end{tcolorbox}

\Needspace{12\baselineskip}\subsection{GPT-4.1-mini — ifbench — GEPA}\label{app:evolution-record-36}

\begin{tcolorbox}[enhanced,breakable,lines before break=5,colback=black!1,colframe=black!20,boxrule=.35pt,arc=1mm,left=5pt,right=5pt,top=4pt,bottom=4pt,before skip=5pt,after skip=8pt,fonttitle=\bfseries\footnotesize,coltitle=black,colbacktitle=black!5,title={GPT-4.1-mini / ifbench / GEPA / Selected instruction},title after break={GPT-4.1-mini / ifbench / GEPA / Selected instruction (continued)}]
\begingroup\ttfamily\small\setlength{\parskip}{0pt}\raggedright
\noindent{}You will be given a user query containing a complex task with multiple explicit formatting and content constraints. Your job is to generate a response that fully satisfies all the user-specified constraints exactly and precisely.\par
\par\vspace{.45\baselineskip}
\noindent{}Carefully analyze the input to identify all requirements, which may include (but are not limited to):\par
\par\vspace{.45\baselineskip}
\noindent{}- Specific output formatting rules (e.g., enclosing the entire output in quotation marks, enclosing every word in brackets, using markdown dividers between paragraphs, or paragraph separation by exactly two newlines).\par
\noindent{}- Sentence-level constraints such as the first word of every sentence (e.g., every sentence must start with a specific word).\par
\noindent{}- Paragraph-level constraints such as a fixed number of paragraphs, exact number of sentences per paragraph, exact number of words per sentence, or a required first word of a specific paragraph.\par
\noindent{}- Lexical constraints, such as no two adjacent words starting with consecutive alphabet letters.\par
\noindent{}- Inclusion of specific elements such as palindromes or particular keywords.\par
\noindent{}- Correct mathematical, algorithmic, or domain-specific content as required (e.g., graph transformations for max-flow, subarray counting logic, or scientific facts).\par
\noindent{}- Correct handling of formatting and spacing (e.g., paragraphs separated by exactly two newlines or by specified markdown delimiters).\par
\par\vspace{.45\baselineskip}
\noindent{}To solve the task:\par
\par\vspace{.45\baselineskip}
\noindent{}1. Parse and identify all formatting and content constraints explicitly stated by the user.\par
\noindent{}2. Plan your answer structure to strictly follow these constraints, including the order and formatting of paragraphs and sentences.\par
\noindent{}3. When generating sentences, ensure the first word is exactly as required and that all lexical constraints hold.\par
\noindent{}4. When counting paragraphs, sentences, and words, be precise to meet the stated counts exactly.\par
\noindent{}5. For content-related tasks, ensure the explanation or solution is correct and logically sound.\par
\noindent{}6. For formatting constraints like enclosing the entire response in quotation marks or wrapping each word in brackets, apply these exactly and consistently.\par
\noindent{}7. If the user requests a palindrome or other special content, include it following the formatting and placement rules.\par
\noindent{}8. Do not omit or alter any of the user’s explicit constraints.\par
\par\vspace{.45\baselineskip}
\noindent{}Your final output must be the fully formatted, fully constraint-compliant response to the user query, with no deviations or omissions.\par
\par\vspace{.45\baselineskip}
\noindent{}Remember: Passing all explicit constraints exactly is mandatory. If you cannot meet all constraints, do not guess or approximate—respond exactly as requested.\par
\par\vspace{.45\baselineskip}
\noindent{}At the end, do not add any commentary or explanations—only output the response as per the instructions above.\par
\endgroup
\end{tcolorbox}

\Needspace{12\baselineskip}\subsection{GPT-4.1-mini — ifbench — MIPROv2}\label{app:evolution-record-37}

\begin{tcolorbox}[enhanced,breakable,lines before break=5,colback=black!1,colframe=black!20,boxrule=.35pt,arc=1mm,left=5pt,right=5pt,top=4pt,bottom=4pt,before skip=5pt,after skip=8pt,fonttitle=\bfseries\footnotesize,coltitle=black,colbacktitle=black!5,title={GPT-4.1-mini / ifbench / MIPROv2 / Selected instruction},title after break={GPT-4.1-mini / ifbench / MIPROv2 / Selected instruction (continued)}]
\begingroup\ttfamily\small\setlength{\parskip}{0pt}\raggedright
\noindent{}Given a `question` that may involve multiple constraints spanning lexical usage, style, formatting, and structural requirements, generate an `answer` that strictly adheres to all specified constraints simultaneously. Your response should be meticulously formatted, clearly demonstrating fulfillment of each constraint—such as mandated word frequencies, numeric placeholder usage, stylistic caps-lock requirements, or JSON-like metadata representations—while solving the problem or responding to the prompt accurately and coherently. Preserve a high level of precision in lexical choices, syntactic structure, and formatting details, and explicitly incorporate required placeholders or annotations as instructed. When applicable, include explanatory notes or warnings as specified, and ensure the final output is machine-friendly and easily parsed for constraint compliance.\par
\endgroup
\end{tcolorbox}

\Needspace{12\baselineskip}\subsection{GPT-4.1-mini — bbeh — GEPA}\label{app:evolution-record-38}

\begin{tcolorbox}[enhanced,breakable,lines before break=5,colback=black!1,colframe=black!20,boxrule=.35pt,arc=1mm,left=5pt,right=5pt,top=4pt,bottom=4pt,before skip=5pt,after skip=8pt,fonttitle=\bfseries\footnotesize,coltitle=black,colbacktitle=black!5,title={GPT-4.1-mini / bbeh / GEPA / Selected instruction},title after break={GPT-4.1-mini / bbeh / GEPA / Selected instruction (continued)}]
\begingroup\ttfamily\small\setlength{\parskip}{0pt}\raggedright
\noindent{}You will be given a reasoning problem involving detailed domain-specific rules or data, such as adjective ordering in a constructed language variant, custom alphabet sorting, or scheduling with complex time zone and availability constraints.\par
\par\vspace{.45\baselineskip}
\noindent{}Your task is to:\par
\par\vspace{.45\baselineskip}
\noindent{}1. Carefully analyze and fully understand the problem statement and all given data or example patterns. Extract explicit rules or orders from the examples if needed (e.g., adjective order categories and their sequence, or the modified alphabet order).\par
\par\vspace{.45\baselineskip}
\noindent{}2. Systematically apply these rules to the inputs. For adjective order problems, identify the categories of adjectives (material/\allowbreak{}nationality, size/\allowbreak{}shape, age/\allowbreak{}condition, color, opinion/\allowbreak{}quality, participial/\allowbreak{}purpose) and their correct order, then check each example against this sequence. For sorting with a custom alphabet, explicitly construct the new alphabet order mapping and sort all words accordingly. For scheduling problems, convert all times to a common reference timezone, convert availability and booked intervals into free intervals, consider any special constraints (e.g., clearing specific time slots, required free time after meetings), and find all possible meeting times starting only on the hour or half hour.\par
\par\vspace{.45\baselineskip}
\noindent{}3. When multiple answers are possible (e.g., multiple options with correct adjective order, multiple meeting slots), select and report all that apply, concatenating or listing them exactly as requested.\par
\par\vspace{.45\baselineskip}
\noindent{}4. Provide only the final answer in the exact format requested, with no additional explanation or commentary. If no valid answer exists, output the specified "none" option or "0, 0" as appropriate.\par
\par\vspace{.45\baselineskip}
\noindent{}5. Double-check your answer against the problem constraints to ensure correctness.\par
\par\vspace{.45\baselineskip}
\noindent{}Always end your response with a single line:  \par
\noindent{}The answer is: \$ANSWER  \par
\noindent{}where \$ANSWER is exactly the final answer string, no extra text.\par
\par\vspace{.45\baselineskip}
\noindent{}This approach ensures the reasoning is thorough, the domain-specific rules are correctly applied, and the final output matches the requested format exactly.\par
\endgroup
\end{tcolorbox}

\Needspace{12\baselineskip}\subsection{GPT-4.1-mini — bbeh — MIPROv2}\label{app:evolution-record-39}

\begin{tcolorbox}[enhanced,breakable,lines before break=5,colback=black!1,colframe=black!20,boxrule=.35pt,arc=1mm,left=5pt,right=5pt,top=4pt,bottom=4pt,before skip=5pt,after skip=8pt,fonttitle=\bfseries\footnotesize,coltitle=black,colbacktitle=black!5,title={GPT-4.1-mini / bbeh / MIPROv2 / Selected instruction},title after break={GPT-4.1-mini / bbeh / MIPROv2 / Selected instruction (continued)}]
\begingroup\ttfamily\small\setlength{\parskip}{0pt}\raggedright
\noindent{}You will be given a field named `question` that contains a reasoning problem or multi-part query. Your task is to carefully analyze the question, perform any necessary multi-step logical, numerical, or linguistic reasoning, and then produce a concise and accurate answer in the field `answer`. For questions with multiple sub-questions or multiple-choice selections, identify and select all correct choices as instructed, and format your answer accordingly (e.g., concatenated letters, comma-separated for multiple parts). For spatial or movement problems, track position changes accurately and report the final encountered object or value. For interpretation or humor understanding tasks, evaluate relevance and wit to select the best answer. Always ensure your response strictly follows the answer formatting and instructions provided in the question. The answer should be succinct, directly addressing the prompts without additional explanation unless specifically requested.\par
\endgroup
\end{tcolorbox}

\Needspace{12\baselineskip}\subsection{GPT-4.1-mini — hotpotqa — GEPA}\label{app:evolution-record-40}

\begin{tcolorbox}[enhanced,breakable,lines before break=5,colback=black!1,colframe=black!20,boxrule=.35pt,arc=1mm,left=5pt,right=5pt,top=4pt,bottom=4pt,before skip=5pt,after skip=8pt,fonttitle=\bfseries\footnotesize,coltitle=black,colbacktitle=black!5,title={GPT-4.1-mini / hotpotqa / GEPA / Selected instruction},title after break={GPT-4.1-mini / hotpotqa / GEPA / Selected instruction (continued)}]
\begingroup\ttfamily\small\setlength{\parskip}{0pt}\raggedright
\noindent{}You are provided with multiple factual context snippets containing detailed information about people, places, events, or entities. Following these is a question that requires extracting a concise, precise factual answer strictly from the given contexts.\par
\par\vspace{.45\baselineskip}
\noindent{}Your task is to:\par
\par\vspace{.45\baselineskip}
\noindent{}- Carefully read all context snippets and identify the exact information needed to answer the question.\par
\noindent{}- Extract the answer exactly as stated or clearly implied by the context, preserving the original wording for formal titles, names, dates, or specific terms. For example, use "Mayor of the City of New York" if that exact phrase appears, not a paraphrase.\par
\noindent{}- When the question asks for a specific entity (e.g., a person, place, or official role), answer with that entity’s canonical name or title as given in the context.\par
\noindent{}- In questions that compare or choose between options, answer with the shortest, most direct identifier matching the context (e.g., "Labrador Retriever" rather than a descriptive sentence).\par
\noindent{}- When the question requires bridging information across multiple snippets, synthesize the relevant facts but keep the answer concise and precise, reflecting the original phrasing.\par
\noindent{}- Do not add any information that is not supported by the provided contexts.\par
\noindent{}- Avoid unnecessary detail, explanations, or restatements; provide only the final answer.\par
\noindent{}- End your response with one line exactly: Answer: \$ANSWER.\par
\par\vspace{.45\baselineskip}
\noindent{}This approach ensures your answer aligns with exact-match or token-level evaluation metrics by matching the key terms and phrasing in the context.\par
\endgroup
\end{tcolorbox}

\Needspace{12\baselineskip}\subsection{GPT-4.1-mini — hotpotqa — MIPROv2}\label{app:evolution-record-41}

\begin{tcolorbox}[enhanced,breakable,lines before break=5,colback=black!1,colframe=black!20,boxrule=.35pt,arc=1mm,left=5pt,right=5pt,top=4pt,bottom=4pt,before skip=5pt,after skip=8pt,fonttitle=\bfseries\footnotesize,coltitle=black,colbacktitle=black!5,title={GPT-4.1-mini / hotpotqa / MIPROv2 / Selected instruction},title after break={GPT-4.1-mini / hotpotqa / MIPROv2 / Selected instruction (continued)}]
\begingroup\ttfamily\small\setlength{\parskip}{0pt}\raggedright
\noindent{}You are an elite fact-checker and research analyst working under a tight deadline to support a government inquiry that demands absolutely accurate, concise answers to complex multi-hop questions based on multiple disjoint Wikipedia-style text snippets. Given a question and its related context paragraphs, determine the precise, fact-based answer by integrating information across all relevant documents, paying close attention to entity disambiguation, temporal details, and comparative reasoning. Your response should be the definitive answer supported by the evidence in the context, provided clearly and without unnecessary elaboration.\par
\endgroup
\end{tcolorbox}

\Needspace{12\baselineskip}\subsection{Fine-tuning baseline: DPO}\label{app:evolution-record-42}

The preference-optimization baseline is DPO. The scores below are from its recorded test evaluations.\par

\begingroup\footnotesize\setlength{\tabcolsep}{3pt}
\setlength{\RecordTableWidth}{\dimexpr\linewidth-18pt\relax}
\begin{longtable}{>{\raggedright\arraybackslash}p{0.23166\RecordTableWidth}>{\raggedright\arraybackslash}p{0.23970\RecordTableWidth}>{\raggedright\arraybackslash}p{0.52864\RecordTableWidth}}
\hline
\textbf{Dataset} & \textbf{Verified DPO test (\%)} & \textbf{Source} \\
\hline\endfirsthead
\textbf{Dataset} & \textbf{Verified DPO test (\%)} & \textbf{Source} \\
\hline\endhead
\hline\endfoot
supergpqa & 50.00 & \nolinkurl{server_runs/qwen35_dpo_preference_s45_20260923/test/supergpqa_summary.json} \\
livebench\_\allowbreak{}math & 57.38 & \nolinkurl{server_runs/qwen35_dpo_preference_s45_20260923/test/livebench_math_summary.json} \\
ifbench & 40.33 & \nolinkurl{server_runs/qwen35_dpo_preference_s45_20260923/test/ifbench_summary.json} \\
bbeh & 21.67 & \nolinkurl{server_runs/qwen35_dpo_preference_s45_20260923/test/bbeh_summary.json} \\
hotpotqa & 40.00 & \nolinkurl{server_runs/qwen35_dpo_preference_s45_20260923/test/hotpotqa_summary.json} \\
\end{longtable}\endgroup

\Needspace{12\baselineskip}\subsection{Remaining provenance limitations}\label{app:evolution-record-43}

\noindent\hangindent=1em\hangafter=1\textbullet\ GPT-4.1-mini ifbench GEPA: ambiguous main-table run mapping\par

\endgroup

%% file: appendix/curated/tree_01.tex
\begin{center}
\begin{adjustbox}{max width=\linewidth,max totalheight=.48\textheight}
\begin{tikzpicture}[x=1cm,y=1cm,>=Stealth,
candidate/.style={circle,minimum size=10.5mm,inner sep=1pt,align=center,font=\sffamily\fontsize{8.2}{9.2}\selectfont,draw=blue!45!black,fill=blue!5,line width=.55pt}]
\node[candidate,draw=red!65!black,fill=orange!12,line width=1pt] (n0) at (5.490,0.000) {\textbf{C0}\\50.33};
\node[candidate,draw=red!65!black,fill=orange!12,line width=1pt] (n1) at (2.440,-1.450) {\textbf{C1}\\49.00};
\node[candidate] (n2) at (0.000,-2.900) {\textbf{C2}\\51.00};
\node[candidate] (n3) at (2.440,-2.900) {\textbf{C3}\\50.67};
\node[candidate] (n4) at (1.220,-4.350) {\textbf{C4}\\51.33};
\node[candidate] (n5) at (2.440,-4.350) {\textbf{C5}\\49.00};
\node[candidate] (n6) at (0.000,-4.350) {\textbf{C6}\\52.67};
\node[candidate] (n7) at (6.710,-1.450) {\textbf{C7}\\46.33};
\node[candidate] (n8) at (6.710,-2.900) {\textbf{C8}\\50.00};
\node[candidate,draw=red!65!black,fill=orange!12,line width=1pt,double,double distance=1pt,fill=orange!25,line width=1pt] (n9) at (4.880,-2.900) {\textbf{C9}\\54.67};
\node[candidate] (n10) at (8.540,-1.450) {\textbf{C10}\\49.67};
\node[candidate] (n11) at (3.660,-4.350) {\textbf{C11}\\54.00};
\node[candidate,draw=gray!60,fill=gray!5,dashed,text=gray!75!black] (n12) at (4.880,-4.350) {\textbf{C12}\\--};
\node[candidate,draw=gray!60,fill=gray!5,dashed,text=gray!75!black] (n13) at (6.100,-4.350) {\textbf{C13}\\51.00};
\node[candidate,draw=gray!60,fill=gray!5,dashed,text=gray!75!black] (n14) at (7.320,-4.350) {\textbf{C14}\\48.67};
\draw[->,red!65!black,line width=1.2pt] (n0) -- node[pos=.58,fill=white,inner sep=1.2pt,text=blue!65!black,font=\sffamily\bfseries\fontsize{7}{8}\selectfont] {R} (n1);
\draw[->,gray!55,line width=.6pt] (n1) -- node[pos=.58,fill=white,inner sep=1.2pt,text=blue!65!black,font=\sffamily\bfseries\fontsize{7}{8}\selectfont] {R} (n2);
\draw[->,gray!55,line width=.6pt] (n1) -- node[pos=.58,fill=white,inner sep=1.2pt,text=blue!65!black,font=\sffamily\bfseries\fontsize{7}{8}\selectfont] {R} (n3);
\draw[->,gray!55,line width=.6pt] (n3) -- node[pos=.58,fill=white,inner sep=1.2pt,text=teal!80!black,font=\sffamily\bfseries\fontsize{7}{8}\selectfont] {A} (n4);
\draw[->,gray!55,line width=.6pt] (n3) -- node[pos=.58,fill=white,inner sep=1.2pt,text=blue!65!black,font=\sffamily\bfseries\fontsize{7}{8}\selectfont] {R} (n5);
\draw[->,gray!55,line width=.6pt] (n2) -- node[pos=.58,fill=white,inner sep=1.2pt,text=teal!80!black,font=\sffamily\bfseries\fontsize{7}{8}\selectfont] {A} (n6);
\draw[->,gray!55,line width=.6pt] (n0) -- node[pos=.58,fill=white,inner sep=1.2pt,text=blue!65!black,font=\sffamily\bfseries\fontsize{7}{8}\selectfont] {R} (n7);
\draw[->,gray!55,line width=.6pt] (n7) -- node[pos=.58,fill=white,inner sep=1.2pt,text=teal!80!black,font=\sffamily\bfseries\fontsize{7}{8}\selectfont] {A} (n8);
\draw[->,red!65!black,line width=1.2pt] (n1) -- node[pos=.58,fill=white,inner sep=1.2pt,text=violet!80!black,font=\sffamily\bfseries\fontsize{7}{8}\selectfont] {S} (n9);
\draw[->,gray!55,line width=.6pt] (n0) -- node[pos=.58,fill=white,inner sep=1.2pt,text=blue!65!black,font=\sffamily\bfseries\fontsize{7}{8}\selectfont] {R} (n10);
\draw[->,gray!55,line width=.6pt] (n3) -- node[pos=.58,fill=white,inner sep=1.2pt,text=violet!80!black,font=\sffamily\bfseries\fontsize{7}{8}\selectfont] {S} (n11);
\draw[->,gray!55,line width=.6pt,dashed] (n9) -- node[pos=.58,fill=white,inner sep=1.2pt,text=blue!65!black,font=\sffamily\bfseries\fontsize{7}{8}\selectfont] {R} (n12);
\draw[->,gray!55,line width=.6pt,dashed] (n8) -- node[pos=.58,fill=white,inner sep=1.2pt,text=blue!65!black,font=\sffamily\bfseries\fontsize{7}{8}\selectfont] {R} (n13);
\draw[->,gray!55,line width=.6pt,dashed] (n8) -- node[pos=.58,fill=white,inner sep=1.2pt,text=violet!80!black,font=\sffamily\bfseries\fontsize{7}{8}\selectfont] {S} (n14);
\end{tikzpicture}
\end{adjustbox}
\end{center}

%% file: appendix/curated/tree_02.tex
\begin{center}
\begin{adjustbox}{max width=\linewidth,max totalheight=.48\textheight}
\begin{tikzpicture}[x=1cm,y=1cm,>=Stealth,
candidate/.style={circle,minimum size=10.5mm,inner sep=1pt,align=center,font=\sffamily\fontsize{8.2}{9.2}\selectfont,draw=blue!45!black,fill=blue!5,line width=.55pt}]
\node[candidate,draw=red!65!black,fill=orange!12,line width=1pt,double,double distance=1pt,fill=orange!25,line width=1pt] (n0) at (0.915,0.000) {\textbf{C0}\\50.33};
\node[candidate] (n1) at (0.915,-1.450) {\textbf{C1}\\50.00};
\node[candidate] (n2) at (0.000,-2.900) {\textbf{C2}\\48.33};
\node[candidate] (n3) at (1.830,-2.900) {\textbf{C3}\\49.00};
\node[candidate,draw=gray!60,fill=gray!5,dashed,text=gray!75!black] (n4) at (1.220,-4.350) {\textbf{C4}\\--};
\node[candidate] (n5) at (2.440,-4.350) {\textbf{C5}\\49.67};
\draw[->,gray!55,line width=.6pt] (n0) -- node[pos=.58,fill=white,inner sep=1.2pt,text=blue!65!black,font=\sffamily\bfseries\fontsize{7}{8}\selectfont] {R} (n1);
\draw[->,gray!55,line width=.6pt] (n1) -- node[pos=.58,fill=white,inner sep=1.2pt,text=blue!65!black,font=\sffamily\bfseries\fontsize{7}{8}\selectfont] {R} (n2);
\draw[->,gray!55,line width=.6pt] (n1) -- node[pos=.58,fill=white,inner sep=1.2pt,text=blue!65!black,font=\sffamily\bfseries\fontsize{7}{8}\selectfont] {R} (n3);
\draw[->,gray!55,line width=.6pt,dashed] (n3) -- node[pos=.58,fill=white,inner sep=1.2pt,text=teal!80!black,font=\sffamily\bfseries\fontsize{7}{8}\selectfont] {A} (n4);
\draw[->,gray!55,line width=.6pt] (n3) -- node[pos=.58,fill=white,inner sep=1.2pt,text=blue!65!black,font=\sffamily\bfseries\fontsize{7}{8}\selectfont] {R} (n5);
\end{tikzpicture}
\end{adjustbox}
\end{center}

%% file: appendix/curated/tree_03.tex
\begin{center}
\begin{adjustbox}{max width=\linewidth,max totalheight=.48\textheight}
\begin{tikzpicture}[x=1cm,y=1cm,>=Stealth,
candidate/.style={circle,minimum size=10.5mm,inner sep=1pt,align=center,font=\sffamily\fontsize{8.2}{9.2}\selectfont,draw=blue!45!black,fill=blue!5,line width=.55pt}]
\node[candidate,draw=red!65!black,fill=orange!12,line width=1pt] (n0) at (6.405,0.000) {\textbf{C0}\\63.11};
\node[candidate] (n1) at (3.355,-1.450) {\textbf{C1}\\54.92};
\node[candidate] (n2) at (0.610,-2.900) {\textbf{C2}\\62.30};
\node[candidate] (n3) at (3.355,-2.900) {\textbf{C3}\\59.84};
\node[candidate] (n4) at (2.440,-4.350) {\textbf{C4}\\69.67};
\node[candidate] (n5) at (4.270,-4.350) {\textbf{C5}\\60.66};
\node[candidate] (n6) at (0.000,-4.350) {\textbf{C6}\\71.31};
\node[candidate,draw=red!65!black,fill=orange!12,line width=1pt] (n7) at (9.455,-1.450) {\textbf{C7}\\61.48};
\node[candidate] (n8) at (7.930,-2.900) {\textbf{C8}\\65.57};
\node[candidate,draw=gray!60,fill=gray!5,dashed,text=gray!75!black] (n9) at (1.220,-4.350) {\textbf{C9}\\66.39};
\node[candidate] (n10) at (7.320,-4.350) {\textbf{C10}\\72.95};
\node[candidate] (n11) at (6.100,-2.900) {\textbf{C11}\\72.13};
\node[candidate,draw=gray!60,fill=gray!5,dashed,text=gray!75!black] (n12) at (7.320,-5.800) {\textbf{C12}\\72.13};
\node[candidate,draw=gray!60,fill=gray!5,dashed,text=gray!75!black] (n13) at (8.540,-4.350) {\textbf{C13}\\--};
\node[candidate,draw=red!65!black,fill=orange!12,line width=1pt,double,double distance=1pt,fill=orange!25,line width=1pt] (n14) at (10.980,-2.900) {\textbf{C14}\\73.77};
\node[candidate] (n15) at (9.760,-4.350) {\textbf{C15}\\68.03};
\node[candidate] (n16) at (3.660,-5.800) {\textbf{C16}\\70.49};
\node[candidate] (n17) at (4.880,-5.800) {\textbf{C17}\\67.21};
\node[candidate,draw=gray!60,fill=gray!5,dashed,text=gray!75!black] (n18) at (10.980,-4.350) {\textbf{C18}\\70.49};
\node[candidate,draw=gray!60,fill=gray!5,dashed,text=gray!75!black] (n19) at (12.200,-4.350) {\textbf{C19}\\--};
\draw[->,gray!55,line width=.6pt] (n0) -- node[pos=.58,fill=white,inner sep=1.2pt,text=blue!65!black,font=\sffamily\bfseries\fontsize{7}{8}\selectfont] {R} (n1);
\draw[->,gray!55,line width=.6pt] (n1) -- node[pos=.58,fill=white,inner sep=1.2pt,text=blue!65!black,font=\sffamily\bfseries\fontsize{7}{8}\selectfont] {R} (n2);
\draw[->,gray!55,line width=.6pt] (n1) -- node[pos=.58,fill=white,inner sep=1.2pt,text=blue!65!black,font=\sffamily\bfseries\fontsize{7}{8}\selectfont] {R} (n3);
\draw[->,gray!55,line width=.6pt] (n3) -- node[pos=.58,fill=white,inner sep=1.2pt,text=teal!80!black,font=\sffamily\bfseries\fontsize{7}{8}\selectfont] {A} (n4);
\draw[->,gray!55,line width=.6pt] (n3) -- node[pos=.58,fill=white,inner sep=1.2pt,text=blue!65!black,font=\sffamily\bfseries\fontsize{7}{8}\selectfont] {R} (n5);
\draw[->,gray!55,line width=.6pt] (n2) -- node[pos=.58,fill=white,inner sep=1.2pt,text=teal!80!black,font=\sffamily\bfseries\fontsize{7}{8}\selectfont] {A} (n6);
\draw[->,red!65!black,line width=1.2pt] (n0) -- node[pos=.58,fill=white,inner sep=1.2pt,text=blue!65!black,font=\sffamily\bfseries\fontsize{7}{8}\selectfont] {R} (n7);
\draw[->,gray!55,line width=.6pt] (n7) -- node[pos=.58,fill=white,inner sep=1.2pt,text=teal!80!black,font=\sffamily\bfseries\fontsize{7}{8}\selectfont] {A} (n8);
\draw[->,gray!55,line width=.6pt,dashed] (n2) -- node[pos=.58,fill=white,inner sep=1.2pt,text=violet!80!black,font=\sffamily\bfseries\fontsize{7}{8}\selectfont] {S} (n9);
\draw[->,gray!55,line width=.6pt] (n8) -- node[pos=.58,fill=white,inner sep=1.2pt,text=blue!65!black,font=\sffamily\bfseries\fontsize{7}{8}\selectfont] {R} (n10);
\draw[->,gray!55,line width=.6pt] (n1) -- node[pos=.58,fill=white,inner sep=1.2pt,text=violet!80!black,font=\sffamily\bfseries\fontsize{7}{8}\selectfont] {S} (n11);
\draw[->,gray!55,line width=.6pt,dashed] (n10) -- node[pos=.58,fill=white,inner sep=1.2pt,text=blue!65!black,font=\sffamily\bfseries\fontsize{7}{8}\selectfont] {R} (n12);
\draw[->,gray!55,line width=.6pt,dashed] (n8) -- node[pos=.58,fill=white,inner sep=1.2pt,text=blue!65!black,font=\sffamily\bfseries\fontsize{7}{8}\selectfont] {R} (n13);
\draw[->,red!65!black,line width=1.2pt] (n7) -- node[pos=.58,fill=white,inner sep=1.2pt,text=violet!80!black,font=\sffamily\bfseries\fontsize{7}{8}\selectfont] {S} (n14);
\draw[->,gray!55,line width=.6pt] (n14) -- node[pos=.58,fill=white,inner sep=1.2pt,text=blue!65!black,font=\sffamily\bfseries\fontsize{7}{8}\selectfont] {R} (n15);
\draw[->,gray!55,line width=.6pt] (n5) -- node[pos=.58,fill=white,inner sep=1.2pt,text=teal!80!black,font=\sffamily\bfseries\fontsize{7}{8}\selectfont] {A} (n16);
\draw[->,gray!55,line width=.6pt] (n5) -- node[pos=.58,fill=white,inner sep=1.2pt,text=violet!80!black,font=\sffamily\bfseries\fontsize{7}{8}\selectfont] {S} (n17);
\draw[->,gray!55,line width=.6pt,dashed] (n14) -- node[pos=.58,fill=white,inner sep=1.2pt,text=blue!65!black,font=\sffamily\bfseries\fontsize{7}{8}\selectfont] {R} (n18);
\draw[->,gray!55,line width=.6pt,dashed] (n14) -- node[pos=.58,fill=white,inner sep=1.2pt,text=violet!80!black,font=\sffamily\bfseries\fontsize{7}{8}\selectfont] {S} (n19);
\end{tikzpicture}
\end{adjustbox}
\end{center}

%% file: appendix/curated/tree_04.tex
\begin{center}
\begin{adjustbox}{max width=\linewidth,max totalheight=.48\textheight}
\begin{tikzpicture}[x=1cm,y=1cm,>=Stealth,
candidate/.style={circle,minimum size=10.5mm,inner sep=1pt,align=center,font=\sffamily\fontsize{8.2}{9.2}\selectfont,draw=blue!45!black,fill=blue!5,line width=.55pt}]
\node[candidate,draw=red!65!black,fill=orange!12,line width=1pt] (n0) at (5.795,0.000) {\textbf{C0}\\63.11};
\node[candidate,draw=red!65!black,fill=orange!12,line width=1pt] (n1) at (3.050,-1.450) {\textbf{C1}\\56.56};
\node[candidate] (n2) at (0.000,-2.900) {\textbf{C2}\\58.20};
\node[candidate,draw=red!65!black,fill=orange!12,line width=1pt] (n3) at (1.830,-2.900) {\textbf{C3}\\58.20};
\node[candidate,draw=red!65!black,fill=orange!12,line width=1pt] (n4) at (1.830,-4.350) {\textbf{C4}\\69.67};
\node[candidate,draw=gray!60,fill=gray!5,dashed,text=gray!75!black] (n5) at (1.220,-5.800) {\textbf{C5}\\--};
\node[candidate,draw=gray!60,fill=gray!5,dashed,text=gray!75!black] (n6) at (3.660,-2.900) {\textbf{C6}\\--};
\node[candidate] (n7) at (7.320,-1.450) {\textbf{C7}\\66.39};
\node[candidate,draw=red!65!black,fill=orange!12,line width=1pt,double,double distance=1pt,fill=orange!25,line width=1pt] (n8) at (2.440,-5.800) {\textbf{C8}\\72.95};
\node[candidate,draw=gray!60,fill=gray!5,dashed,text=gray!75!black] (n9) at (0.000,-4.350) {\textbf{C9}\\--};
\node[candidate] (n10) at (4.880,-2.900) {\textbf{C10}\\59.02};
\node[candidate] (n11) at (6.100,-2.900) {\textbf{C11}\\68.03};
\node[candidate] (n12) at (8.540,-1.450) {\textbf{C12}\\65.57};
\node[candidate,draw=gray!60,fill=gray!5,dashed,text=gray!75!black] (n13) at (2.440,-7.250) {\textbf{C13}\\--};
\draw[->,red!65!black,line width=1.2pt] (n0) -- node[pos=.58,fill=white,inner sep=1.2pt,text=blue!65!black,font=\sffamily\bfseries\fontsize{7}{8}\selectfont] {R} (n1);
\draw[->,gray!55,line width=.6pt] (n1) -- node[pos=.58,fill=white,inner sep=1.2pt,text=blue!65!black,font=\sffamily\bfseries\fontsize{7}{8}\selectfont] {R} (n2);
\draw[->,red!65!black,line width=1.2pt] (n1) -- node[pos=.58,fill=white,inner sep=1.2pt,text=blue!65!black,font=\sffamily\bfseries\fontsize{7}{8}\selectfont] {R} (n3);
\draw[->,red!65!black,line width=1.2pt] (n3) -- node[pos=.58,fill=white,inner sep=1.2pt,text=teal!80!black,font=\sffamily\bfseries\fontsize{7}{8}\selectfont] {A} (n4);
\draw[->,gray!55,line width=.6pt,dashed] (n4) -- node[pos=.58,fill=white,inner sep=1.2pt,text=blue!65!black,font=\sffamily\bfseries\fontsize{7}{8}\selectfont] {R} (n5);
\draw[->,gray!55,line width=.6pt,dashed] (n1) -- node[pos=.58,fill=white,inner sep=1.2pt,text=teal!80!black,font=\sffamily\bfseries\fontsize{7}{8}\selectfont] {A} (n6);
\draw[->,gray!55,line width=.6pt] (n0) -- node[pos=.58,fill=white,inner sep=1.2pt,text=blue!65!black,font=\sffamily\bfseries\fontsize{7}{8}\selectfont] {R} (n7);
\draw[->,red!65!black,line width=1.2pt] (n4) -- node[pos=.58,fill=white,inner sep=1.2pt,text=teal!80!black,font=\sffamily\bfseries\fontsize{7}{8}\selectfont] {A} (n8);
\draw[->,gray!55,line width=.6pt,dashed] (n2) -- node[pos=.58,fill=white,inner sep=1.2pt,text=blue!65!black,font=\sffamily\bfseries\fontsize{7}{8}\selectfont] {R} (n9);
\draw[->,gray!55,line width=.6pt] (n1) -- node[pos=.58,fill=white,inner sep=1.2pt,text=blue!65!black,font=\sffamily\bfseries\fontsize{7}{8}\selectfont] {R} (n10);
\draw[->,gray!55,line width=.6pt] (n1) -- node[pos=.58,fill=white,inner sep=1.2pt,text=teal!80!black,font=\sffamily\bfseries\fontsize{7}{8}\selectfont] {A} (n11);
\draw[->,gray!55,line width=.6pt] (n0) -- node[pos=.58,fill=white,inner sep=1.2pt,text=blue!65!black,font=\sffamily\bfseries\fontsize{7}{8}\selectfont] {R} (n12);
\draw[->,gray!55,line width=.6pt,dashed] (n8) -- node[pos=.58,fill=white,inner sep=1.2pt,text=blue!65!black,font=\sffamily\bfseries\fontsize{7}{8}\selectfont] {R} (n13);
\end{tikzpicture}
\end{adjustbox}
\end{center}

%% file: appendix/curated/tree_05.tex
\begin{center}
\begin{adjustbox}{max width=\linewidth,max totalheight=.48\textheight}
\begin{tikzpicture}[x=1cm,y=1cm,>=Stealth,
candidate/.style={circle,minimum size=10.5mm,inner sep=1pt,align=center,font=\sffamily\fontsize{8.2}{9.2}\selectfont,draw=blue!45!black,fill=blue!5,line width=.55pt}]
\node[candidate,draw=red!65!black,fill=orange!12,line width=1pt] (n0) at (9.607,0.000) {\textbf{C0}\\33.67};
\node[candidate,draw=red!65!black,fill=orange!12,line width=1pt] (n1) at (7.015,-1.450) {\textbf{C1}\\36.67};
\node[candidate,draw=red!65!black,fill=orange!12,line width=1pt] (n2) at (3.050,-2.900) {\textbf{C2}\\36.33};
\node[candidate] (n3) at (8.540,-2.900) {\textbf{C3}\\37.00};
\node[candidate,draw=gray!60,fill=gray!5,dashed,text=gray!75!black] (n4) at (7.320,-4.350) {\textbf{C4}\\--};
\node[candidate] (n5) at (8.540,-4.350) {\textbf{C5}\\36.00};
\node[candidate,draw=gray!60,fill=gray!5,dashed,text=gray!75!black] (n6) at (0.000,-4.350) {\textbf{C6}\\--};
\node[candidate,draw=gray!60,fill=gray!5,dashed,text=gray!75!black] (n7) at (12.200,-1.450) {\textbf{C7}\\--};
\node[candidate] (n8) at (9.760,-4.350) {\textbf{C8}\\37.33};
\node[candidate] (n9) at (8.540,-5.800) {\textbf{C9}\\36.00};
\node[candidate] (n10) at (10.980,-2.900) {\textbf{C10}\\35.67};
\node[candidate] (n11) at (1.220,-4.350) {\textbf{C11}\\36.33};
\node[candidate] (n12) at (9.760,-5.800) {\textbf{C12}\\37.67};
\node[candidate] (n13) at (2.440,-4.350) {\textbf{C13}\\36.00};
\node[candidate,draw=red!65!black,fill=orange!12,line width=1pt] (n14) at (4.270,-4.350) {\textbf{C14}\\38.67};
\node[candidate,draw=red!65!black,fill=orange!12,line width=1pt] (n15) at (4.270,-5.800) {\textbf{C15}\\39.33};
\node[candidate] (n16) at (6.100,-4.350) {\textbf{C16}\\34.00};
\node[candidate] (n17) at (9.760,-7.250) {\textbf{C17}\\39.00};
\node[candidate,draw=red!65!black,fill=orange!12,line width=1pt,double,double distance=1pt,fill=orange!25,line width=1pt] (n18) at (3.660,-7.250) {\textbf{C18}\\42.33};
\node[candidate] (n19) at (4.880,-7.250) {\textbf{C19}\\35.33};
\node[candidate] (n20) at (3.660,-8.700) {\textbf{C20}\\38.33};
\draw[->,red!65!black,line width=1.2pt] (n0) -- node[pos=.58,fill=white,inner sep=1.2pt,text=blue!65!black,font=\sffamily\bfseries\fontsize{7}{8}\selectfont] {R} (n1);
\draw[->,red!65!black,line width=1.2pt] (n1) -- node[pos=.58,fill=white,inner sep=1.2pt,text=blue!65!black,font=\sffamily\bfseries\fontsize{7}{8}\selectfont] {R} (n2);
\draw[->,gray!55,line width=.6pt] (n1) -- node[pos=.58,fill=white,inner sep=1.2pt,text=blue!65!black,font=\sffamily\bfseries\fontsize{7}{8}\selectfont] {R} (n3);
\draw[->,gray!55,line width=.6pt,dashed] (n3) -- node[pos=.58,fill=white,inner sep=1.2pt,text=teal!80!black,font=\sffamily\bfseries\fontsize{7}{8}\selectfont] {A} (n4);
\draw[->,gray!55,line width=.6pt] (n3) -- node[pos=.58,fill=white,inner sep=1.2pt,text=blue!65!black,font=\sffamily\bfseries\fontsize{7}{8}\selectfont] {R} (n5);
\draw[->,gray!55,line width=.6pt,dashed] (n2) -- node[pos=.58,fill=white,inner sep=1.2pt,text=teal!80!black,font=\sffamily\bfseries\fontsize{7}{8}\selectfont] {A} (n6);
\draw[->,gray!55,line width=.6pt,dashed] (n0) -- node[pos=.58,fill=white,inner sep=1.2pt,text=blue!65!black,font=\sffamily\bfseries\fontsize{7}{8}\selectfont] {R} (n7);
\draw[->,gray!55,line width=.6pt] (n3) -- node[pos=.58,fill=white,inner sep=1.2pt,text=teal!80!black,font=\sffamily\bfseries\fontsize{7}{8}\selectfont] {A} (n8);
\draw[->,gray!55,line width=.6pt] (n5) -- node[pos=.58,fill=white,inner sep=1.2pt,text=violet!80!black,font=\sffamily\bfseries\fontsize{7}{8}\selectfont] {S} (n9);
\draw[->,gray!55,line width=.6pt] (n1) -- node[pos=.58,fill=white,inner sep=1.2pt,text=blue!65!black,font=\sffamily\bfseries\fontsize{7}{8}\selectfont] {R} (n10);
\draw[->,gray!55,line width=.6pt] (n2) -- node[pos=.58,fill=white,inner sep=1.2pt,text=violet!80!black,font=\sffamily\bfseries\fontsize{7}{8}\selectfont] {S} (n11);
\draw[->,gray!55,line width=.6pt] (n8) -- node[pos=.58,fill=white,inner sep=1.2pt,text=blue!65!black,font=\sffamily\bfseries\fontsize{7}{8}\selectfont] {R} (n12);
\draw[->,gray!55,line width=.6pt] (n2) -- node[pos=.58,fill=white,inner sep=1.2pt,text=blue!65!black,font=\sffamily\bfseries\fontsize{7}{8}\selectfont] {R} (n13);
\draw[->,red!65!black,line width=1.2pt] (n2) -- node[pos=.58,fill=white,inner sep=1.2pt,text=violet!80!black,font=\sffamily\bfseries\fontsize{7}{8}\selectfont] {S} (n14);
\draw[->,red!65!black,line width=1.2pt] (n14) -- node[pos=.58,fill=white,inner sep=1.2pt,text=blue!65!black,font=\sffamily\bfseries\fontsize{7}{8}\selectfont] {R} (n15);
\draw[->,gray!55,line width=.6pt] (n2) -- node[pos=.58,fill=white,inner sep=1.2pt,text=teal!80!black,font=\sffamily\bfseries\fontsize{7}{8}\selectfont] {A} (n16);
\draw[->,gray!55,line width=.6pt] (n12) -- node[pos=.58,fill=white,inner sep=1.2pt,text=violet!80!black,font=\sffamily\bfseries\fontsize{7}{8}\selectfont] {S} (n17);
\draw[->,red!65!black,line width=1.2pt] (n15) -- node[pos=.58,fill=white,inner sep=1.2pt,text=blue!65!black,font=\sffamily\bfseries\fontsize{7}{8}\selectfont] {R} (n18);
\draw[->,gray!55,line width=.6pt] (n15) -- node[pos=.58,fill=white,inner sep=1.2pt,text=violet!80!black,font=\sffamily\bfseries\fontsize{7}{8}\selectfont] {S} (n19);
\draw[->,gray!55,line width=.6pt] (n18) -- node[pos=.58,fill=white,inner sep=1.2pt,text=violet!80!black,font=\sffamily\bfseries\fontsize{7}{8}\selectfont] {S} (n20);
\end{tikzpicture}
\end{adjustbox}
\end{center}

%% file: appendix/curated/tree_06.tex
\begin{center}
\begin{adjustbox}{max width=\linewidth,max totalheight=.48\textheight}
\begin{tikzpicture}[x=1cm,y=1cm,>=Stealth,
candidate/.style={circle,minimum size=10.5mm,inner sep=1pt,align=center,font=\sffamily\fontsize{8.2}{9.2}\selectfont,draw=blue!45!black,fill=blue!5,line width=.55pt}]
\node[candidate,draw=red!65!black,fill=orange!12,line width=1pt] (n0) at (5.032,0.000) {\textbf{C0}\\33.33};
\node[candidate,draw=red!65!black,fill=orange!12,line width=1pt] (n1) at (2.745,-1.450) {\textbf{C1}\\34.33};
\node[candidate,draw=red!65!black,fill=orange!12,line width=1pt,double,double distance=1pt,fill=orange!25,line width=1pt] (n2) at (0.610,-2.900) {\textbf{C2}\\38.33};
\node[candidate,draw=gray!60,fill=gray!5,dashed,text=gray!75!black] (n3) at (2.440,-2.900) {\textbf{C3}\\--};
\node[candidate] (n4) at (4.880,-2.900) {\textbf{C4}\\37.67};
\node[candidate] (n5) at (3.660,-4.350) {\textbf{C5}\\35.33};
\node[candidate,draw=gray!60,fill=gray!5,dashed,text=gray!75!black] (n6) at (0.000,-4.350) {\textbf{C6}\\--};
\node[candidate] (n7) at (7.320,-1.450) {\textbf{C7}\\35.33};
\node[candidate] (n8) at (4.880,-4.350) {\textbf{C8}\\36.33};
\node[candidate] (n9) at (1.220,-4.350) {\textbf{C9}\\37.00};
\node[candidate,draw=gray!60,fill=gray!5,dashed,text=gray!75!black] (n10) at (6.100,-4.350) {\textbf{C10}\\--};
\draw[->,red!65!black,line width=1.2pt] (n0) -- node[pos=.58,fill=white,inner sep=1.2pt,text=blue!65!black,font=\sffamily\bfseries\fontsize{7}{8}\selectfont] {R} (n1);
\draw[->,red!65!black,line width=1.2pt] (n1) -- node[pos=.58,fill=white,inner sep=1.2pt,text=blue!65!black,font=\sffamily\bfseries\fontsize{7}{8}\selectfont] {R} (n2);
\draw[->,gray!55,line width=.6pt,dashed] (n1) -- node[pos=.58,fill=white,inner sep=1.2pt,text=blue!65!black,font=\sffamily\bfseries\fontsize{7}{8}\selectfont] {R} (n3);
\draw[->,gray!55,line width=.6pt] (n1) -- node[pos=.58,fill=white,inner sep=1.2pt,text=teal!80!black,font=\sffamily\bfseries\fontsize{7}{8}\selectfont] {A} (n4);
\draw[->,gray!55,line width=.6pt] (n4) -- node[pos=.58,fill=white,inner sep=1.2pt,text=blue!65!black,font=\sffamily\bfseries\fontsize{7}{8}\selectfont] {R} (n5);
\draw[->,gray!55,line width=.6pt,dashed] (n2) -- node[pos=.58,fill=white,inner sep=1.2pt,text=teal!80!black,font=\sffamily\bfseries\fontsize{7}{8}\selectfont] {A} (n6);
\draw[->,gray!55,line width=.6pt] (n0) -- node[pos=.58,fill=white,inner sep=1.2pt,text=blue!65!black,font=\sffamily\bfseries\fontsize{7}{8}\selectfont] {R} (n7);
\draw[->,gray!55,line width=.6pt] (n4) -- node[pos=.58,fill=white,inner sep=1.2pt,text=teal!80!black,font=\sffamily\bfseries\fontsize{7}{8}\selectfont] {A} (n8);
\draw[->,gray!55,line width=.6pt] (n2) -- node[pos=.58,fill=white,inner sep=1.2pt,text=blue!65!black,font=\sffamily\bfseries\fontsize{7}{8}\selectfont] {R} (n9);
\draw[->,gray!55,line width=.6pt,dashed] (n4) -- node[pos=.58,fill=white,inner sep=1.2pt,text=blue!65!black,font=\sffamily\bfseries\fontsize{7}{8}\selectfont] {R} (n10);
\end{tikzpicture}
\end{adjustbox}
\end{center}

%% file: appendix/curated/tree_07.tex
\begin{center}
\begin{adjustbox}{max width=\linewidth,max totalheight=.48\textheight}
\begin{tikzpicture}[x=1cm,y=1cm,>=Stealth,
candidate/.style={circle,minimum size=10.5mm,inner sep=1pt,align=center,font=\sffamily\fontsize{8.2}{9.2}\selectfont,draw=blue!45!black,fill=blue!5,line width=.55pt}]
\node[candidate,draw=red!65!black,fill=orange!12,line width=1pt] (n0) at (5.185,0.000) {\textbf{C0}\\22.00};
\node[candidate,draw=red!65!black,fill=orange!12,line width=1pt] (n1) at (3.050,-1.450) {\textbf{C1}\\24.00};
\node[candidate] (n2) at (0.000,-2.900) {\textbf{C2}\\25.00};
\node[candidate] (n3) at (1.830,-2.900) {\textbf{C3}\\22.33};
\node[candidate] (n4) at (1.220,-4.350) {\textbf{C4}\\34.67};
\node[candidate] (n5) at (2.440,-4.350) {\textbf{C5}\\26.33};
\node[candidate] (n6) at (0.000,-4.350) {\textbf{C6}\\32.67};
\node[candidate] (n7) at (7.320,-1.450) {\textbf{C7}\\23.67};
\node[candidate] (n8) at (7.320,-2.900) {\textbf{C8}\\32.67};
\node[candidate] (n9) at (2.440,-5.800) {\textbf{C9}\\37.67};
\node[candidate] (n10) at (0.000,-5.800) {\textbf{C10}\\34.33};
\node[candidate] (n11) at (0.000,-7.250) {\textbf{C11}\\39.33};
\node[candidate] (n12) at (0.000,-8.700) {\textbf{C12}\\37.33};
\node[candidate,draw=gray!60,fill=gray!5,dashed,text=gray!75!black] (n13) at (3.660,-2.900) {\textbf{C13}\\24.00};
\node[candidate,draw=red!65!black,fill=orange!12,line width=1pt,double,double distance=1pt,fill=orange!25,line width=1pt] (n14) at (4.880,-2.900) {\textbf{C14}\\39.33};
\node[candidate] (n15) at (4.880,-4.350) {\textbf{C15}\\32.33};
\node[candidate] (n16) at (6.100,-2.900) {\textbf{C16}\\33.33};
\draw[->,red!65!black,line width=1.2pt] (n0) -- node[pos=.58,fill=white,inner sep=1.2pt,text=blue!65!black,font=\sffamily\bfseries\fontsize{7}{8}\selectfont] {R} (n1);
\draw[->,gray!55,line width=.6pt] (n1) -- node[pos=.58,fill=white,inner sep=1.2pt,text=blue!65!black,font=\sffamily\bfseries\fontsize{7}{8}\selectfont] {R} (n2);
\draw[->,gray!55,line width=.6pt] (n1) -- node[pos=.58,fill=white,inner sep=1.2pt,text=blue!65!black,font=\sffamily\bfseries\fontsize{7}{8}\selectfont] {R} (n3);
\draw[->,gray!55,line width=.6pt] (n3) -- node[pos=.58,fill=white,inner sep=1.2pt,text=teal!80!black,font=\sffamily\bfseries\fontsize{7}{8}\selectfont] {A} (n4);
\draw[->,gray!55,line width=.6pt] (n3) -- node[pos=.58,fill=white,inner sep=1.2pt,text=blue!65!black,font=\sffamily\bfseries\fontsize{7}{8}\selectfont] {R} (n5);
\draw[->,gray!55,line width=.6pt] (n2) -- node[pos=.58,fill=white,inner sep=1.2pt,text=teal!80!black,font=\sffamily\bfseries\fontsize{7}{8}\selectfont] {A} (n6);
\draw[->,gray!55,line width=.6pt] (n0) -- node[pos=.58,fill=white,inner sep=1.2pt,text=blue!65!black,font=\sffamily\bfseries\fontsize{7}{8}\selectfont] {R} (n7);
\draw[->,gray!55,line width=.6pt] (n7) -- node[pos=.58,fill=white,inner sep=1.2pt,text=teal!80!black,font=\sffamily\bfseries\fontsize{7}{8}\selectfont] {A} (n8);
\draw[->,gray!55,line width=.6pt] (n5) -- node[pos=.58,fill=white,inner sep=1.2pt,text=violet!80!black,font=\sffamily\bfseries\fontsize{7}{8}\selectfont] {S} (n9);
\draw[->,gray!55,line width=.6pt] (n6) -- node[pos=.58,fill=white,inner sep=1.2pt,text=blue!65!black,font=\sffamily\bfseries\fontsize{7}{8}\selectfont] {R} (n10);
\draw[->,gray!55,line width=.6pt] (n10) -- node[pos=.58,fill=white,inner sep=1.2pt,text=violet!80!black,font=\sffamily\bfseries\fontsize{7}{8}\selectfont] {S} (n11);
\draw[->,gray!55,line width=.6pt] (n11) -- node[pos=.58,fill=white,inner sep=1.2pt,text=blue!65!black,font=\sffamily\bfseries\fontsize{7}{8}\selectfont] {R} (n12);
\draw[->,gray!55,line width=.6pt,dashed] (n1) -- node[pos=.58,fill=white,inner sep=1.2pt,text=blue!65!black,font=\sffamily\bfseries\fontsize{7}{8}\selectfont] {R} (n13);
\draw[->,red!65!black,line width=1.2pt] (n1) -- node[pos=.58,fill=white,inner sep=1.2pt,text=violet!80!black,font=\sffamily\bfseries\fontsize{7}{8}\selectfont] {S} (n14);
\draw[->,gray!55,line width=.6pt] (n14) -- node[pos=.58,fill=white,inner sep=1.2pt,text=blue!65!black,font=\sffamily\bfseries\fontsize{7}{8}\selectfont] {R} (n15);
\draw[->,gray!55,line width=.6pt] (n1) -- node[pos=.58,fill=white,inner sep=1.2pt,text=teal!80!black,font=\sffamily\bfseries\fontsize{7}{8}\selectfont] {A} (n16);
\end{tikzpicture}
\end{adjustbox}
\end{center}

%% file: appendix/curated/tree_08.tex
\begin{center}
\begin{adjustbox}{max width=\linewidth,max totalheight=.48\textheight}
\begin{tikzpicture}[x=1cm,y=1cm,>=Stealth,
candidate/.style={circle,minimum size=10.5mm,inner sep=1pt,align=center,font=\sffamily\fontsize{8.2}{9.2}\selectfont,draw=blue!45!black,fill=blue!5,line width=.55pt}]
\node[candidate,draw=red!65!black,fill=orange!12,line width=1pt] (n0) at (5.643,0.000) {\textbf{C0}\\22.00};
\node[candidate,draw=red!65!black,fill=orange!12,line width=1pt] (n1) at (2.745,-1.450) {\textbf{C1}\\23.00};
\node[candidate,draw=gray!60,fill=gray!5,dashed,text=gray!75!black] (n2) at (0.000,-2.900) {\textbf{C2}\\--};
\node[candidate,draw=red!65!black,fill=orange!12,line width=1pt] (n3) at (2.745,-2.900) {\textbf{C3}\\27.67};
\node[candidate] (n4) at (5.490,-2.900) {\textbf{C4}\\31.67};
\node[candidate,draw=gray!60,fill=gray!5,dashed,text=gray!75!black] (n5) at (4.880,-4.350) {\textbf{C5}\\--};
\node[candidate,draw=red!65!black,fill=orange!12,line width=1pt] (n6) at (1.830,-4.350) {\textbf{C6}\\34.67};
\node[candidate] (n7) at (7.320,-1.450) {\textbf{C7}\\23.67};
\node[candidate,draw=red!65!black,fill=orange!12,line width=1pt,double,double distance=1pt,fill=orange!25,line width=1pt] (n8) at (1.220,-5.800) {\textbf{C8}\\35.67};
\node[candidate,draw=gray!60,fill=gray!5,dashed,text=gray!75!black] (n9) at (6.100,-4.350) {\textbf{C9}\\--};
\node[candidate] (n10) at (3.660,-4.350) {\textbf{C10}\\26.00};
\node[candidate] (n11) at (3.660,-5.800) {\textbf{C11}\\32.00};
\node[candidate] (n12) at (8.540,-1.450) {\textbf{C12}\\22.67};
\node[candidate] (n13) at (2.440,-5.800) {\textbf{C13}\\35.00};
\draw[->,red!65!black,line width=1.2pt] (n0) -- node[pos=.58,fill=white,inner sep=1.2pt,text=blue!65!black,font=\sffamily\bfseries\fontsize{7}{8}\selectfont] {R} (n1);
\draw[->,gray!55,line width=.6pt,dashed] (n1) -- node[pos=.58,fill=white,inner sep=1.2pt,text=blue!65!black,font=\sffamily\bfseries\fontsize{7}{8}\selectfont] {R} (n2);
\draw[->,red!65!black,line width=1.2pt] (n1) -- node[pos=.58,fill=white,inner sep=1.2pt,text=blue!65!black,font=\sffamily\bfseries\fontsize{7}{8}\selectfont] {R} (n3);
\draw[->,gray!55,line width=.6pt] (n1) -- node[pos=.58,fill=white,inner sep=1.2pt,text=teal!80!black,font=\sffamily\bfseries\fontsize{7}{8}\selectfont] {A} (n4);
\draw[->,gray!55,line width=.6pt,dashed] (n4) -- node[pos=.58,fill=white,inner sep=1.2pt,text=blue!65!black,font=\sffamily\bfseries\fontsize{7}{8}\selectfont] {R} (n5);
\draw[->,red!65!black,line width=1.2pt] (n3) -- node[pos=.58,fill=white,inner sep=1.2pt,text=teal!80!black,font=\sffamily\bfseries\fontsize{7}{8}\selectfont] {A} (n6);
\draw[->,gray!55,line width=.6pt] (n0) -- node[pos=.58,fill=white,inner sep=1.2pt,text=blue!65!black,font=\sffamily\bfseries\fontsize{7}{8}\selectfont] {R} (n7);
\draw[->,red!65!black,line width=1.2pt] (n6) -- node[pos=.58,fill=white,inner sep=1.2pt,text=teal!80!black,font=\sffamily\bfseries\fontsize{7}{8}\selectfont] {A} (n8);
\draw[->,gray!55,line width=.6pt,dashed] (n4) -- node[pos=.58,fill=white,inner sep=1.2pt,text=blue!65!black,font=\sffamily\bfseries\fontsize{7}{8}\selectfont] {R} (n9);
\draw[->,gray!55,line width=.6pt] (n3) -- node[pos=.58,fill=white,inner sep=1.2pt,text=blue!65!black,font=\sffamily\bfseries\fontsize{7}{8}\selectfont] {R} (n10);
\draw[->,gray!55,line width=.6pt] (n10) -- node[pos=.58,fill=white,inner sep=1.2pt,text=teal!80!black,font=\sffamily\bfseries\fontsize{7}{8}\selectfont] {A} (n11);
\draw[->,gray!55,line width=.6pt] (n0) -- node[pos=.58,fill=white,inner sep=1.2pt,text=blue!65!black,font=\sffamily\bfseries\fontsize{7}{8}\selectfont] {R} (n12);
\draw[->,gray!55,line width=.6pt] (n6) -- node[pos=.58,fill=white,inner sep=1.2pt,text=blue!65!black,font=\sffamily\bfseries\fontsize{7}{8}\selectfont] {R} (n13);
\end{tikzpicture}
\end{adjustbox}
\end{center}

%% file: appendix/curated/tree_09.tex
\begin{center}
\begin{adjustbox}{max width=\linewidth,max totalheight=.48\textheight}
\begin{tikzpicture}[x=1cm,y=1cm,>=Stealth,
candidate/.style={circle,minimum size=10.5mm,inner sep=1pt,align=center,font=\sffamily\fontsize{8.2}{9.2}\selectfont,draw=blue!45!black,fill=blue!5,line width=.55pt}]
\node[candidate,draw=red!65!black,fill=orange!12,line width=1pt] (n0) at (3.203,0.000) {\textbf{C0}\\41.00};
\node[candidate,draw=red!65!black,fill=orange!12,line width=1pt] (n1) at (1.525,-1.450) {\textbf{C1}\\43.33};
\node[candidate] (n2) at (0.610,-2.900) {\textbf{C2}\\43.00};
\node[candidate] (n3) at (3.660,-1.450) {\textbf{C3}\\42.67};
\node[candidate] (n4) at (4.880,-1.450) {\textbf{C4}\\44.67};
\node[candidate,draw=red!65!black,fill=orange!12,line width=1pt,double,double distance=1pt,fill=orange!25,line width=1pt] (n5) at (2.440,-2.900) {\textbf{C5}\\49.67};
\node[candidate] (n6) at (0.000,-4.350) {\textbf{C6}\\43.67};
\node[candidate] (n7) at (2.440,-4.350) {\textbf{C7}\\35.67};
\node[candidate] (n8) at (1.220,-4.350) {\textbf{C8}\\47.33};
\node[candidate] (n9) at (3.660,-2.900) {\textbf{C9}\\44.67};
\node[candidate] (n10) at (0.000,-5.800) {\textbf{C10}\\39.33};
\draw[->,red!65!black,line width=1.2pt] (n0) -- node[pos=.58,fill=white,inner sep=1.2pt,text=blue!65!black,font=\sffamily\bfseries\fontsize{7}{8}\selectfont] {R} (n1);
\draw[->,gray!55,line width=.6pt] (n1) -- node[pos=.58,fill=white,inner sep=1.2pt,text=blue!65!black,font=\sffamily\bfseries\fontsize{7}{8}\selectfont] {R} (n2);
\draw[->,gray!55,line width=.6pt] (n0) -- node[pos=.58,fill=white,inner sep=1.2pt,text=blue!65!black,font=\sffamily\bfseries\fontsize{7}{8}\selectfont] {R} (n3);
\draw[->,gray!55,line width=.6pt] (n0) -- node[pos=.58,fill=white,inner sep=1.2pt,text=blue!65!black,font=\sffamily\bfseries\fontsize{7}{8}\selectfont] {R} (n4);
\draw[->,red!65!black,line width=1.2pt] (n1) -- node[pos=.58,fill=white,inner sep=1.2pt,text=violet!80!black,font=\sffamily\bfseries\fontsize{7}{8}\selectfont] {S} (n5);
\draw[->,gray!55,line width=.6pt] (n2) -- node[pos=.58,fill=white,inner sep=1.2pt,text=teal!80!black,font=\sffamily\bfseries\fontsize{7}{8}\selectfont] {A} (n6);
\draw[->,gray!55,line width=.6pt] (n5) -- node[pos=.58,fill=white,inner sep=1.2pt,text=teal!80!black,font=\sffamily\bfseries\fontsize{7}{8}\selectfont] {A} (n7);
\draw[->,gray!55,line width=.6pt] (n2) -- node[pos=.58,fill=white,inner sep=1.2pt,text=violet!80!black,font=\sffamily\bfseries\fontsize{7}{8}\selectfont] {S} (n8);
\draw[->,gray!55,line width=.6pt] (n3) -- node[pos=.58,fill=white,inner sep=1.2pt,text=teal!80!black,font=\sffamily\bfseries\fontsize{7}{8}\selectfont] {A} (n9);
\draw[->,gray!55,line width=.6pt] (n6) -- node[pos=.58,fill=white,inner sep=1.2pt,text=teal!80!black,font=\sffamily\bfseries\fontsize{7}{8}\selectfont] {A} (n10);
\end{tikzpicture}
\end{adjustbox}
\end{center}

%% file: appendix/curated/tree_10.tex
\begin{center}
\begin{adjustbox}{max width=\linewidth,max totalheight=.48\textheight}
\begin{tikzpicture}[x=1cm,y=1cm,>=Stealth,
candidate/.style={circle,minimum size=10.5mm,inner sep=1pt,align=center,font=\sffamily\fontsize{8.2}{9.2}\selectfont,draw=blue!45!black,fill=blue!5,line width=.55pt}]
\node[candidate,draw=red!65!black,fill=orange!12,line width=1pt] (n0) at (2.592,0.000) {\textbf{C0}\\41.00};
\node[candidate,draw=gray!60,fill=gray!5,dashed,text=gray!75!black] (n1) at (0.000,-1.450) {\textbf{C1}\\--};
\node[candidate,draw=red!65!black,fill=orange!12,line width=1pt] (n2) at (5.185,-1.450) {\textbf{C2}\\35.00};
\node[candidate] (n3) at (1.830,-2.900) {\textbf{C3}\\34.67};
\node[candidate] (n4) at (4.270,-2.900) {\textbf{C4}\\35.67};
\node[candidate,draw=red!65!black,fill=orange!12,line width=1pt] (n5) at (6.100,-2.900) {\textbf{C5}\\35.33};
\node[candidate] (n6) at (7.320,-2.900) {\textbf{C6}\\33.67};
\node[candidate] (n7) at (1.220,-4.350) {\textbf{C7}\\35.33};
\node[candidate,draw=red!65!black,fill=orange!12,line width=1pt,double,double distance=1pt,fill=orange!25,line width=1pt] (n8) at (6.100,-4.350) {\textbf{C8}\\44.00};
\node[candidate] (n9) at (2.440,-4.350) {\textbf{C9}\\33.00};
\node[candidate] (n10) at (7.320,-4.350) {\textbf{C10}\\35.00};
\node[candidate] (n11) at (8.540,-2.900) {\textbf{C11}\\39.33};
\node[candidate,draw=gray!60,fill=gray!5,dashed,text=gray!75!black] (n12) at (3.660,-4.350) {\textbf{C12}\\--};
\node[candidate] (n13) at (4.880,-4.350) {\textbf{C13}\\34.67};
\draw[->,gray!55,line width=.6pt,dashed] (n0) -- node[pos=.58,fill=white,inner sep=1.2pt,text=blue!65!black,font=\sffamily\bfseries\fontsize{7}{8}\selectfont] {R} (n1);
\draw[->,red!65!black,line width=1.2pt] (n0) -- node[pos=.58,fill=white,inner sep=1.2pt,text=blue!65!black,font=\sffamily\bfseries\fontsize{7}{8}\selectfont] {R} (n2);
\draw[->,gray!55,line width=.6pt] (n2) -- node[pos=.58,fill=white,inner sep=1.2pt,text=blue!65!black,font=\sffamily\bfseries\fontsize{7}{8}\selectfont] {R} (n3);
\draw[->,gray!55,line width=.6pt] (n2) -- node[pos=.58,fill=white,inner sep=1.2pt,text=teal!80!black,font=\sffamily\bfseries\fontsize{7}{8}\selectfont] {A} (n4);
\draw[->,red!65!black,line width=1.2pt] (n2) -- node[pos=.58,fill=white,inner sep=1.2pt,text=teal!80!black,font=\sffamily\bfseries\fontsize{7}{8}\selectfont] {A} (n5);
\draw[->,gray!55,line width=.6pt] (n2) -- node[pos=.58,fill=white,inner sep=1.2pt,text=blue!65!black,font=\sffamily\bfseries\fontsize{7}{8}\selectfont] {R} (n6);
\draw[->,gray!55,line width=.6pt] (n3) -- node[pos=.58,fill=white,inner sep=1.2pt,text=blue!65!black,font=\sffamily\bfseries\fontsize{7}{8}\selectfont] {R} (n7);
\draw[->,red!65!black,line width=1.2pt] (n5) -- node[pos=.58,fill=white,inner sep=1.2pt,text=teal!80!black,font=\sffamily\bfseries\fontsize{7}{8}\selectfont] {A} (n8);
\draw[->,gray!55,line width=.6pt] (n3) -- node[pos=.58,fill=white,inner sep=1.2pt,text=blue!65!black,font=\sffamily\bfseries\fontsize{7}{8}\selectfont] {R} (n9);
\draw[->,gray!55,line width=.6pt] (n6) -- node[pos=.58,fill=white,inner sep=1.2pt,text=blue!65!black,font=\sffamily\bfseries\fontsize{7}{8}\selectfont] {R} (n10);
\draw[->,gray!55,line width=.6pt] (n2) -- node[pos=.58,fill=white,inner sep=1.2pt,text=blue!65!black,font=\sffamily\bfseries\fontsize{7}{8}\selectfont] {R} (n11);
\draw[->,gray!55,line width=.6pt,dashed] (n4) -- node[pos=.58,fill=white,inner sep=1.2pt,text=blue!65!black,font=\sffamily\bfseries\fontsize{7}{8}\selectfont] {R} (n12);
\draw[->,gray!55,line width=.6pt] (n4) -- node[pos=.58,fill=white,inner sep=1.2pt,text=blue!65!black,font=\sffamily\bfseries\fontsize{7}{8}\selectfont] {R} (n13);
\end{tikzpicture}
\end{adjustbox}
\end{center}

%% file: appendix/curated/tree_11.tex
\begin{center}
\begin{adjustbox}{max width=\linewidth,max totalheight=.48\textheight}
\begin{tikzpicture}[x=1cm,y=1cm,>=Stealth,
candidate/.style={circle,minimum size=10.5mm,inner sep=1pt,align=center,font=\sffamily\fontsize{8.2}{9.2}\selectfont,draw=blue!45!black,fill=blue!5,line width=.55pt}]
\node[candidate,draw=red!65!black,fill=orange!12,line width=1pt] (n0) at (3.355,0.000) {\textbf{C0}\\48.33};
\node[candidate,draw=red!65!black,fill=orange!12,line width=1pt] (n1) at (3.355,-1.450) {\textbf{C1}\\51.00};
\node[candidate] (n2) at (1.220,-2.900) {\textbf{C2}\\50.00};
\node[candidate] (n3) at (3.660,-2.900) {\textbf{C3}\\51.33};
\node[candidate,draw=gray!60,fill=gray!5,dashed,text=gray!75!black] (n4) at (0.000,-4.350) {\textbf{C4}\\--};
\node[candidate,draw=red!65!black,fill=orange!12,line width=1pt] (n5) at (5.490,-2.900) {\textbf{C5}\\50.00};
\node[candidate,draw=red!65!black,fill=orange!12,line width=1pt,double,double distance=1pt,fill=orange!25,line width=1pt] (n6) at (4.880,-4.350) {\textbf{C6}\\53.33};
\node[candidate] (n7) at (1.220,-4.350) {\textbf{C7}\\49.67};
\node[candidate] (n8) at (2.440,-4.350) {\textbf{C8}\\51.33};
\node[candidate] (n9) at (6.100,-4.350) {\textbf{C9}\\51.00};
\node[candidate,draw=gray!60,fill=gray!5,dashed,text=gray!75!black] (n10) at (3.660,-4.350) {\textbf{C10}\\51.00};
\node[candidate] (n11) at (2.440,-5.800) {\textbf{C11}\\52.33};
\draw[->,red!65!black,line width=1.2pt] (n0) -- node[pos=.58,fill=white,inner sep=1.2pt,text=blue!65!black,font=\sffamily\bfseries\fontsize{7}{8}\selectfont] {R} (n1);
\draw[->,gray!55,line width=.6pt] (n1) -- node[pos=.58,fill=white,inner sep=1.2pt,text=blue!65!black,font=\sffamily\bfseries\fontsize{7}{8}\selectfont] {R} (n2);
\draw[->,gray!55,line width=.6pt] (n1) -- node[pos=.58,fill=white,inner sep=1.2pt,text=teal!80!black,font=\sffamily\bfseries\fontsize{7}{8}\selectfont] {A} (n3);
\draw[->,gray!55,line width=.6pt,dashed] (n2) -- node[pos=.58,fill=white,inner sep=1.2pt,text=teal!80!black,font=\sffamily\bfseries\fontsize{7}{8}\selectfont] {A} (n4);
\draw[->,red!65!black,line width=1.2pt] (n1) -- node[pos=.58,fill=white,inner sep=1.2pt,text=violet!80!black,font=\sffamily\bfseries\fontsize{7}{8}\selectfont] {S} (n5);
\draw[->,red!65!black,line width=1.2pt] (n5) -- node[pos=.58,fill=white,inner sep=1.2pt,text=violet!80!black,font=\sffamily\bfseries\fontsize{7}{8}\selectfont] {S} (n6);
\draw[->,gray!55,line width=.6pt] (n2) -- node[pos=.58,fill=white,inner sep=1.2pt,text=violet!80!black,font=\sffamily\bfseries\fontsize{7}{8}\selectfont] {S} (n7);
\draw[->,gray!55,line width=.6pt] (n2) -- node[pos=.58,fill=white,inner sep=1.2pt,text=violet!80!black,font=\sffamily\bfseries\fontsize{7}{8}\selectfont] {S} (n8);
\draw[->,gray!55,line width=.6pt] (n5) -- node[pos=.58,fill=white,inner sep=1.2pt,text=violet!80!black,font=\sffamily\bfseries\fontsize{7}{8}\selectfont] {S} (n9);
\draw[->,gray!55,line width=.6pt,dashed] (n3) -- node[pos=.58,fill=white,inner sep=1.2pt,text=blue!65!black,font=\sffamily\bfseries\fontsize{7}{8}\selectfont] {R} (n10);
\draw[->,gray!55,line width=.6pt] (n8) -- node[pos=.58,fill=white,inner sep=1.2pt,text=violet!80!black,font=\sffamily\bfseries\fontsize{7}{8}\selectfont] {S} (n11);
\end{tikzpicture}
\end{adjustbox}
\end{center}

%% file: appendix/curated/tree_12.tex
\begin{center}
\begin{adjustbox}{max width=\linewidth,max totalheight=.48\textheight}
\begin{tikzpicture}[x=1cm,y=1cm,>=Stealth,
candidate/.style={circle,minimum size=10.5mm,inner sep=1pt,align=center,font=\sffamily\fontsize{8.2}{9.2}\selectfont,draw=blue!45!black,fill=blue!5,line width=.55pt}]
\node[candidate,draw=red!65!black,fill=orange!12,line width=1pt] (n0) at (2.440,0.000) {\textbf{C0}\\48.33};
\node[candidate,draw=red!65!black,fill=orange!12,line width=1pt] (n1) at (2.440,-1.450) {\textbf{C1}\\50.33};
\node[candidate,draw=gray!60,fill=gray!5,dashed,text=gray!75!black] (n2) at (0.000,-2.900) {\textbf{C2}\\--};
\node[candidate,draw=red!65!black,fill=orange!12,line width=1pt] (n3) at (2.135,-2.900) {\textbf{C3}\\50.67};
\node[candidate,draw=red!65!black,fill=orange!12,line width=1pt] (n4) at (2.135,-4.350) {\textbf{C4}\\51.00};
\node[candidate,draw=gray!60,fill=gray!5,dashed,text=gray!75!black] (n5) at (4.880,-2.900) {\textbf{C5}\\--};
\node[candidate,draw=red!65!black,fill=orange!12,line width=1pt,double,double distance=1pt,fill=orange!25,line width=1pt] (n6) at (2.135,-5.800) {\textbf{C6}\\52.00};
\node[candidate,draw=gray!60,fill=gray!5,dashed,text=gray!75!black] (n7) at (1.220,-7.250) {\textbf{C7}\\51.33};
\node[candidate] (n8) at (3.050,-7.250) {\textbf{C8}\\49.67};
\node[candidate] (n9) at (3.050,-8.700) {\textbf{C9}\\48.33};
\node[candidate] (n10) at (2.440,-10.150) {\textbf{C10}\\48.33};
\node[candidate] (n11) at (3.660,-10.150) {\textbf{C11}\\48.67};
\draw[->,red!65!black,line width=1.2pt] (n0) -- node[pos=.58,fill=white,inner sep=1.2pt,text=blue!65!black,font=\sffamily\bfseries\fontsize{7}{8}\selectfont] {R} (n1);
\draw[->,gray!55,line width=.6pt,dashed] (n1) -- node[pos=.58,fill=white,inner sep=1.2pt,text=blue!65!black,font=\sffamily\bfseries\fontsize{7}{8}\selectfont] {R} (n2);
\draw[->,red!65!black,line width=1.2pt] (n1) -- node[pos=.58,fill=white,inner sep=1.2pt,text=teal!80!black,font=\sffamily\bfseries\fontsize{7}{8}\selectfont] {A} (n3);
\draw[->,red!65!black,line width=1.2pt] (n3) -- node[pos=.58,fill=white,inner sep=1.2pt,text=teal!80!black,font=\sffamily\bfseries\fontsize{7}{8}\selectfont] {A} (n4);
\draw[->,gray!55,line width=.6pt,dashed] (n1) -- node[pos=.58,fill=white,inner sep=1.2pt,text=teal!80!black,font=\sffamily\bfseries\fontsize{7}{8}\selectfont] {A} (n5);
\draw[->,red!65!black,line width=1.2pt] (n4) -- node[pos=.58,fill=white,inner sep=1.2pt,text=blue!65!black,font=\sffamily\bfseries\fontsize{7}{8}\selectfont] {R} (n6);
\draw[->,gray!55,line width=.6pt,dashed] (n6) -- node[pos=.58,fill=white,inner sep=1.2pt,text=blue!65!black,font=\sffamily\bfseries\fontsize{7}{8}\selectfont] {R} (n7);
\draw[->,gray!55,line width=.6pt] (n6) -- node[pos=.58,fill=white,inner sep=1.2pt,text=blue!65!black,font=\sffamily\bfseries\fontsize{7}{8}\selectfont] {R} (n8);
\draw[->,gray!55,line width=.6pt] (n8) -- node[pos=.58,fill=white,inner sep=1.2pt,text=blue!65!black,font=\sffamily\bfseries\fontsize{7}{8}\selectfont] {R} (n9);
\draw[->,gray!55,line width=.6pt] (n9) -- node[pos=.58,fill=white,inner sep=1.2pt,text=blue!65!black,font=\sffamily\bfseries\fontsize{7}{8}\selectfont] {R} (n10);
\draw[->,gray!55,line width=.6pt] (n9) -- node[pos=.58,fill=white,inner sep=1.2pt,text=blue!65!black,font=\sffamily\bfseries\fontsize{7}{8}\selectfont] {R} (n11);
\end{tikzpicture}
\end{adjustbox}
\end{center}

%% file: appendix/curated/tree_13.tex
\begin{center}
\begin{adjustbox}{max width=\linewidth,max totalheight=.48\textheight}
\begin{tikzpicture}[x=1cm,y=1cm,>=Stealth,
candidate/.style={circle,minimum size=10.5mm,inner sep=1pt,align=center,font=\sffamily\fontsize{8.2}{9.2}\selectfont,draw=blue!45!black,fill=blue!5,line width=.55pt}]
\node[candidate,draw=red!65!black,fill=orange!12,line width=1pt] (n0) at (6.405,0.000) {\textbf{C0}\\54.92};
\node[candidate,draw=red!65!black,fill=orange!12,line width=1pt] (n1) at (3.050,-1.450) {\textbf{C1}\\58.20};
\node[candidate] (n2) at (9.760,-1.450) {\textbf{C2}\\52.46};
\node[candidate,draw=red!65!black,fill=orange!12,line width=1pt] (n3) at (0.000,-2.900) {\textbf{C3}\\62.30};
\node[candidate,draw=red!65!black,fill=orange!12,line width=1pt] (n4) at (0.000,-4.350) {\textbf{C4}\\64.75};
\node[candidate] (n5) at (1.830,-2.900) {\textbf{C5}\\57.38};
\node[candidate] (n6) at (6.100,-2.900) {\textbf{C6}\\62.30};
\node[candidate] (n7) at (3.660,-4.350) {\textbf{C7}\\60.66};
\node[candidate] (n8) at (1.220,-4.350) {\textbf{C8}\\57.38};
\node[candidate,draw=red!65!black,fill=orange!12,line width=1pt,double,double distance=1pt,fill=orange!25,line width=1pt] (n9) at (0.000,-5.800) {\textbf{C9}\\68.03};
\node[candidate] (n10) at (4.880,-4.350) {\textbf{C10}\\61.48};
\node[candidate,draw=gray!60,fill=gray!5,dashed,text=gray!75!black] (n11) at (2.440,-4.350) {\textbf{C11}\\52.46};
\node[candidate] (n12) at (6.100,-4.350) {\textbf{C12}\\56.56};
\node[candidate,draw=gray!60,fill=gray!5,dashed,text=gray!75!black] (n13) at (7.320,-4.350) {\textbf{C13}\\--};
\node[candidate] (n14) at (8.540,-4.350) {\textbf{C14}\\65.57};
\draw[->,red!65!black,line width=1.2pt] (n0) -- node[pos=.58,fill=white,inner sep=1.2pt,text=blue!65!black,font=\sffamily\bfseries\fontsize{7}{8}\selectfont] {R} (n1);
\draw[->,gray!55,line width=.6pt] (n0) -- node[pos=.58,fill=white,inner sep=1.2pt,text=blue!65!black,font=\sffamily\bfseries\fontsize{7}{8}\selectfont] {R} (n2);
\draw[->,red!65!black,line width=1.2pt] (n1) -- node[pos=.58,fill=white,inner sep=1.2pt,text=blue!65!black,font=\sffamily\bfseries\fontsize{7}{8}\selectfont] {R} (n3);
\draw[->,red!65!black,line width=1.2pt] (n3) -- node[pos=.58,fill=white,inner sep=1.2pt,text=teal!80!black,font=\sffamily\bfseries\fontsize{7}{8}\selectfont] {A} (n4);
\draw[->,gray!55,line width=.6pt] (n1) -- node[pos=.58,fill=white,inner sep=1.2pt,text=violet!80!black,font=\sffamily\bfseries\fontsize{7}{8}\selectfont] {S} (n5);
\draw[->,gray!55,line width=.6pt] (n1) -- node[pos=.58,fill=white,inner sep=1.2pt,text=violet!80!black,font=\sffamily\bfseries\fontsize{7}{8}\selectfont] {S} (n6);
\draw[->,gray!55,line width=.6pt] (n6) -- node[pos=.58,fill=white,inner sep=1.2pt,text=blue!65!black,font=\sffamily\bfseries\fontsize{7}{8}\selectfont] {R} (n7);
\draw[->,gray!55,line width=.6pt] (n5) -- node[pos=.58,fill=white,inner sep=1.2pt,text=teal!80!black,font=\sffamily\bfseries\fontsize{7}{8}\selectfont] {A} (n8);
\draw[->,red!65!black,line width=1.2pt] (n4) -- node[pos=.58,fill=white,inner sep=1.2pt,text=teal!80!black,font=\sffamily\bfseries\fontsize{7}{8}\selectfont] {A} (n9);
\draw[->,gray!55,line width=.6pt] (n6) -- node[pos=.58,fill=white,inner sep=1.2pt,text=blue!65!black,font=\sffamily\bfseries\fontsize{7}{8}\selectfont] {R} (n10);
\draw[->,gray!55,line width=.6pt,dashed] (n5) -- node[pos=.58,fill=white,inner sep=1.2pt,text=blue!65!black,font=\sffamily\bfseries\fontsize{7}{8}\selectfont] {R} (n11);
\draw[->,gray!55,line width=.6pt] (n6) -- node[pos=.58,fill=white,inner sep=1.2pt,text=violet!80!black,font=\sffamily\bfseries\fontsize{7}{8}\selectfont] {S} (n12);
\draw[->,gray!55,line width=.6pt,dashed] (n6) -- node[pos=.58,fill=white,inner sep=1.2pt,text=blue!65!black,font=\sffamily\bfseries\fontsize{7}{8}\selectfont] {R} (n13);
\draw[->,gray!55,line width=.6pt] (n6) -- node[pos=.58,fill=white,inner sep=1.2pt,text=blue!65!black,font=\sffamily\bfseries\fontsize{7}{8}\selectfont] {R} (n14);
\end{tikzpicture}
\end{adjustbox}
\end{center}

%% file: appendix/curated/tree_14.tex
\begin{center}
\begin{adjustbox}{max width=\linewidth,max totalheight=.48\textheight}
\begin{tikzpicture}[x=1cm,y=1cm,>=Stealth,
candidate/.style={circle,minimum size=10.5mm,inner sep=1pt,align=center,font=\sffamily\fontsize{8.2}{9.2}\selectfont,draw=blue!45!black,fill=blue!5,line width=.55pt}]
\node[candidate,draw=red!65!black,fill=orange!12,line width=1pt] (n0) at (5.032,0.000) {\textbf{C0}\\54.92};
\node[candidate,draw=red!65!black,fill=orange!12,line width=1pt] (n1) at (1.525,-1.450) {\textbf{C1}\\56.56};
\node[candidate] (n2) at (6.100,-1.450) {\textbf{C2}\\57.38};
\node[candidate,draw=gray!60,fill=gray!5,dashed,text=gray!75!black] (n3) at (0.000,-2.900) {\textbf{C3}\\--};
\node[candidate] (n4) at (6.100,-2.900) {\textbf{C4}\\59.84};
\node[candidate,draw=red!65!black,fill=orange!12,line width=1pt] (n5) at (3.050,-2.900) {\textbf{C5}\\59.02};
\node[candidate,draw=red!65!black,fill=orange!12,line width=1pt] (n6) at (3.050,-4.350) {\textbf{C6}\\63.11};
\node[candidate,draw=gray!60,fill=gray!5,dashed,text=gray!75!black] (n7) at (1.220,-5.800) {\textbf{C7}\\62.30};
\node[candidate] (n8) at (6.100,-4.350) {\textbf{C8}\\60.66};
\node[candidate] (n9) at (7.320,-1.450) {\textbf{C9}\\56.56};
\node[candidate] (n10) at (6.100,-5.800) {\textbf{C10}\\59.02};
\node[candidate,draw=red!65!black,fill=orange!12,line width=1pt,double,double distance=1pt,fill=orange!25,line width=1pt] (n11) at (2.440,-5.800) {\textbf{C11}\\68.85};
\node[candidate,draw=gray!60,fill=gray!5,dashed,text=gray!75!black] (n12) at (7.320,-2.900) {\textbf{C12}\\--};
\node[candidate,draw=gray!60,fill=gray!5,dashed,text=gray!75!black] (n13) at (3.660,-5.800) {\textbf{C13}\\--};
\node[candidate,draw=gray!60,fill=gray!5,dashed,text=gray!75!black] (n14) at (4.880,-5.800) {\textbf{C14}\\--};
\node[candidate,draw=gray!60,fill=gray!5,dashed,text=gray!75!black] (n15) at (8.540,-1.450) {\textbf{C15}\\55.74};
\node[candidate] (n16) at (6.100,-7.250) {\textbf{C16}\\60.66};
\draw[->,red!65!black,line width=1.2pt] (n0) -- node[pos=.58,fill=white,inner sep=1.2pt,text=blue!65!black,font=\sffamily\bfseries\fontsize{7}{8}\selectfont] {R} (n1);
\draw[->,gray!55,line width=.6pt] (n0) -- node[pos=.58,fill=white,inner sep=1.2pt,text=blue!65!black,font=\sffamily\bfseries\fontsize{7}{8}\selectfont] {R} (n2);
\draw[->,gray!55,line width=.6pt,dashed] (n1) -- node[pos=.58,fill=white,inner sep=1.2pt,text=blue!65!black,font=\sffamily\bfseries\fontsize{7}{8}\selectfont] {R} (n3);
\draw[->,gray!55,line width=.6pt] (n2) -- node[pos=.58,fill=white,inner sep=1.2pt,text=teal!80!black,font=\sffamily\bfseries\fontsize{7}{8}\selectfont] {A} (n4);
\draw[->,red!65!black,line width=1.2pt] (n1) -- node[pos=.58,fill=white,inner sep=1.2pt,text=teal!80!black,font=\sffamily\bfseries\fontsize{7}{8}\selectfont] {A} (n5);
\draw[->,red!65!black,line width=1.2pt] (n5) -- node[pos=.58,fill=white,inner sep=1.2pt,text=teal!80!black,font=\sffamily\bfseries\fontsize{7}{8}\selectfont] {A} (n6);
\draw[->,gray!55,line width=.6pt,dashed] (n6) -- node[pos=.58,fill=white,inner sep=1.2pt,text=blue!65!black,font=\sffamily\bfseries\fontsize{7}{8}\selectfont] {R} (n7);
\draw[->,gray!55,line width=.6pt] (n4) -- node[pos=.58,fill=white,inner sep=1.2pt,text=teal!80!black,font=\sffamily\bfseries\fontsize{7}{8}\selectfont] {A} (n8);
\draw[->,gray!55,line width=.6pt] (n0) -- node[pos=.58,fill=white,inner sep=1.2pt,text=blue!65!black,font=\sffamily\bfseries\fontsize{7}{8}\selectfont] {R} (n9);
\draw[->,gray!55,line width=.6pt] (n8) -- node[pos=.58,fill=white,inner sep=1.2pt,text=blue!65!black,font=\sffamily\bfseries\fontsize{7}{8}\selectfont] {R} (n10);
\draw[->,red!65!black,line width=1.2pt] (n6) -- node[pos=.58,fill=white,inner sep=1.2pt,text=blue!65!black,font=\sffamily\bfseries\fontsize{7}{8}\selectfont] {R} (n11);
\draw[->,gray!55,line width=.6pt,dashed] (n9) -- node[pos=.58,fill=white,inner sep=1.2pt,text=teal!80!black,font=\sffamily\bfseries\fontsize{7}{8}\selectfont] {A} (n12);
\draw[->,gray!55,line width=.6pt,dashed] (n6) -- node[pos=.58,fill=white,inner sep=1.2pt,text=blue!65!black,font=\sffamily\bfseries\fontsize{7}{8}\selectfont] {R} (n13);
\draw[->,gray!55,line width=.6pt,dashed] (n6) -- node[pos=.58,fill=white,inner sep=1.2pt,text=blue!65!black,font=\sffamily\bfseries\fontsize{7}{8}\selectfont] {R} (n14);
\draw[->,gray!55,line width=.6pt,dashed] (n0) -- node[pos=.58,fill=white,inner sep=1.2pt,text=blue!65!black,font=\sffamily\bfseries\fontsize{7}{8}\selectfont] {R} (n15);
\draw[->,gray!55,line width=.6pt] (n10) -- node[pos=.58,fill=white,inner sep=1.2pt,text=blue!65!black,font=\sffamily\bfseries\fontsize{7}{8}\selectfont] {R} (n16);
\end{tikzpicture}
\end{adjustbox}
\end{center}

%% file: appendix/curated/tree_15.tex
\begin{center}
\begin{adjustbox}{max width=\linewidth,max totalheight=.48\textheight}
\begin{tikzpicture}[x=1cm,y=1cm,>=Stealth,
candidate/.style={circle,minimum size=10.5mm,inner sep=1pt,align=center,font=\sffamily\fontsize{8.2}{9.2}\selectfont,draw=blue!45!black,fill=blue!5,line width=.55pt}]
\node[candidate,draw=red!65!black,fill=orange!12,line width=1pt] (n0) at (2.745,0.000) {\textbf{C0}\\29.00};
\node[candidate,draw=gray!60,fill=gray!5,dashed,text=gray!75!black] (n1) at (0.000,-1.450) {\textbf{C1}\\--};
\node[candidate,draw=red!65!black,fill=orange!12,line width=1pt] (n2) at (5.490,-1.450) {\textbf{C2}\\31.33};
\node[candidate] (n3) at (2.440,-2.900) {\textbf{C3}\\31.33};
\node[candidate,draw=gray!60,fill=gray!5,dashed,text=gray!75!black] (n4) at (1.220,-4.350) {\textbf{C4}\\--};
\node[candidate,draw=red!65!black,fill=orange!12,line width=1pt] (n5) at (4.880,-2.900) {\textbf{C5}\\34.67};
\node[candidate] (n6) at (6.710,-2.900) {\textbf{C6}\\35.67};
\node[candidate,draw=gray!60,fill=gray!5,dashed,text=gray!75!black] (n7) at (6.100,-4.350) {\textbf{C7}\\--};
\node[candidate,draw=red!65!black,fill=orange!12,line width=1pt,double,double distance=1pt,fill=orange!25,line width=1pt] (n8) at (4.880,-4.350) {\textbf{C8}\\38.00};
\node[candidate,draw=gray!60,fill=gray!5,dashed,text=gray!75!black] (n9) at (2.440,-4.350) {\textbf{C9}\\--};
\node[candidate] (n10) at (3.660,-4.350) {\textbf{C10}\\32.67};
\node[candidate] (n11) at (8.540,-2.900) {\textbf{C11}\\33.00};
\node[candidate] (n12) at (8.540,-4.350) {\textbf{C12}\\35.67};
\node[candidate] (n13) at (7.320,-4.350) {\textbf{C13}\\36.67};
\draw[->,gray!55,line width=.6pt,dashed] (n0) -- node[pos=.58,fill=white,inner sep=1.2pt,text=blue!65!black,font=\sffamily\bfseries\fontsize{7}{8}\selectfont] {R} (n1);
\draw[->,red!65!black,line width=1.2pt] (n0) -- node[pos=.58,fill=white,inner sep=1.2pt,text=blue!65!black,font=\sffamily\bfseries\fontsize{7}{8}\selectfont] {R} (n2);
\draw[->,gray!55,line width=.6pt] (n2) -- node[pos=.58,fill=white,inner sep=1.2pt,text=blue!65!black,font=\sffamily\bfseries\fontsize{7}{8}\selectfont] {R} (n3);
\draw[->,gray!55,line width=.6pt,dashed] (n3) -- node[pos=.58,fill=white,inner sep=1.2pt,text=teal!80!black,font=\sffamily\bfseries\fontsize{7}{8}\selectfont] {A} (n4);
\draw[->,red!65!black,line width=1.2pt] (n2) -- node[pos=.58,fill=white,inner sep=1.2pt,text=violet!80!black,font=\sffamily\bfseries\fontsize{7}{8}\selectfont] {S} (n5);
\draw[->,gray!55,line width=.6pt] (n2) -- node[pos=.58,fill=white,inner sep=1.2pt,text=violet!80!black,font=\sffamily\bfseries\fontsize{7}{8}\selectfont] {S} (n6);
\draw[->,gray!55,line width=.6pt,dashed] (n6) -- node[pos=.58,fill=white,inner sep=1.2pt,text=blue!65!black,font=\sffamily\bfseries\fontsize{7}{8}\selectfont] {R} (n7);
\draw[->,red!65!black,line width=1.2pt] (n5) -- node[pos=.58,fill=white,inner sep=1.2pt,text=teal!80!black,font=\sffamily\bfseries\fontsize{7}{8}\selectfont] {A} (n8);
\draw[->,gray!55,line width=.6pt,dashed] (n3) -- node[pos=.58,fill=white,inner sep=1.2pt,text=teal!80!black,font=\sffamily\bfseries\fontsize{7}{8}\selectfont] {A} (n9);
\draw[->,gray!55,line width=.6pt] (n3) -- node[pos=.58,fill=white,inner sep=1.2pt,text=blue!65!black,font=\sffamily\bfseries\fontsize{7}{8}\selectfont] {R} (n10);
\draw[->,gray!55,line width=.6pt] (n2) -- node[pos=.58,fill=white,inner sep=1.2pt,text=blue!65!black,font=\sffamily\bfseries\fontsize{7}{8}\selectfont] {R} (n11);
\draw[->,gray!55,line width=.6pt] (n11) -- node[pos=.58,fill=white,inner sep=1.2pt,text=violet!80!black,font=\sffamily\bfseries\fontsize{7}{8}\selectfont] {S} (n12);
\draw[->,gray!55,line width=.6pt] (n6) -- node[pos=.58,fill=white,inner sep=1.2pt,text=blue!65!black,font=\sffamily\bfseries\fontsize{7}{8}\selectfont] {R} (n13);
\end{tikzpicture}
\end{adjustbox}
\end{center}

%% file: appendix/curated/tree_16.tex
\begin{center}
\begin{adjustbox}{max width=\linewidth,max totalheight=.48\textheight}
\begin{tikzpicture}[x=1cm,y=1cm,>=Stealth,
candidate/.style={circle,minimum size=10.5mm,inner sep=1pt,align=center,font=\sffamily\fontsize{8.2}{9.2}\selectfont,draw=blue!45!black,fill=blue!5,line width=.55pt}]
\node[candidate,draw=red!65!black,fill=orange!12,line width=1pt] (n0) at (4.880,0.000) {\textbf{C0}\\29.00};
\node[candidate,draw=red!65!black,fill=orange!12,line width=1pt] (n1) at (1.830,-1.450) {\textbf{C1}\\31.33};
\node[candidate] (n2) at (7.930,-1.450) {\textbf{C2}\\34.00};
\node[candidate,draw=gray!60,fill=gray!5,dashed,text=gray!75!black] (n3) at (0.000,-2.900) {\textbf{C3}\\--};
\node[candidate,draw=gray!60,fill=gray!5,dashed,text=gray!75!black] (n4) at (4.880,-2.900) {\textbf{C4}\\--};
\node[candidate] (n5) at (1.220,-2.900) {\textbf{C5}\\32.00};
\node[candidate] (n6) at (6.100,-2.900) {\textbf{C6}\\35.67};
\node[candidate,draw=gray!60,fill=gray!5,dashed,text=gray!75!black] (n7) at (6.100,-4.350) {\textbf{C7}\\--};
\node[candidate] (n8) at (1.220,-4.350) {\textbf{C8}\\33.33};
\node[candidate] (n9) at (7.320,-2.900) {\textbf{C9}\\34.67};
\node[candidate,draw=gray!60,fill=gray!5,dashed,text=gray!75!black] (n10) at (8.540,-2.900) {\textbf{C10}\\--};
\node[candidate,draw=red!65!black,fill=orange!12,line width=1pt,double,double distance=1pt,fill=orange!25,line width=1pt] (n11) at (2.440,-2.900) {\textbf{C11}\\36.00};
\node[candidate] (n12) at (3.660,-2.900) {\textbf{C12}\\32.00};
\node[candidate] (n13) at (9.760,-2.900) {\textbf{C13}\\28.67};
\node[candidate] (n14) at (10.980,-2.900) {\textbf{C14}\\31.00};
\node[candidate] (n15) at (2.440,-4.350) {\textbf{C15}\\33.00};
\node[candidate] (n16) at (2.440,-5.800) {\textbf{C16}\\34.33};
\draw[->,red!65!black,line width=1.2pt] (n0) -- node[pos=.58,fill=white,inner sep=1.2pt,text=blue!65!black,font=\sffamily\bfseries\fontsize{7}{8}\selectfont] {R} (n1);
\draw[->,gray!55,line width=.6pt] (n0) -- node[pos=.58,fill=white,inner sep=1.2pt,text=blue!65!black,font=\sffamily\bfseries\fontsize{7}{8}\selectfont] {R} (n2);
\draw[->,gray!55,line width=.6pt,dashed] (n1) -- node[pos=.58,fill=white,inner sep=1.2pt,text=blue!65!black,font=\sffamily\bfseries\fontsize{7}{8}\selectfont] {R} (n3);
\draw[->,gray!55,line width=.6pt,dashed] (n2) -- node[pos=.58,fill=white,inner sep=1.2pt,text=teal!80!black,font=\sffamily\bfseries\fontsize{7}{8}\selectfont] {A} (n4);
\draw[->,gray!55,line width=.6pt] (n1) -- node[pos=.58,fill=white,inner sep=1.2pt,text=teal!80!black,font=\sffamily\bfseries\fontsize{7}{8}\selectfont] {A} (n5);
\draw[->,gray!55,line width=.6pt] (n2) -- node[pos=.58,fill=white,inner sep=1.2pt,text=teal!80!black,font=\sffamily\bfseries\fontsize{7}{8}\selectfont] {A} (n6);
\draw[->,gray!55,line width=.6pt,dashed] (n6) -- node[pos=.58,fill=white,inner sep=1.2pt,text=blue!65!black,font=\sffamily\bfseries\fontsize{7}{8}\selectfont] {R} (n7);
\draw[->,gray!55,line width=.6pt] (n5) -- node[pos=.58,fill=white,inner sep=1.2pt,text=teal!80!black,font=\sffamily\bfseries\fontsize{7}{8}\selectfont] {A} (n8);
\draw[->,gray!55,line width=.6pt] (n2) -- node[pos=.58,fill=white,inner sep=1.2pt,text=teal!80!black,font=\sffamily\bfseries\fontsize{7}{8}\selectfont] {A} (n9);
\draw[->,gray!55,line width=.6pt,dashed] (n2) -- node[pos=.58,fill=white,inner sep=1.2pt,text=blue!65!black,font=\sffamily\bfseries\fontsize{7}{8}\selectfont] {R} (n10);
\draw[->,red!65!black,line width=1.2pt] (n1) -- node[pos=.58,fill=white,inner sep=1.2pt,text=blue!65!black,font=\sffamily\bfseries\fontsize{7}{8}\selectfont] {R} (n11);
\draw[->,gray!55,line width=.6pt] (n1) -- node[pos=.58,fill=white,inner sep=1.2pt,text=teal!80!black,font=\sffamily\bfseries\fontsize{7}{8}\selectfont] {A} (n12);
\draw[->,gray!55,line width=.6pt] (n2) -- node[pos=.58,fill=white,inner sep=1.2pt,text=blue!65!black,font=\sffamily\bfseries\fontsize{7}{8}\selectfont] {R} (n13);
\draw[->,gray!55,line width=.6pt] (n2) -- node[pos=.58,fill=white,inner sep=1.2pt,text=blue!65!black,font=\sffamily\bfseries\fontsize{7}{8}\selectfont] {R} (n14);
\draw[->,gray!55,line width=.6pt] (n11) -- node[pos=.58,fill=white,inner sep=1.2pt,text=blue!65!black,font=\sffamily\bfseries\fontsize{7}{8}\selectfont] {R} (n15);
\draw[->,gray!55,line width=.6pt] (n15) -- node[pos=.58,fill=white,inner sep=1.2pt,text=blue!65!black,font=\sffamily\bfseries\fontsize{7}{8}\selectfont] {R} (n16);
\end{tikzpicture}
\end{adjustbox}
\end{center}

%% file: appendix/curated/tree_17.tex
\begin{center}
\begin{adjustbox}{max width=\linewidth,max totalheight=.48\textheight}
\begin{tikzpicture}[x=1cm,y=1cm,>=Stealth,
candidate/.style={circle,minimum size=10.5mm,inner sep=1pt,align=center,font=\sffamily\fontsize{8.2}{9.2}\selectfont,draw=blue!45!black,fill=blue!5,line width=.55pt}]
\node[candidate,draw=red!65!black,fill=orange!12,line width=1pt] (n0) at (4.270,0.000) {\textbf{C0}\\26.67};
\node[candidate,draw=gray!60,fill=gray!5,dashed,text=gray!75!black] (n1) at (0.000,-1.450) {\textbf{C1}\\--};
\node[candidate,draw=red!65!black,fill=orange!12,line width=1pt] (n2) at (3.965,-1.450) {\textbf{C2}\\27.00};
\node[candidate] (n3) at (1.220,-2.900) {\textbf{C3}\\25.33};
\node[candidate] (n4) at (3.660,-2.900) {\textbf{C4}\\30.67};
\node[candidate,draw=gray!60,fill=gray!5,dashed,text=gray!75!black] (n5) at (2.440,-4.350) {\textbf{C5}\\--};
\node[candidate,draw=red!65!black,fill=orange!12,line width=1pt] (n6) at (6.710,-2.900) {\textbf{C6}\\29.67};
\node[candidate] (n7) at (8.540,-1.450) {\textbf{C7}\\28.00};
\node[candidate] (n8) at (3.660,-4.350) {\textbf{C8}\\31.00};
\node[candidate,draw=red!65!black,fill=orange!12,line width=1pt,double,double distance=1pt,fill=orange!25,line width=1pt] (n9) at (6.100,-4.350) {\textbf{C9}\\33.33};
\node[candidate,draw=gray!60,fill=gray!5,dashed,text=gray!75!black] (n10) at (4.880,-4.350) {\textbf{C10}\\--};
\node[candidate] (n11) at (1.220,-4.350) {\textbf{C11}\\31.33};
\node[candidate,draw=gray!60,fill=gray!5,dashed,text=gray!75!black] (n12) at (6.100,-5.800) {\textbf{C12}\\--};
\node[candidate] (n13) at (7.320,-4.350) {\textbf{C13}\\31.33};
\node[candidate] (n14) at (7.320,-5.800) {\textbf{C14}\\30.33};
\draw[->,gray!55,line width=.6pt,dashed] (n0) -- node[pos=.58,fill=white,inner sep=1.2pt,text=blue!65!black,font=\sffamily\bfseries\fontsize{7}{8}\selectfont] {R} (n1);
\draw[->,red!65!black,line width=1.2pt] (n0) -- node[pos=.58,fill=white,inner sep=1.2pt,text=blue!65!black,font=\sffamily\bfseries\fontsize{7}{8}\selectfont] {R} (n2);
\draw[->,gray!55,line width=.6pt] (n2) -- node[pos=.58,fill=white,inner sep=1.2pt,text=blue!65!black,font=\sffamily\bfseries\fontsize{7}{8}\selectfont] {R} (n3);
\draw[->,gray!55,line width=.6pt] (n2) -- node[pos=.58,fill=white,inner sep=1.2pt,text=teal!80!black,font=\sffamily\bfseries\fontsize{7}{8}\selectfont] {A} (n4);
\draw[->,gray!55,line width=.6pt,dashed] (n4) -- node[pos=.58,fill=white,inner sep=1.2pt,text=blue!65!black,font=\sffamily\bfseries\fontsize{7}{8}\selectfont] {R} (n5);
\draw[->,red!65!black,line width=1.2pt] (n2) -- node[pos=.58,fill=white,inner sep=1.2pt,text=teal!80!black,font=\sffamily\bfseries\fontsize{7}{8}\selectfont] {A} (n6);
\draw[->,gray!55,line width=.6pt] (n0) -- node[pos=.58,fill=white,inner sep=1.2pt,text=blue!65!black,font=\sffamily\bfseries\fontsize{7}{8}\selectfont] {R} (n7);
\draw[->,gray!55,line width=.6pt] (n4) -- node[pos=.58,fill=white,inner sep=1.2pt,text=teal!80!black,font=\sffamily\bfseries\fontsize{7}{8}\selectfont] {A} (n8);
\draw[->,red!65!black,line width=1.2pt] (n6) -- node[pos=.58,fill=white,inner sep=1.2pt,text=violet!80!black,font=\sffamily\bfseries\fontsize{7}{8}\selectfont] {S} (n9);
\draw[->,gray!55,line width=.6pt,dashed] (n4) -- node[pos=.58,fill=white,inner sep=1.2pt,text=blue!65!black,font=\sffamily\bfseries\fontsize{7}{8}\selectfont] {R} (n10);
\draw[->,gray!55,line width=.6pt] (n3) -- node[pos=.58,fill=white,inner sep=1.2pt,text=violet!80!black,font=\sffamily\bfseries\fontsize{7}{8}\selectfont] {S} (n11);
\draw[->,gray!55,line width=.6pt,dashed] (n9) -- node[pos=.58,fill=white,inner sep=1.2pt,text=blue!65!black,font=\sffamily\bfseries\fontsize{7}{8}\selectfont] {R} (n12);
\draw[->,gray!55,line width=.6pt] (n6) -- node[pos=.58,fill=white,inner sep=1.2pt,text=blue!65!black,font=\sffamily\bfseries\fontsize{7}{8}\selectfont] {R} (n13);
\draw[->,gray!55,line width=.6pt] (n13) -- node[pos=.58,fill=white,inner sep=1.2pt,text=violet!80!black,font=\sffamily\bfseries\fontsize{7}{8}\selectfont] {S} (n14);
\end{tikzpicture}
\end{adjustbox}
\end{center}

%% file: appendix/curated/tree_18.tex
\begin{center}
\begin{adjustbox}{max width=\linewidth,max totalheight=.48\textheight}
\begin{tikzpicture}[x=1cm,y=1cm,>=Stealth,
candidate/.style={circle,minimum size=10.5mm,inner sep=1pt,align=center,font=\sffamily\fontsize{8.2}{9.2}\selectfont,draw=blue!45!black,fill=blue!5,line width=.55pt}]
\node[candidate,draw=red!65!black,fill=orange!12,line width=1pt] (n0) at (5.185,0.000) {\textbf{C0}\\26.67};
\node[candidate,draw=red!65!black,fill=orange!12,line width=1pt] (n1) at (3.050,-1.450) {\textbf{C1}\\29.33};
\node[candidate,draw=red!65!black,fill=orange!12,line width=1pt] (n2) at (0.000,-2.900) {\textbf{C2}\\30.00};
\node[candidate] (n3) at (3.355,-2.900) {\textbf{C3}\\27.67};
\node[candidate] (n4) at (1.830,-4.350) {\textbf{C4}\\30.33};
\node[candidate] (n5) at (3.660,-4.350) {\textbf{C5}\\24.67};
\node[candidate,draw=red!65!black,fill=orange!12,line width=1pt,double,double distance=1pt,fill=orange!25,line width=1pt] (n6) at (0.000,-4.350) {\textbf{C6}\\32.67};
\node[candidate,draw=gray!60,fill=gray!5,dashed,text=gray!75!black] (n7) at (7.320,-1.450) {\textbf{C7}\\--};
\node[candidate] (n8) at (4.880,-4.350) {\textbf{C8}\\31.67};
\node[candidate,draw=gray!60,fill=gray!5,dashed,text=gray!75!black] (n9) at (1.220,-5.800) {\textbf{C9}\\30.33};
\node[candidate] (n10) at (6.100,-2.900) {\textbf{C10}\\28.67};
\node[candidate] (n11) at (2.440,-5.800) {\textbf{C11}\\30.33};
\draw[->,red!65!black,line width=1.2pt] (n0) -- node[pos=.58,fill=white,inner sep=1.2pt,text=blue!65!black,font=\sffamily\bfseries\fontsize{7}{8}\selectfont] {R} (n1);
\draw[->,red!65!black,line width=1.2pt] (n1) -- node[pos=.58,fill=white,inner sep=1.2pt,text=blue!65!black,font=\sffamily\bfseries\fontsize{7}{8}\selectfont] {R} (n2);
\draw[->,gray!55,line width=.6pt] (n1) -- node[pos=.58,fill=white,inner sep=1.2pt,text=blue!65!black,font=\sffamily\bfseries\fontsize{7}{8}\selectfont] {R} (n3);
\draw[->,gray!55,line width=.6pt] (n3) -- node[pos=.58,fill=white,inner sep=1.2pt,text=teal!80!black,font=\sffamily\bfseries\fontsize{7}{8}\selectfont] {A} (n4);
\draw[->,gray!55,line width=.6pt] (n3) -- node[pos=.58,fill=white,inner sep=1.2pt,text=blue!65!black,font=\sffamily\bfseries\fontsize{7}{8}\selectfont] {R} (n5);
\draw[->,red!65!black,line width=1.2pt] (n2) -- node[pos=.58,fill=white,inner sep=1.2pt,text=teal!80!black,font=\sffamily\bfseries\fontsize{7}{8}\selectfont] {A} (n6);
\draw[->,gray!55,line width=.6pt,dashed] (n0) -- node[pos=.58,fill=white,inner sep=1.2pt,text=blue!65!black,font=\sffamily\bfseries\fontsize{7}{8}\selectfont] {R} (n7);
\draw[->,gray!55,line width=.6pt] (n3) -- node[pos=.58,fill=white,inner sep=1.2pt,text=teal!80!black,font=\sffamily\bfseries\fontsize{7}{8}\selectfont] {A} (n8);
\draw[->,gray!55,line width=.6pt,dashed] (n4) -- node[pos=.58,fill=white,inner sep=1.2pt,text=blue!65!black,font=\sffamily\bfseries\fontsize{7}{8}\selectfont] {R} (n9);
\draw[->,gray!55,line width=.6pt] (n1) -- node[pos=.58,fill=white,inner sep=1.2pt,text=blue!65!black,font=\sffamily\bfseries\fontsize{7}{8}\selectfont] {R} (n10);
\draw[->,gray!55,line width=.6pt] (n4) -- node[pos=.58,fill=white,inner sep=1.2pt,text=teal!80!black,font=\sffamily\bfseries\fontsize{7}{8}\selectfont] {A} (n11);
\end{tikzpicture}
\end{adjustbox}
\end{center}

%% file: appendix/curated/tree_19.tex
\begin{center}
\begin{adjustbox}{max width=\linewidth,max totalheight=.48\textheight}
\begin{tikzpicture}[x=1cm,y=1cm,>=Stealth,
candidate/.style={circle,minimum size=10.5mm,inner sep=1pt,align=center,font=\sffamily\fontsize{8.2}{9.2}\selectfont,draw=blue!45!black,fill=blue!5,line width=.55pt}]
\node[candidate,draw=red!65!black,fill=orange!12,line width=1pt] (n0) at (5.947,0.000) {\textbf{C0}\\40.67};
\node[candidate,draw=red!65!black,fill=orange!12,line width=1pt] (n1) at (2.745,-1.450) {\textbf{C1}\\44.00};
\node[candidate] (n2) at (0.000,-2.900) {\textbf{C2}\\46.33};
\node[candidate] (n3) at (1.830,-2.900) {\textbf{C3}\\46.33};
\node[candidate,draw=red!65!black,fill=orange!12,line width=1pt] (n4) at (5.490,-2.900) {\textbf{C4}\\46.67};
\node[candidate] (n5) at (0.000,-4.350) {\textbf{C5}\\47.67};
\node[candidate] (n6) at (3.660,-4.350) {\textbf{C6}\\46.33};
\node[candidate] (n7) at (1.220,-4.350) {\textbf{C7}\\48.00};
\node[candidate] (n8) at (4.880,-4.350) {\textbf{C8}\\45.67};
\node[candidate] (n9) at (6.100,-4.350) {\textbf{C9}\\47.00};
\node[candidate,draw=red!65!black,fill=orange!12,line width=1pt,double,double distance=1pt,fill=orange!25,line width=1pt] (n10) at (7.320,-4.350) {\textbf{C10}\\48.67};
\node[candidate] (n11) at (2.440,-4.350) {\textbf{C11}\\46.67};
\node[candidate] (n12) at (3.660,-5.800) {\textbf{C12}\\46.33};
\node[candidate] (n13) at (9.150,-1.450) {\textbf{C13}\\35.00};
\node[candidate] (n14) at (8.540,-2.900) {\textbf{C14}\\43.33};
\node[candidate,draw=gray!60,fill=gray!5,dashed,text=gray!75!black] (n15) at (9.760,-2.900) {\textbf{C15}\\40.33};
\draw[->,red!65!black,line width=1.2pt] (n0) -- node[pos=.58,fill=white,inner sep=1.2pt,text=blue!65!black,font=\sffamily\bfseries\fontsize{7}{8}\selectfont] {R} (n1);
\draw[->,gray!55,line width=.6pt] (n1) -- node[pos=.58,fill=white,inner sep=1.2pt,text=teal!80!black,font=\sffamily\bfseries\fontsize{7}{8}\selectfont] {A} (n2);
\draw[->,gray!55,line width=.6pt] (n1) -- node[pos=.58,fill=white,inner sep=1.2pt,text=teal!80!black,font=\sffamily\bfseries\fontsize{7}{8}\selectfont] {A} (n3);
\draw[->,red!65!black,line width=1.2pt] (n1) -- node[pos=.58,fill=white,inner sep=1.2pt,text=teal!80!black,font=\sffamily\bfseries\fontsize{7}{8}\selectfont] {A} (n4);
\draw[->,gray!55,line width=.6pt] (n2) -- node[pos=.58,fill=white,inner sep=1.2pt,text=teal!80!black,font=\sffamily\bfseries\fontsize{7}{8}\selectfont] {A} (n5);
\draw[->,gray!55,line width=.6pt] (n4) -- node[pos=.58,fill=white,inner sep=1.2pt,text=violet!80!black,font=\sffamily\bfseries\fontsize{7}{8}\selectfont] {S} (n6);
\draw[->,gray!55,line width=.6pt] (n3) -- node[pos=.58,fill=white,inner sep=1.2pt,text=teal!80!black,font=\sffamily\bfseries\fontsize{7}{8}\selectfont] {A} (n7);
\draw[->,gray!55,line width=.6pt] (n4) -- node[pos=.58,fill=white,inner sep=1.2pt,text=violet!80!black,font=\sffamily\bfseries\fontsize{7}{8}\selectfont] {S} (n8);
\draw[->,gray!55,line width=.6pt] (n4) -- node[pos=.58,fill=white,inner sep=1.2pt,text=violet!80!black,font=\sffamily\bfseries\fontsize{7}{8}\selectfont] {S} (n9);
\draw[->,red!65!black,line width=1.2pt] (n4) -- node[pos=.58,fill=white,inner sep=1.2pt,text=teal!80!black,font=\sffamily\bfseries\fontsize{7}{8}\selectfont] {A} (n10);
\draw[->,gray!55,line width=.6pt] (n3) -- node[pos=.58,fill=white,inner sep=1.2pt,text=teal!80!black,font=\sffamily\bfseries\fontsize{7}{8}\selectfont] {A} (n11);
\draw[->,gray!55,line width=.6pt] (n6) -- node[pos=.58,fill=white,inner sep=1.2pt,text=blue!65!black,font=\sffamily\bfseries\fontsize{7}{8}\selectfont] {R} (n12);
\draw[->,gray!55,line width=.6pt] (n0) -- node[pos=.58,fill=white,inner sep=1.2pt,text=blue!65!black,font=\sffamily\bfseries\fontsize{7}{8}\selectfont] {R} (n13);
\draw[->,gray!55,line width=.6pt] (n13) -- node[pos=.58,fill=white,inner sep=1.2pt,text=violet!80!black,font=\sffamily\bfseries\fontsize{7}{8}\selectfont] {S} (n14);
\draw[->,gray!55,line width=.6pt,dashed] (n13) -- node[pos=.58,fill=white,inner sep=1.2pt,text=teal!80!black,font=\sffamily\bfseries\fontsize{7}{8}\selectfont] {A} (n15);
\end{tikzpicture}
\end{adjustbox}
\end{center}

%% file: appendix/curated/tree_20.tex
\begin{center}
\begin{adjustbox}{max width=\linewidth,max totalheight=.48\textheight}
\begin{tikzpicture}[x=1cm,y=1cm,>=Stealth,
candidate/.style={circle,minimum size=10.5mm,inner sep=1pt,align=center,font=\sffamily\fontsize{8.2}{9.2}\selectfont,draw=blue!45!black,fill=blue!5,line width=.55pt}]
\node[candidate,draw=red!65!black,fill=orange!12,line width=1pt] (n0) at (3.355,0.000) {\textbf{C0}\\40.67};
\node[candidate,draw=red!65!black,fill=orange!12,line width=1pt] (n1) at (1.830,-1.450) {\textbf{C1}\\30.67};
\node[candidate] (n2) at (0.000,-2.900) {\textbf{C2}\\35.67};
\node[candidate,draw=red!65!black,fill=orange!12,line width=1pt] (n3) at (1.830,-2.900) {\textbf{C3}\\32.33};
\node[candidate,draw=red!65!black,fill=orange!12,line width=1pt,double,double distance=1pt,fill=orange!25,line width=1pt] (n4) at (1.830,-4.350) {\textbf{C4}\\45.67};
\node[candidate,draw=gray!60,fill=gray!5,dashed,text=gray!75!black] (n5) at (1.220,-5.800) {\textbf{C5}\\--};
\node[candidate] (n6) at (3.660,-2.900) {\textbf{C6}\\41.67};
\node[candidate] (n7) at (4.880,-1.450) {\textbf{C7}\\43.33};
\node[candidate] (n8) at (3.660,-4.350) {\textbf{C8}\\42.00};
\node[candidate] (n9) at (2.440,-5.800) {\textbf{C9}\\43.33};
\node[candidate] (n10) at (0.000,-4.350) {\textbf{C10}\\36.67};
\draw[->,red!65!black,line width=1.2pt] (n0) -- node[pos=.58,fill=white,inner sep=1.2pt,text=blue!65!black,font=\sffamily\bfseries\fontsize{7}{8}\selectfont] {R} (n1);
\draw[->,gray!55,line width=.6pt] (n1) -- node[pos=.58,fill=white,inner sep=1.2pt,text=blue!65!black,font=\sffamily\bfseries\fontsize{7}{8}\selectfont] {R} (n2);
\draw[->,red!65!black,line width=1.2pt] (n1) -- node[pos=.58,fill=white,inner sep=1.2pt,text=blue!65!black,font=\sffamily\bfseries\fontsize{7}{8}\selectfont] {R} (n3);
\draw[->,red!65!black,line width=1.2pt] (n3) -- node[pos=.58,fill=white,inner sep=1.2pt,text=teal!80!black,font=\sffamily\bfseries\fontsize{7}{8}\selectfont] {A} (n4);
\draw[->,gray!55,line width=.6pt,dashed] (n4) -- node[pos=.58,fill=white,inner sep=1.2pt,text=blue!65!black,font=\sffamily\bfseries\fontsize{7}{8}\selectfont] {R} (n5);
\draw[->,gray!55,line width=.6pt] (n1) -- node[pos=.58,fill=white,inner sep=1.2pt,text=teal!80!black,font=\sffamily\bfseries\fontsize{7}{8}\selectfont] {A} (n6);
\draw[->,gray!55,line width=.6pt] (n0) -- node[pos=.58,fill=white,inner sep=1.2pt,text=blue!65!black,font=\sffamily\bfseries\fontsize{7}{8}\selectfont] {R} (n7);
\draw[->,gray!55,line width=.6pt] (n6) -- node[pos=.58,fill=white,inner sep=1.2pt,text=teal!80!black,font=\sffamily\bfseries\fontsize{7}{8}\selectfont] {A} (n8);
\draw[->,gray!55,line width=.6pt] (n4) -- node[pos=.58,fill=white,inner sep=1.2pt,text=blue!65!black,font=\sffamily\bfseries\fontsize{7}{8}\selectfont] {R} (n9);
\draw[->,gray!55,line width=.6pt] (n2) -- node[pos=.58,fill=white,inner sep=1.2pt,text=blue!65!black,font=\sffamily\bfseries\fontsize{7}{8}\selectfont] {R} (n10);
\end{tikzpicture}
\end{adjustbox}
\end{center}